\documentclass[sigconf]{acmart}

\usepackage{subfigure}
\usepackage{balance}
\usepackage[ruled,linesnumbered,noline,noend]{algorithm2e}
\usepackage[english]{babel}
\SetKwInOut{Input}{Input}
\SetKwInOut{Output}{Output}
\SetKwInOut{Parameters}{Parameters\hspace{0.15em}}

\AtBeginDocument{%
  \providecommand\BibTeX{{%
    \normalfont B\kern-0.5em{\scshape i\kern-0.25em b}\kern-0.8em\TeX}}}

\copyrightyear{2025}
\acmYear{2025}
\setcopyright{acmlicensed}
\acmConference[KDD '25]{Proceedings of the 31st ACM SIGKDD Conference on Knowledge Discovery and Data Mining V.1}{August 3--7, 2025}{Toronto, ON, Canada}
\acmBooktitle{Proceedings of the 31st ACM SIGKDD Conference on Knowledge Discovery and Data Mining V.1 (KDD '25), August 3--7, 2025, Toronto, ON, Canada}
\acmDOI{10.1145/3690624.3709268}
\acmISBN{979-8-4007-1245-6/25/08}

\begin{document}

\title{Simplicial SMOTE: Oversampling Solution to the Imbalanced Learning Problem}

\author{Oleg Kachan \orcid{0000-0002-7679-755X}}
\affiliation{
  \institution{Sber AI Lab}
  \city{Moscow}
  \country{Russia}
}
\affiliation{
 \institution{HSE University}
 \city{Moscow}
 \country{Russia}
}
\email{oleg.n.kachan@gmail.com}
\author{Andrey Savchenko \orcid{0000-0001-6196-0564}}
\affiliation{
  \institution{Sber AI Lab}
  \city{Moscow}
  \country{Russia}
}
\affiliation{
 \institution{HSE University}
 \city{Moscow}
 \country{Russia}
}
\email{avsavchenko@hse.ru}
\author{Gleb Gusev \orcid{0009-0003-7298-1848}}
\affiliation{
  \institution{Sber AI Lab}
  \city{Moscow}
  \country{Russia}
}
\email{gleb57@gmail.com}

\begin{abstract}
SMOTE (Synthetic Minority Oversampling Technique) is the established geometric approach to random oversampling to balance classes in the imbalanced learning problem, followed by many extensions. Its idea is to introduce synthetic data points of the minor class, with each new point being the convex combination of an existing data point and one of its $k$-nearest neighbors.

In this paper, by viewing SMOTE as sampling from the edges of a geometric neighborhood graph and borrowing tools from the topological data analysis, we propose a novel technique, Simplicial SMOTE, that samples from the simplices of a geometric neighborhood simplicial complex. A new synthetic point is defined by the barycentric coordinates w.r.t. a simplex spanned by an arbitrary number of data points being sufficiently close rather than a pair. Such a replacement of the geometric data model results in better coverage of the underlying data distribution compared to existing geometric sampling methods and allows the generation of synthetic points of the minority class closer to the majority class on the decision boundary.

We experimentally demonstrate that our Simplicial SMOTE outperforms several popular geometric sampling methods, including the original SMOTE. Moreover, we show that simplicial sampling can be easily integrated into existing SMOTE extensions. We generalize and evaluate simplicial extensions of the classic Borderline SMOTE, Safe-level SMOTE, and ADASYN algorithms, all of which outperform their graph-based counterparts.
\end{abstract}


\begin{teaserfigure}
 \includegraphics[width=\textwidth]{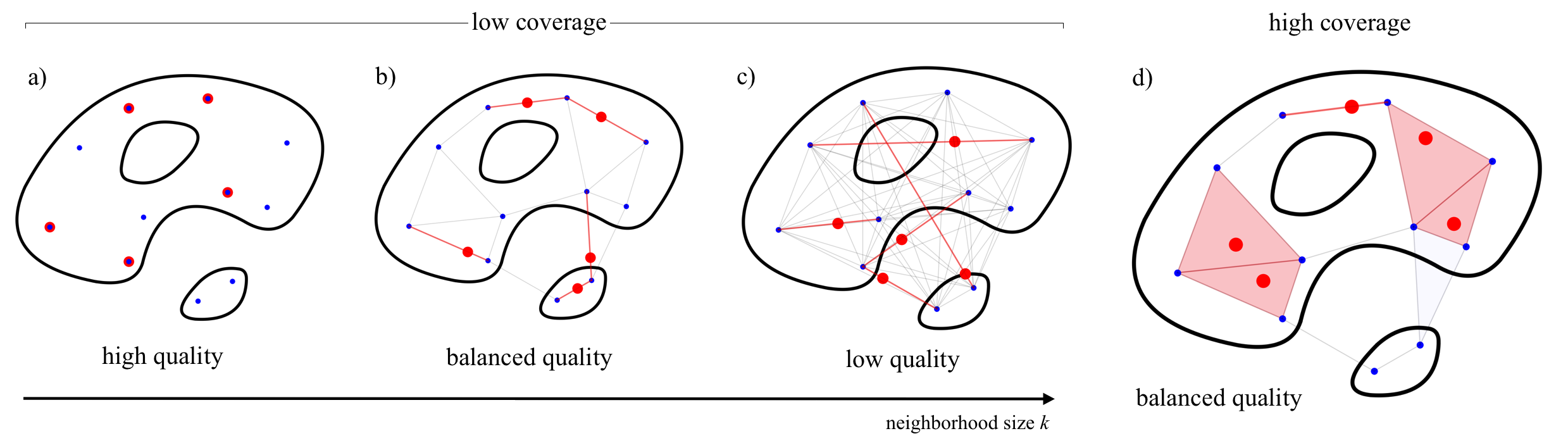}
 \caption{
 Geometric oversampling algorithms: a) random oversampling, b) SMOTE, c) global sampling, d) Simplicial SMOTE. With no inductive assumptions on data, random oversampling just duplicates existing points. Assuming that synthetic data points lie within a convex hull of existing points, global methods do not respect the intrinsic properties of data such as clusters and holes, resulting in low sample quality. While SMOTE, being a local method, improves on this, it still models the data with a union of one-dimensional segments, unable to sample all of the data support. Simplicial SMOTE, by modeling data with a union of higher-dimensional simplices, samples dense areas of the data support while avoiding sampling from topological holes, effectively improving coverage of the data distribution. 
 True data distribution is shown in black, data points are shown in blue, geometric graph- or simplicial-based models are shown in light gray, selected edges or simplices to sample from, and sampled synthetic points are shown in red.
 }
 \label{figure:teaser}
 \Description{Geometric oversampling algorithms: a) random oversampling, b) SMOTE, c) global sampling, d) Simplicial SMOTE (proposed)}
\end{teaserfigure}

\maketitle




\section{Introduction}

The imbalanced learning problem is learning from data when the minority class is dominated by the majority one \cite{chen2024survey}. Many problems in data analysis are inherently imbalanced in areas such as finance (fraud detection) \cite{Wang2019}, marketing (churn prediction) \cite{Liu2018},  medicine (medical diagnosis) \cite{Han2019}, industry (predictive maintenance) \cite{Sridhar2021}, image recognition \cite{savchenko2016search,savchenko2020ad}, etc. Often, the rare minority class (a credit fraud, a canceled subscription, the presence of a disease, an equipment failure) is of much more interest than the more common majority one. The class imbalance causes the bias of a classifier towards the majority class \cite{Wallace2011}, as the naive classifier assigning all data points to the majority class will achieve an accuracy equal to the majority class proportion.


Many techniques exist for the imbalanced learning problem, including undersampling and oversampling \cite{bespalov2022lambo}. Several resampling methods are geometric in nature (Fig.~\ref{figure:teaser}), having in common the reliance on a geometric model of data, i.e., introducing new points within a neighborhood of existing data points or by their interpolation. Geometric resampling methods differ in terms of neighborhood size or locality. For random oversampling \cite{Batista2004} that duplicates the existing points, the neighborhood of each point includes only the point itself. Global sampling \cite{Zhang2018} introduces new synthetic points as the convex combination of randomly selected pairs of points. Here, the neighborhood of each point includes all points of its class. Synthetic Minority Oversampling Technique (SMOTE) \cite{Chawla2002} introduces new synthetic points as the convex combination of pairs consisting of a data point and its nearest neighbors. Thus, the neighborhood of each point includes points of its class being sufficiently close.

From the data modeling standpoint, geometric resampling replaces the original empirical distribution with data-augmented density. Such an approach proved to improve the solution of the original problem if the density estimator parameters are chosen correctly \cite{Chapelle2000}. Data models can be quantified by two dual metrics \cite{Kynkaanniemi2019}. First, sample quality (precision) is how well it models the data, answering the question, ``How many model samples are within the data support?'' Second, data coverage (recall) is how well the model covers the data, answering the question, ``How many data samples are within the model support?''. Often modeling of a subset on the decision boundary will be sufficient, as it is not necessary to model the whole minor class distribution for the discriminative downstream tasks \cite{Han2005}.

We highlight the issues of existing geometric resampling methods to be addressed. First, the low data coverage of low-dimensional geometric models which use single or pairs of points to generate synthetic ones. Second, the low sample quality of global neighborhood methods for topologically and geometrically complex data distributions, as by modeling data globally by a convex hull they do not respect the topology (multiple clusters and topological holes) and local geometry (curved areas) of the data distribution. 

\begin{figure*}[t!] 

\includegraphics[width=\textwidth]{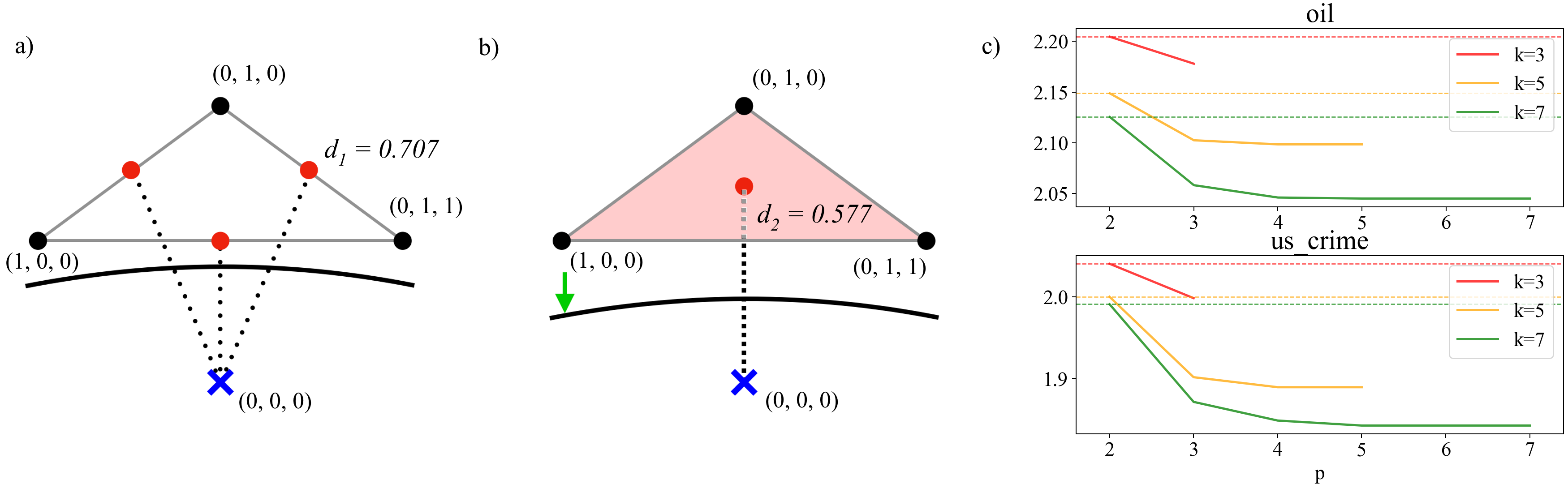}

\vspace{-0.5em}
\caption{For the configuration of three points of the minor class (black circles) equidistant to a point of the major class (blue cross) b) Simplicial SMOTE will generate synthetic points of the minor class (red circles) closer to the point of the major class (projection distance to the 2-simplex $d_2 = 0.577$), than a) SMOTE (projection distance to any edge $d_1 = 0.707$), effectively moving the local decision boundary. c) Mean projection distance to the geometric model of minority class gets smaller with increasing maximal relation arity parameter $p$. Distance to the simplicial model is shown as solid lines for different values of neighborhood size parameter $k$, distance to the graph model is shown as a dashed line of the same color.
}
\label{figure:decision_boundary}
\Description{Visualization sample of a) SMOTE and b) Simplicial SMOTE}
\end{figure*}

Existing geometric sampling methods, either global or local, have low coverage due to modeling data with a union of one-dimensional segments or edges of a geometric neighborhood graph. Such graphs model the data as the union of one-dimensional segments, which is insufficient to sample from high-dimensional spaces. For example, even for a two-dimensional dataset, one could not introduce samples from the entire convex hull spanned by data points using SMOTE or global sampling. Instead, if we model the data with the union of convex regions whose dimension is equal to the feature space, for example, a union of triangles, which are two-dimensional simplices, we could sample it, see Figure \ref{figure:teaser}.

Moreover, sampling from a simplex on the borderline between classes will result in synthetic points of the minor class being closer to the points of the major class compared to sampling from its edges only, as shown in Figure \ref{figure:decision_boundary}. To get the idea, consider the standard simplex $(x, y, z)$ in $\mathbb{R}^3$, with coordinates of spanning points $x=(1, 0, 0)$, $y=(0, 1, 0)$ and $z=(0, 0, 1)$ (Fig. \ref{figure:decision_boundary}, left). Then the orthogonal projection of the origin $(0, 0, 0)$ onto the edge $(x, y)$ would be $p_1 = (0.5, 0.5, 0)$, with the distance to the origin $d_1 = 0.7071$, and the orthogonal projection of the origin to the simplex $p_2 = (1/3, 1/3, 1/3)$, with the distance to the origin $d_2 = 0.5774$. Hence, generating points by considering higher-dimensional simplices (Fig. \ref{figure:decision_boundary}, right) would result in synthetic points of the minor class being closer to the points of the majority class, effectively moving the decision boundary away from the minor class.





With this in mind, we introduce the generalization of SMOTE, namely Simplicial SMOTE \footnote{Code is available at: \url{https://github.com/oleg-kachan/simplicial-smote-kdd25}.}, modeling the data with a union of higher-dimensional simplices of the clique complex of a neighborhood graph. That is, a position of a new synthetic point is defined by the barycentric coordinates w.r.t. a simplex spanned by an arbitrary number of data points being sufficiently close, i.e., being in the $p$-ary neighborhood relation, effectively increasing the data coverage and moving the decision boundary.

\subsubsection*{Our contribution:}

\begin{itemize}
    \item We propose the novel geometric oversampling approach, Simplicial SMOTE, in which new points are sampled from simplices of a geometric neighborhood simplicial complex. As a result, the true data distribution is better covered. Moreover, synthetic points in the minority class can be generated closer to the majority class data points.

    \item We experimentally demonstrate that the proposed technique is characterized by a significant increase in performance for various classifiers and datasets. Compared to the original SMOTE, our simplicial generalization achieves 4.5\% improvement in F1 score on average and up to 29.3\% individually (``car\_eval\_4'' dataset) for k-NN, and 5\% improvement on average and up to 25.7\% individually (``oil'' dataset) for the gradient boosting classifier.

    \item As the proposed simplicial sampling is orthogonal to the sampling scheme of SMOTE, we have shown how the known variants, such as Borderline SMOTE, Safe-level SMOTE, and ADASYN, can be generalized to use the simplicial sampling. We provided their evaluation, with all simplicial extensions outperforming their graph-based counterparts.
\end{itemize}

\section{Related work}





The original SMOTE algorithm introduces synthetic points from the geometric model of the minority class. Several variants of SMOTE instead propose to sample synthetic points from the minority class part of the decision manifold, i.e., the minority points lying on the boundary between classes. The decision manifold is estimated in several ways. For example,  Borderline SMOTE \citep{Han2005} estimates the decision manifold by taking the minority class local density around each minority data point. The SVM SMOTE \citep{Nguyen2011} first takes the points corresponding to the support vectors of the SVM classifier.



In Safe-level SMOTE \citep{Bunkhumpornpat2009}, a value called safe level ratio is assigned to each edge of the neighborhood graph built over minority class instances, which is the ratio of the numbers of minority class instances for a point $x$ and its neighbor $x'$. If the number of the minority class instances in the neighborhoods of $x$ and $x'$ are zero, no synthetic examples are generated from that edge. Otherwise, a new synthetic sample is a convex combination of the points, and the coefficient depends on the ratio, being close to the minority example with more neighbors of the minority class.

In ADASYN \citep{He2008}, for each minority point, a ratio of majority examples in the neighborhood is computed. The new points are the convex combination of minority class points, with the number of synthetic examples generated using a given minority example being inversely proportional to that ratio.

MWMOTE \citep{Barua2012} first identifies the hard-to-learn informative minority class samples and then generates the synthetic samples from the weighted informative minority class samples using a clustering approach. In Density-based SMOTE (DBSMOTE) \citep{Bunkhumpornpat2012}, minority class examples are partitioned into disjointed clusters by the DBSCAN algorithm \citep{Ester1996}. The new points are the random convex combinations of two points from the random edge of the shortest path connecting minority points with the pseudo-centroid point, which is the closest to the cluster centroid. LVQ-SMOTE \citep{Nakamura2013} oversamples the minority class, first approximating is using a set of prototype points obtained by LVQ (Learning Vector Quantization) algorithm \cite{DeVries2016}.

Global sampling, seen as a geometric method using a complete graph as the data model, introduces new synthetic points as a convex combination of a pair of existing points randomly chosen from a dataset \citep{Zhang2018}. Fitting parametric distributions to data, such as the Gaussian distribution, is also used for the minority class oversampling in the imbalanced data classification problem \citep{Xie2020}.
\section{Proposed approach}





Our work improves the SMOTE modeling and sampling scheme by modeling data with a geometric simplicial complex \citep{Boissonnat2018,Dey2022}, which is the higher-dimensional generalization of a graph. Contrary to global sampling methods or fitting Gaussian distribution, it respects local topological features of data such as clusters and topological holes \cite{kachan2020persistent}. Geometric sampling methods assume that a synthetic point combines (several) existing data point(s). When designing such algorithms, one should decide upon 1) a neighborhood size of each data point, ranging from a point itself to all points from the dataset, and 2) a set of data points used to synthesize a new point. Neighborhood relations can describe the former, while the latter corresponds to the relation arity. While popular sampling techniques model data with a complete or local graph based on the binary neighborhood relations, our choice is to model the data with a simplicial complex based on neighborhood relations of arity greater than $2$.

Consider a complete graph $H_n$ with a vertex set $X \in \mathbb{R}^d$ of cardinality $n$. A \emph{neighborhood graph} $G = (X, E) \subseteq H_n$ is a subgraph of $H_n$ such that the edge set $E \subseteq {n \choose 2}$ is instantiated according to a relation $R$ defining a neighborhood of each point $\mathcal{N}(x) = \{ x' \mid xRx' \}$.

For example, let $X$ be endowed with a distance function $d: X \times X \rightarrow \mathbb{R}_+$. A (symmetrized) $k$-nearest neighbor relation $R^{k\mathrm{NN}}$ on $X$ defining a \emph{$k$-nearest neighbors neighborhood graph} parameterized by $k \in \mathbb{N} \setminus \{0\}$ is
\begin{equation}
    R^{k\mathrm{NN}}(k) = \left\{\hspace{0.15em}(x, y)~\big\vert~d(x, y) \leq \textrm{min}_k d(x, z),~z \in X \hspace{0.15em}\right\},
\end{equation}
where $\min_k(\cdot)$ denotes the $k$-th minimum, hence $\arg\min_k(x, z)$ is the $k$-th neighbor of $x$.

An $\varepsilon$-ball relation $R^{\varepsilon}$ on $X$ defining the \emph{$\varepsilon$-ball neighborhood graph} given a scale parameter $\varepsilon \in \mathbb{R}_{\geq 0}$ is
\begin{equation}
    R^{\varepsilon}(\varepsilon) = \left\{(x, y)~\big\vert~d(x, y) \leq \varepsilon \right\},
\end{equation}
meaning that balls $B_x(\varepsilon/2) \cap B_{x'}(\varepsilon/2) \neq \emptyset$ of radius $\varepsilon/2$ centered at $x$ and $x'$ intersect.

Given a binary relation $R$, a \emph{$p$-ary relation} $P$ of $R$ is defined as a subset $Y$ of $X$ of cardinality $p$ such that $Y \times Y \subseteq R$, that is, a set $Y$ iff $xRx'$ for any pair $(x, x') \in Y$. A \emph{maximal $p$-ary relation} $C$ of $R$ defined as a subset $Y$ of $X$ of cardinality $p$ that is maximal concerning inclusion \citep{Henry2011}.



Whether a binary neighborhood relation corresponds to an edge in a neighborhood graph, a $p$-ary relation corresponds to a graph's $(p+1)$-clique, more generally, a $p$-simplex in a neighborhood simplicial complex over the vertex set $X$. Points can belong to more than one simplex. All simplices containing a point $x$ are subsets of its neighborhood $\mathcal{N}(x)$ \citep{Henry2011}.

\subsection{Simplicial SMOTE}

We propose a simple, yet effective generalization of SMOTE by considering a general $p$-ary neighborhood relation. That is, instead of a binary relation leading to neighborhood graphs, we believe the $p$-ary relation leads to a neighborhood simplicial complex, resulting in a high-dimensional data model, contrary to a graph that is locally $1$-dimensional.

Consider a dataset $\mathcal{X} = \{ (\mathbf{x}_i, y_i) \}_{i \in 1, \dots, n}$, where $\mathbf{x}_i \in \mathbb{R}^d$ and $y_i \in \{-1, +1\}$. By convention, we denote the \emph{minority class} $\mathcal{X}^+ = \{ \mathbf{x}_i \mid y_i = +1 \}_{i \in 1, \dots, n^+}$ as positive, and the \emph{majority class} $\mathcal{X}^- = \{ \mathbf{x}_j \mid y_j = -1 \}_{j\in 1, \dots, n^-}$ as negative of sizes $n^+ < n^-$ respectively, with $n = n^+ + n^-$. To balance classes, we need to introduce $m = n^- - n^+$ synthetic points of minor class $\mathcal{\hat{X}}^+ = \{\ ( \mathbf{\hat{x}}_\ell, y_\ell = +1) \}_{\ell \in 1, \dots, m}$.  

\subsubsection*{Constructing a simplicial complex from data}

The neighborhood simplicial complex can be viewed dually. Combinatorially, it is a collection of subsets of a given set, satisfying the closure of a neighborhood relation. Geometrically, it is a union of convex hulls of those subsets from which to sample synthetic points.

Given a set $X$, an \emph{(abstract) simplicial complex} $K$ is a collection of subsets of $X$ called simplices such that if a simplex $\sigma$ is in $K$, then all of its subsets $\tau \subseteq \sigma$ are also in $K$. That is, combinatorially a \emph{$p$-simplex} $\sigma$ is a subset of $p+1$ points of $X$. We say that a $p$-simplex is of \emph{dimension} $p$.

Neighborhood simplicial complexes can be constructed naively, by definition, by enumerating all subsets of $X$ to check whether they satisfy a closure of neighborhood relation, i.e., whether all pairs of a subset satisfy a binary relation. Zomorodian \cite{Zomorodian2010} showed that it is related to the clique enumeration problems in the neighborhood graphs. That is, given any graph $G$, its \emph{clique complex} is a simplicial complex K(G), which has the same vertices and edges as $G$, and $(p+1)$-cliques of $G$ are $p$-simplices of $K(G)$.

For example, the \emph{Vietoris-Rips complex} is the clique complex of the $\varepsilon$-ball neighborhood graph. We consider the clique complex of the symmetric $k$-nearest neighbor neighborhood graph for our algorithm, with the number of nearest neighbors $k$ being the first hyperparameter.


A \emph{$p$-skeleton} $L$ of a simplicial complex $K$ is the subcomplex of $K$ with the dimension of simplices at most $p$. Algorithmically, this corresponds to finding cliques up to dimension $p+1$ instead of maximal cliques.

\subsubsection*{Sampling from a simplicial complex}

To obtain $m$ synthetic points, we first sample uniformly $m$ maximal simplices from the $p$-skeleton of the clique complex $K(G)$ with replacement, followed by sampling a single point from each simplex.

Let $\mathrm{\Lambda}_p$ be the set of all vectors of $p+1$ elements, such that $\lambda_i \geq 0$ and $\sum_{i=0}^k \lambda_i = 1$. Given a set of $p+1$ points $\{\mathbf{x}_i\}_{i=0}^p$ in an $d$-dimensional Euclidean space, represented by a matrix $\mathbf{X} \in \mathbb{R}^{p \times d}$, a \emph{geometric $p$-simplex} $\sigma$ is defined

\begin{equation}
    \sigma_{\mathbf{x}_0,\dots,\mathbf{x}_p} = \left\{ \sum_{i=0}^p \lambda_i \mathbf{x}_i \bigm\vert \boldsymbol{\lambda} \in \mathrm{\Lambda}_p \right\}.
\end{equation}

We call the elements of $\boldsymbol{\lambda}$ \emph{barycentric coordinates} w.r.t. the points spanning a simplex. Barycentric coordinates could be mapped into Euclidean coordinates, resulting in a synthetic point:
\begin{align}
    \hat{\mathbf{x}}_{\mathbf{x}_0,\dots,\mathbf{x}_p}(\boldsymbol{\lambda})
    &= \lambda_0\mathbf{x}_0 + \dots + \lambda_p\mathbf{x}_p\\
    \nonumber &= \boldsymbol{\lambda}^T \mathbf{X}.
\end{align}

To sample uniformly from a $p$-simplex, we sample barycentric coordinates $\boldsymbol{\lambda} \in \mathbb{R}^{p+1}$ according to the symmetric Dirichlet distribution $\boldsymbol{\lambda} \sim Dir(\boldsymbol{\alpha})$, where $\boldsymbol{\lambda} = (1, \dots, 1) \in \mathbb{R}^{p+1}$.

\begin{algorithm}[t]

\caption{Simplicial SMOTE}
\label{algorithm}

\Input{Minority class points $\mathbf{X}^+$.}
\Parameters{Neighborhood size $k$,\\maximal relation arity $p \geq k$,\\
}
\Output{Synthetic minority class points $\hat{\mathbf{X}}^+$.}
Construct a k-NN neighborhood graph $G_k(X^+)$.\\
Compute a $p$-skeleton $(K_p \circ G_k)(X^+)$ of a clique complex $(K \circ G_k)(X^+)$, get its maximal simplices $\Sigma_p^{MAX}$\\
Sample $m = n^- - n^+$ simplices $\sigma_i^{(p_i)}$ of dimension $p_i$, $\Sigma = \{ \sigma^{(p_i)}_i \}_{i \in 1,\cdots,m}$ from $\Sigma_{p}^{MAX}$.\\
\For{$i \in 1,\dots,m$} {
    Sample barycentric coordinates $\boldsymbol{\lambda}_i \sim \mathrm{Dir}(\boldsymbol{\alpha})$, where $\boldsymbol{\alpha} = (1, \dots, 1) \in \mathbb{R}^{(p+1)}$.\\
    Compute Euclidean coordinates $\hat{\mathbf{x}}_i = \boldsymbol{\lambda}_i^T \mathbf{X}_i$ w.r.t. a simplex $\sigma^{(p_i)}_i = (\mathbf{x}_0, \dots, \mathbf{x}_{p_i}) \in \Sigma$ of dimension $p_i$.
}

\Return{$\{ \hat{\mathbf{x}}_i\}_{i \in 1,...,m}$}

\end{algorithm}


We outline the Simplicial SMOTE method in Algorithm \ref{algorithm}. Note that it has only two hyperparameters: the neighborhood size ($k$ for kNN neighborhood graph) and the maximal arity of neighborhood relation $p$. It is worth noting that the proposed sampling scheme is orthogonal to the original SMOTE and its known modifications and could be used to complement them. Let us consider the details in the next Subsection.

\begin{figure*}[t] 
\includegraphics[width=\textwidth]{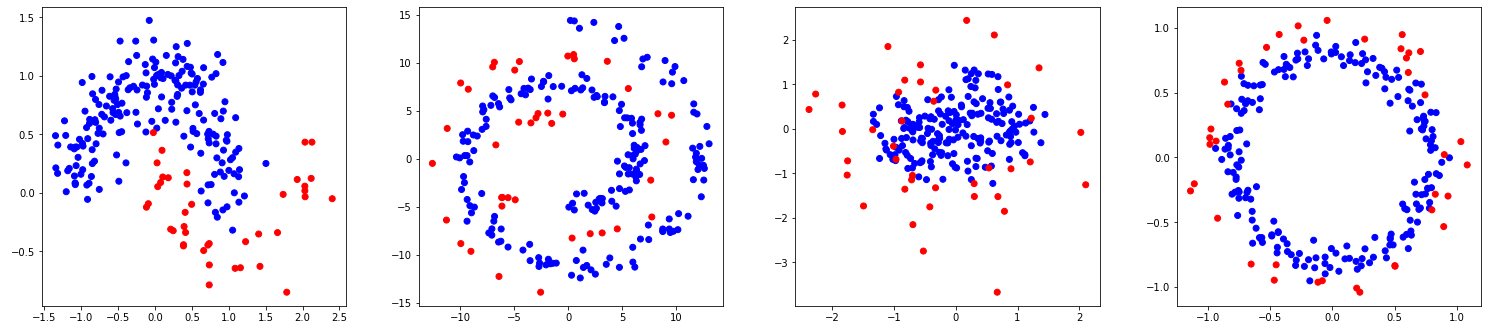}
\vspace{-0.75em}
\caption{Synthetic data: a) moons, b) swiss rolls, c) a Gaussian inside a sphere, d) a sphere inside a sphere.}

\label{figure:synthetic}
\Description{Synthetic data: a) moons, b) swiss rolls, c) a Gaussian inside a sphere, d) a sphere inside a sphere.}
\end{figure*}


\subsection{Simplicial generalizations of SMOTE variants}




The original SMOTE algorithm constructs the minority neighborhood graph and samples points from its edges without considering the majority class. Several variants of the SMOTE algorithm improve reinforcing the points close to the boundary between types by considering the density of the majority class relative to the points from the minority. We generalize the known modifications of the SMOTE mentioned in the previous Section, namely, Borderline SMOTE \cite{Han2005}, Safe-level SMOTE \cite{Bunkhumpornpat2009} and ADASYN \cite{He2008}, to use the simplicial sampling scheme. We denote the \emph{minority neighborhood} of $\mathbf{x}_i$ are the points of the minority class within a given neighborhood of a point $\mathcal{N}^+(\mathbf{x}_i) = \{ \mathbf{x}_j \mid \mathbf{x}_i \sim \mathbf{x}_j, y_j = 1 \}_{j \in 1, \dots, k^+}$, and the \emph{majority neighborhood} of $\mathbf{x}_i$ as $\mathcal{N}^-(\mathbf{x}_i) = \{ \mathbf{x}_j \mid \mathbf{x}_i \sim \mathbf{x}_j, y_j = 0 \}_{j \in 1, \dots, k^-}$ of sizes $k^+$ and $k^-$ respectively. The majority and minority points ratios within a given neighborhood are defined as $\mathrm{\Delta^+}(\mathbf{x}_i) = k^+/k$ and $\mathrm{\Delta^-}(\mathbf{x}_i) = k^-/k$ respectively.

\subsubsection{Simplicial Borderline SMOTE}

The extension assumption is that the examples on the borderline and the ones nearby are more apt to be misclassified than the ones far from the borderline and, thus, more important for classification. The examples far from the borderline may contribute little to classification results.

The \emph{borderline subset} of the minority class $B(\mathcal{X}^+)$ is defined

\begin{equation}
    B(\mathcal{X}^+) = \left\{ \mathbf{x}_i, y_i = 0 \bigm\vert \frac{|\mathcal{N}^+(\mathbf{x}_i)|}{|\mathcal{N}(\mathbf{x}_i)|} < 1/2, |\mathcal{N}^+(\mathbf{x}_i)| \neq 0 \right\}
\end{equation}
that is the points whose the larger part of the nearest neighbors belong to the majority class, except those whose nearest neighbors are completely majority class instances and are considered noise. The new points are the convex combination of the simplices of a simplicial complex built upon the borderline points and their nearest neighbors from the minority class.

\subsubsection{Simplicial Safe-level SMOTE}

The original SMOTE algorithm considers sampling from a $k$-simplex according to the Dirichlet distribution $\mathrm{Dir}(\boldsymbol{\alpha})$, where $\boldsymbol{\alpha} \in \mathbb{R}_{> 0}^k$. Without any further assumptions, the distribution is symmetric, i.e., all of the vector $\boldsymbol{\alpha}$ elements have the same value (usually $1$, resulting in the uniform distribution on a simplex). Safe-level SMOTE modifies the elements of $\boldsymbol{\alpha}$ by setting them based on the ratio of majority neighborhood ratios, resulting in synthetic points being generated closer to safer minority points, i.e., having a larger proportion of neighbors of the same class. A simplicial generalization is to set the parameter $\alpha_i = 1 / \Delta^+(x_i)$.

\subsubsection{Simplicial ADASYN}

While Borderline SMOTE answers the question from which simplex to sample, selecting simplices spanned by borderline points, and Safe-level SMOTE answers the question from where precisely on a simplex to sample sampling closer to safer points, ADASYN answers the question of how much to sample from a simplex, inversely proportional to the average safety of points. Therefore, its simplicial generalization is to average arbitrary safety values instead of just a pair.

Borderline Simplicial SMOTE benefited most from the proposed sampling scheme, showing increased performance relative to both original SMOTE and Borderline SMOTE.

\begin{table}[b!]

\caption{Synthetic data classification results.}
\label{table:results_synthetic}

\centering

\renewcommand{\arraystretch}{1.15}

\resizebox{\columnwidth}{!}{

\begin{tabular}{lrrrrrr}
\toprule
{} &  \fontsize{8}{9}\selectfont\textbf{Imbalanced} &  \fontsize{8}{9}\selectfont\textbf{Gaussian} &  \fontsize{8}{9}\selectfont\textbf{Random} &   \fontsize{8}{9}\selectfont\textbf{Global} &   \fontsize{8}{9}\selectfont\textbf{SMOTE} &      \fontsize{8}{9}\selectfont\textbf{Simplicial} \\ \midrule
moons           &      0.9511 &    0.8830 &  0.9485 &  0.9348 &  \textbf{0.9694} &  \textbf{0.9694} \\
swiss\_rolls     &      0.5317 &    0.6673 &  0.7168 &  0.6774 &  \textbf{0.7208} &  0.6823 \\
g\_circle &      0.7129 &    0.6750 &  0.7089 &  0.6542 &  0.6937 &  \textbf{0.7269} \\
circles         &      0.6541 &    0.7060 &  0.6777 &  0.6356 &  0.7005 &  \textbf{0.7139} \\ \midrule
\textbf{rank}            &      4.0000 &    4.5000 &  3.2500 &  5.2500 &  2.3750 &  \textbf{1.6250} \\ \bottomrule
\end{tabular}

}

\end{table}

\subsection{Complexity analysis} 

The algorithm's complexity depends on the complexity of the neighborhood graph construction and expansion. The naive nearest neighbor search has a complexity of $O(n^2)$, while the approximated nearest neighbor search lowers it to $O(n)$.

From the computational complexity perspective, the difference between SMOTE and Simplicial SMOTE is the clique finding step in the neighborhood graph, such as the k-nearest neighbor graph. The computational complexity of the clique finding step depends only on the density of the neighborhood graph, which is controlled by the hyperparameter ($k$), but not by the data dimensionality. Indeed, enumerating all maximal cliques in a graph with $n$ vertices and $m$ edges is an NP-complete problem, requiring exponential time in the worst case. Up to $n^{n/3}$ maximal cliques exist in a graph with $n$ vertices \citep{Moon1965}. Yet, as the neighborhood graphs are sparse, various bounds were given regarding the number of edges, node degree, and arboricity of a graph. In a graph with maximum degree $\delta$ the time complexity of maximal clique enumeration (MCE) is $O(\delta^4)$ per clique and $O((n - \delta)3^{\delta/3}\delta^4)$ total \citep{Makino2004,Eppstein2013}. The \emph{arboricity} is the minimum number of edge-disjoint spanning forests into which the graph can be decomposed. For a graph of arboricity $a$, the complexity of MCE is $O(am)$ \citep{Chiba1985}.


Enumeration of all cliques up to the size $p$ can be done in either inductive, incremental, or top-down enumeration approach after solving the MCE problem \citep{Zomorodian2010}. Recently, an algorithm conjectured to be optimal was proposed for this task, which was shown to be approximately a magnitude faster than the incremental algorithm in practice \cite{Rieser2023}.

\section{Results}

\subsection{Synthetic data}

First, we evaluated the proposed method, comparing it with the original SMOTE \citep{Chawla2002}, sampling from the Gaussian distribution fitted to data \citep{Xie2020}, as well as random \citep{Batista2004} and global oversampling \citep{Zhang2018} algorithms on several synthetic datasets to emphasize the importance of modeling the data locally, as well as the advantage of the simplicial complex data model.

As the synthetic data, we have generated the following datasets partially using the \texttt{scikit-learn} library \citep{Pedregosa2011}, shown in Figure \ref{figure:synthetic}: moons, swiss rolls, a Gaussian inside a circle, and a circle inside a circle. As all model datasets have complex topological and geometric structure, global methods are conjectured to underperform by generating synthetic points of minority class within the support of the majority class, hence of low quality.

All synthetic datasets consist of $n = 350$ points, with the size of the minority class $n^+ = 50$ (shown in red) and the size of the majority class $n^- = 300$ (shown in blue), i.e., class imbalance ratio is equal to $6$.

For SMOTE and Simplicial SMOTE, we performed a grid search for the neighborhood size parameter $k$ of the kNN neighborhood graph ranging from $3$ to $8$ with a step $1$. We report the F1 score averaged over $5$ runs using $4$-fold cross-validation in Table \ref{table:results_synthetic} for the $k$-nearest neighbors classifier with default hyperparameters from the \texttt{scikit-learn} library \citep{Pedregosa2011}.

Results show when data is of complex topological and geometric structure, global methods such as global sampling from a complete graph or fitting a Gaussian distribution underperform, having the low sample quality compared to local techniques such as SMOTE and Simplicial SMOTE, as well as the simple random oversampling. Simplicial SMOTE has performed the best, achieving the highest rank among all sampling methods, and is generally better than its original graph-based counterpart.

\subsection{Real data}

\begin{table*}[h!]

\caption{Classification results on benchmark datasets for the $k$-NN classifier. F1 score averaged over $5$ repeats of $5$-fold (outer) cross-validation is reported. \textcolor{red}{\textbf{Best}} and \textcolor{blue}{\textbf{second-best}} results are highlighted. Results are \underline{underlined} when the Simplicial SMOTE and simplicial generalizations of Borderline SMOTE, Safe-level SMOTE and ADASYN methods are better or equal to SMOTE and the original versions, respectively.}
\label{table:knn_f1}


\centering

\renewcommand{\arraystretch}{1.2}

\resizebox{\textwidth}{!}{

\begin{tabular}{l|rrrrrrrrrr|rrrr}
\toprule
 & \fontsize{8}{9}\selectfont\textbf{Imbalanced} & \fontsize{8}{9}\selectfont\textbf{Random} & \fontsize{8}{9}\selectfont\textbf{Global} & \fontsize{8}{9}\selectfont\textbf{SMOTE} & \fontsize{8}{9}\selectfont\textbf{Border.} & \fontsize{8}{9}\selectfont\textbf{Safelevel} & \fontsize{8}{9}\selectfont\textbf{ADASYN} & \fontsize{8}{9}\selectfont\textbf{MWMOTE} & \fontsize{8}{9}\selectfont\textbf{DBSMOTE} & \fontsize{8}{9}\selectfont\textbf{LVQ} & \fontsize{8}{9}\selectfont\textbf{Simplicial} & \fontsize{8}{9}\selectfont\textbf{S-Border.} & \fontsize{8}{9}\selectfont\textbf{S-Safe.} & \fontsize{8}{9}\selectfont\textbf{S-ADASYN} \\  \midrule

ecoli           &     0.5780 &     0.5501 &     0.5864 &     0.5822 &     0.5853 &     \underline{0.5781} &     0.5705 &     0.5980 &     0.6336 &     0.5827 &     \underline{0.6275} &     \underline{0.6151} &     0.5688 &     \underline{0.6280} \\
optical\_digits &     0.9670 &     0.9491 &     0.9427 &     0.9415 &     \underline{0.9559} &     \underline{0.9439} &     \underline{0.9442} &     0.9376 &     0.9491 &     0.9618 &     \underline{0.9443} &     0.9557 &     0.9425 &     0.9423 \\
pen\_digits     &     0.9927 &     0.9906 &     0.9895 &     0.9906 &     0.9925 &     0.9900 &     \underline{0.9917} &     0.9907 &     0.9922 &     0.9928 &     \underline{0.9915} &     \underline{0.9925} &     \underline{0.9912} &     0.9911 \\
abalone         &     0.1808 &     0.3326 &     0.3842 &     0.3501 &     0.3586 &     \underline{0.3727} &     0.3448 &     0.3753 &     0.3281 &     0.3078 &     \underline{0.3698} &     \underline{0.3725} &     0.3614 &     \underline{0.3594} \\
sick\_euthyroid &     0.5565 &     0.5694 &     0.5872 &     0.5708 &     0.5684 &     0.5246 &     0.5650 &     0.5654 &     0.6081 &     0.5800 &     \underline{0.6049} &     \underline{0.5981} &     \underline{0.5962} &     \underline{0.6047} \\
spectrometer    &     0.7618 &     0.8493 &     0.8382 &     0.8430 &     \underline{0.8543} &     0.8274 &     0.8423 &     0.8551 &     0.7968 &     0.8209 &     \underline{0.8548} &     0.8485 &     \underline{0.8386} &     \underline{0.8465} \\
car\_eval\_34   &     0.6018 &     0.5830 &     0.5718 &     0.5774 &     0.5913 &     0.5886 &     0.5851 &     0.6490 &     0.5830 &     0.7165 &     \underline{0.6296} &     \underline{0.6081} &     \underline{0.6321} &     \underline{0.6341} \\
us\_crime       &     0.3676 &     0.4429 &     0.4404 &     0.4188 &     0.4567 &     \underline{0.4346} &     0.4144 &     0.4062 &     0.4429 &     0.4634 &     \underline{0.4313} &     \underline{0.4687} &     0.4210 &     \underline{0.4236} \\
yeast\_ml8      &     0.0375 &     0.1426 &     0.1613 &     0.1592 &     \underline{0.1659} &     0.1542 &     0.1578 &     0.1603 &     0.1426 &     0.1643 &     \underline{0.1598} &     0.1658 &     \underline{0.1601} &     \underline{0.1586} \\
scene           &     0.1004 &     0.2500 &     0.2499 &     \underline{0.2386} &     0.2471 &     \underline{0.2452} &     \underline{0.2309} &     0.2364 &     0.1101 &     0.2579 &     0.2251 &     \underline{0.2482} &     0.2219 &     0.2259 \\
libras\_move    &     0.6997 &     0.8031 &     0.7702 &     \underline{0.7672} &     \underline{0.7661} &     0.7362 &     0.7519 &     0.7874 &     0.8031 &     0.8029 &     0.7560 &     0.7568 &     \underline{0.7640} &     \underline{0.7525} \\
thyroid\_sick   &     0.4966 &     0.5239 &     0.5246 &     0.5255 &     0.5307 &     0.4620 &     0.5214 &     0.5245 &     0.5095 &     0.5031 &     \underline{0.5504} &     \underline{0.5459} &     \underline{0.5271} &     \underline{0.5579} \\
coil\_2000      &     0.0457 &     0.1745 &     0.1726 &     0.1709 &     \underline{0.1759} &     0.1677 &     0.1711 &     0.1748 &     0.0533 &     0.1168 &     \underline{0.1714} &     0.1748 &     \underline{0.1703} &     \underline{0.1743} \\
solar\_flare\_m0 &     0.0510 &     0.2192 &     0.2043 &     0.2077 &     0.2262 &     0.2280 &     \underline{0.2188} &     0.2120 &     0.0486 &     0.2126 &     \underline{0.2189} &     \underline{0.2337} &     \underline{0.2340} &     0.2171 \\
oil             &     0.3156 &     0.4467 &     0.4592 &     0.4428 &     0.4674 &     0.3784 &     0.4240 &     0.4191 &     0.4467 &     0.4626 &     \underline{0.5074} &    \underline{0.5062} &     \underline{0.4267} &     \underline{0.4777} \\
car\_eval\_4    &     0.1294 &     0.3815 &     0.4810 &     0.4443 &     0.4472 &     0.4121 &     0.4371 &     0.5240 &     0.3815 &     0.6771 &     \underline{0.5749} &     \underline{0.5773} &     \underline{0.6052} &     \underline{0.5676} \\
wine\_quality   &     0.1292 &     0.2983 &     0.2097 &     \underline{0.2558} &     \underline{0.2774} &     0.2460 &     \underline{0.2537} &     0.2163 &     0.1487 &     0.2244 &     0.2533 &     0.2741 &     \underline{0.2731} &     0.2534 \\
letter\_img     &     0.9722 &     0.9526 &     0.9086 &     0.9410 &     0.9608 &     0.9293 &     \underline{0.9531} &     0.9128 &     0.9661 &     0.9673 &     \underline{0.9556} &     \underline{0.9610} &     \underline{0.9547} &     0.9491 \\
yeast\_me2      &     0.2296 &     0.3192 &     0.2705 &     0.3043 &     0.3621 &     0.2943 &     0.3018 &     0.3227 &     0.2952 &     0.2890 &     \underline{0.3364} &     \underline{0.3759} &     \underline{0.3057} &     \underline{0.3279} \\
ozone\_level    &     0.1608 &     0.2528 &     0.2086 &     0.2087 &     0.2270 &     \underline{0.2516} &     \underline{0.2095} &     0.2108 &     0.2528 &     0.2352 &     \underline{0.2111} &     \underline{0.2443} &     0.2056 &     0.2046 \\
abalone\_19     &     0.0000 &     0.0277 &     0.0500 &     0.0377 &     \underline{0.0482} &     0.0289 &     0.0408 &     0.0366 &     0.0212 &     0.0312 &     \underline{0.0512} &     0.0452 &     \underline{0.0434} &     \underline{0.0498} \\ \midrule
\textbf{mean}   &     0.3988 &     0.4790 &     0.4767 &     0.4751 &     0.4888 &     0.4664 &     0.4729 &     0.4817 &     0.4530 &     0.4938 &     \textcolor{blue}{\textbf{\underline{0.4964}}} &     \textcolor{red}{\textbf{\underline{0.5032}}} &     \underline{0.4878} &     \underline{0.4927} \\
\textbf{rank}   &    11.5714 &     7.3095 &     8.0476 &     9.0000 &     \textcolor{blue}{\textbf{4.6905}} &     9.7143 &     9.2381 &     7.5714 &     8.5476 &     6.2381 &     \underline{5.3333} &     \textcolor{red}{\textbf{\underline{3.6429}}} &     \underline{7.4286} &     \underline{6.6667} \\ \bottomrule

\end{tabular}

}



\end{table*}

\begin{figure*}[h!] 

\centering
\includegraphics[width=0.9\textwidth]{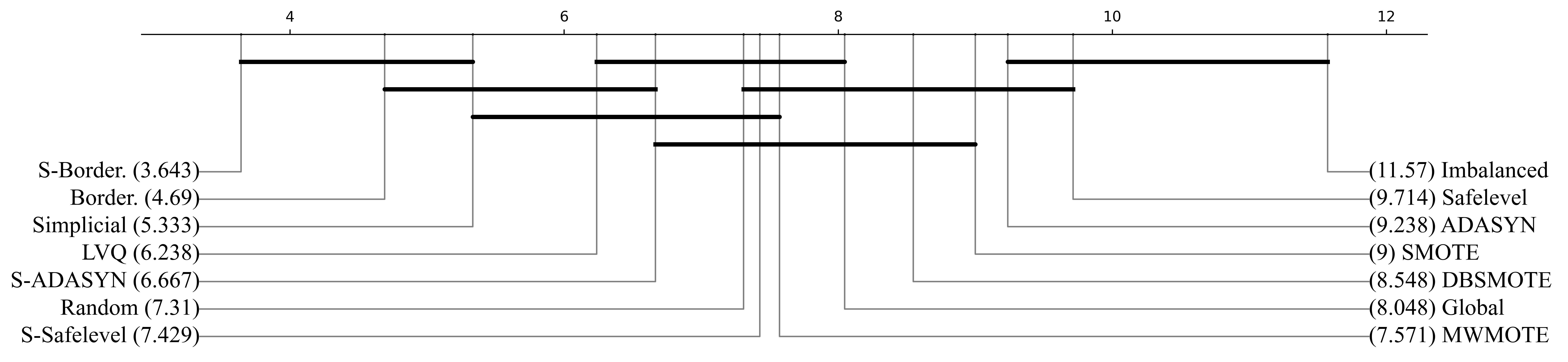}

\vspace{-0.5em}
\caption{Critical difference diagram for the $k$-NN classifier and F1 score.}
\label{cd_knn_f1}
\Description{Critical difference diagram for the k-NN classifier and F1 score.}
\end{figure*}


In this Subsection, we compared the proposed Simplicial SMOTE method to the original SMOTE, random and global oversampling, as well as the simplicial generalizations of classic SMOTE extensions such as Borderline SMOTE, Safe-level SMOTE, and ADASYN to its original versions. We also included several more recent geometric sampling methods, such as MWMOTE, DBSMOTE, and LVQ-SMOTE, all of which had achieved rank one for at least one combination of a classifier and metric in the extensive evaluation of $85$ SMOTE extensions \cite{Kovacs2019}.

As most of the existing works on SMOTE and its variants, we considered the binary classification case, yet our method can be used to handle the multi-class scenario by oversampling all classes except the major one. The evaluation was performed on $21$ benchmark datasets from UCI/LIBSVM repositories common in the imbalanced learning literature (Table~\ref{table:datasets}). The class imbalance ratio ranges from $9$ to $130$. Data dimensionality ranges from $7$ to $294$. Each dataset was normalized to zero mean and unit variance.


We used nested cross-validation with the inner cross-validation with $25$ repeats and $4$ splits for samplers hyperparameter search, and the outer cross-validation with $5$ repeats and $5$ splits for model evaluation. For all SMOTE-based methods, we performed a grid search for the neighborhood size parameter $k$ of the kNN neighborhood graph within a dataset-specific range, depending on the number of data points and features. The neighborhood size $k$ ranged from $3$ to $\lceil \sqrt[3]{n^+} + \log{d} \rceil$ with a step $2$, where $n^+$ is the minority class size and $d$ is the dimension of the dataset. For Simplicial SMOTE and the simplicial generalizations of Borderline SMOTE, Safe-level SMOTE, and ADASYN, analyzed the optimal value of an additional hyperparameter, namely, maximal simplex dimensionality $p$. The maximal simplex dimension $p$ ranged from $3$ to $k$, with a step $1$. The dependence on hyperparameters $k$ and $p$ of the classification performance in terms of F1 score is presented in Appendix \ref{appendix_sensitivity}.

\begin{figure*}[h!] 

\centering
\includegraphics[width=0.9\textwidth]{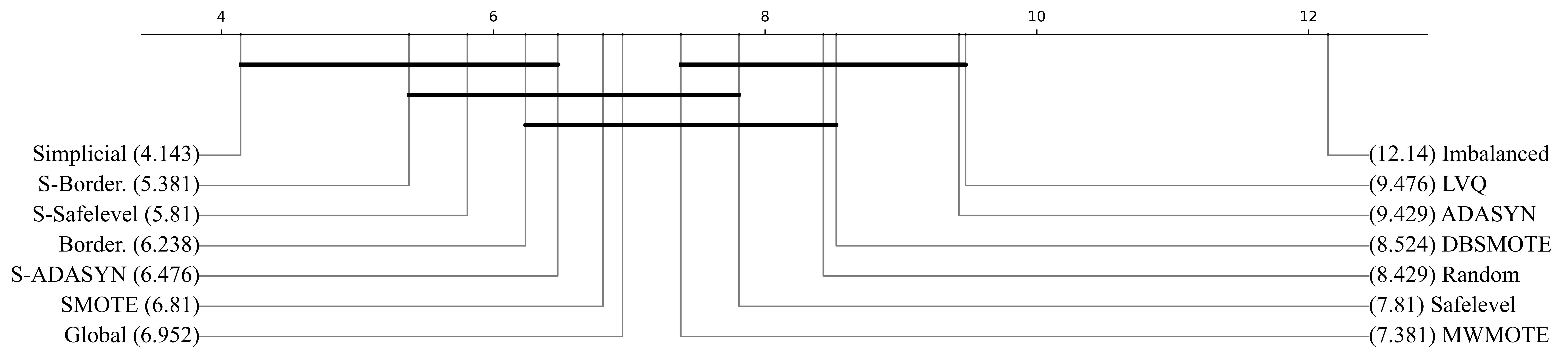}

\vspace{-0.5em}

\caption{Critical difference diagram for the gradient boosting classifier and F1 score.}
\label{cd_gb_f1}
\Description{Critical difference diagram for the gradient boosting classifier and F1 score.}
\end{figure*}

We report the F1 score for classifiers from the scikit-learn library, namely, $k$-nearest neighbors (k-NN) (Table \ref{table:knn_f1}), gradient boosting (Table \ref{table:gb_f1}). We used default hyperparameters for $k$-nearest neighbor classifier and set maximum tree depth to $2$ for gradient boosting~\cite{prokhorenkova2018catboost}. We performed statistical significance testing using the Friedman test with Conover post-hoc analysis \cite{Demvsar2006}. We provide the critical difference diagrams for k-NN and gradient boosting classifiers in Figs.~\ref{cd_knn_f1},~\ref{cd_gb_f1}, respectively.



In addition to F1-score, we considered the Matthew's correlation coefficient (MCC) scores in Appendix \ref{appendix_more_results} as the complementary metric. Indeed, F$1$ score emphasizes the correct classification of the minor class, while MCC considers all the four rates of the confusion matrix \cite{Chicco2020}. 

Classification results on benchmark imbalanced datasets show the advantage of the proposed Simplicial SMOTE method over its competitors, including the original SMOTE in terms of F1 and Matthew's correlation coefficient scores in terms of ranks and the mean value of the metrics across all datasets. Compared to the original SMOTE, our simplicial generalization achieves 4.5\% improvement in F1 score on average and up to 29.3\% individually (``car\_eval\_4'' dataset) for k-NN, and 5\% improvement on average and up to 25.7\% individually (``oil'' dataset) for the gradient boosting classifier. The simplicial generalizations of Borderline SMOTE, Safe-level SMOTE, and ADASYN also outperformed their original versions.

\subsection{Running time}

We provide the running time of an experiment on benchmark datasets. We run $5$-fold cross-validation repeated $5$ times for the neighborhood size parameter k=10. The maximal simplex dimension $p$ was set to $3$. Computation was run on 2x Intel(R) Xeon(R) Gold 6248R CPU @ 3.00GHz system, with 48 cores and 96 threads total. SMOTE, k=10 -- 15.03 sec (0.65 sec per dataset on average), Simplicial SMOTE, k=10, p=3 -- 20.79 sec (0.90 sec per dataset on average). Hence, the running time is only $1.4$ times slower for the Simplicial SMOTE compared to the original SMOTE algorithm on the benchmark datasets. However, while our approach takes more time to oversample the dataset due to the additional clique search step to build the simplicial complex model of data, in the overall pipeline, oversampling takes only a fraction of the time compared to the classifier fitting, especially for complex techniques, such as gradient boosting~\cite{prokhorenkova2018catboost}.


\section{Conclusion}

\begin{table*}[h!]

\caption{Classification results on benchmark datasets for the gradient boosting classifier. F1 score averaged over $5$ repeats of $5$-fold (outer) cross-validation is reported. \textcolor{red}{\underline{\textbf{Best}}} and \textcolor{blue}{\textbf{second-best}} results are highlighted. Results are \underline{underlined} when the Simplicial SMOTE and simplicial generalizations of Borderline SMOTE, Safe-level SMOTE and ADASYN methods are better or equal to SMOTE and the original versions, respectively.}
\label{table:gb_f1}


\centering

\renewcommand{\arraystretch}{1.2}

\resizebox{\textwidth}{!}{

\begin{tabular}{l|rrrrrrrrrr|rrrr}
\toprule
 & \fontsize{8}{9}\selectfont\textbf{Imbalanced} & \fontsize{8}{9}\selectfont\textbf{Random} & \fontsize{8}{9}\selectfont\textbf{Global} & \fontsize{8}{9}\selectfont\textbf{SMOTE} & \fontsize{8}{9}\selectfont\textbf{Border.} & \fontsize{8}{9}\selectfont\textbf{Safelevel} & \fontsize{8}{9}\selectfont\textbf{ADASYN} & \fontsize{8}{9}\selectfont\textbf{MWMOTE} & \fontsize{8}{9}\selectfont\textbf{DBSMOTE} & \fontsize{8}{9}\selectfont\textbf{LVQ} & \fontsize{8}{9}\selectfont\textbf{Simplicial} & \fontsize{8}{9}\selectfont\textbf{S-Border.} & \fontsize{8}{9}\selectfont\textbf{S-Safe.} & \fontsize{8}{9}\selectfont\textbf{S-ADASYN} \\  \midrule

ecoli           &     0.5628 &     0.5735 &     0.6048 &     0.5965 &     0.5696 &     0.5718 &     0.5875 &     0.6024 &     0.5893 &     0.5767 &     \underline{0.6282} &     \underline{0.6003} &     \underline{0.5839} &     \underline{0.6230} \\
optical\_digits &     0.5586 &     0.6698 &     0.7381 &     0.7193 &     0.6779 &     0.7178 &     0.6824 &     0.7269 &     0.6717 &     0.6627 &     \underline{0.7551} &     \underline{0.6876} &     \underline{0.7624} &     \underline{0.7339} \\
pen\_digits     &     0.6719 &     0.8017 &     0.6951 &     0.8110 &     \underline{0.7006} &     0.8118 &     0.6857 &     0.7270 &     0.8044 &     0.8005 &     \underline{0.8290} &     0.6916 &     \underline{0.8278} &     \underline{0.7268} \\
abalone         &     0.0000 &     0.3700 &     0.3983 &     0.3769 &     0.3792 &     0.3799 &     0.3716 &     0.3879 &     0.3785 &     0.3747 &     \underline{0.3883} &     \underline{0.3950} &     \underline{0.3865} &     \underline{0.3847} \\
sick\_euthyroid &     0.8494 &     0.8243 &     0.8214 &     0.8288 &     0.8247 &     0.7334 &     0.8273 &     0.8297 &     0.8397 &     0.8109 &     \underline{0.8382} &     \underline{0.8321} &     \underline{0.8401} &     \underline{0.8310} \\
spectrometer    &     0.6129 &     0.7237 &     0.6315 &     0.7186 &     0.7453 &     \underline{0.7697} &     0.7025 &     0.6878 &     0.7828 &     0.6183 &     \underline{0.8068} &     \underline{0.7456} &     0.7426 &     \underline{0.7792} \\
car\_eval\_34   &     0.2588 &     0.6426 &     0.7485 &     0.7058 &     0.7120 &     0.6743 &     \underline{0.7187} &     0.6990 &     0.6429 &     0.5837 &     \underline{0.7278} &     \underline{0.7131} &     \underline{0.7278} &     0.7019 \\
us\_crime       &     0.4243 &     0.4639 &     0.4692 &     0.4702 &     0.4787 &     \underline{0.4753} &     \underline{0.4623} &     0.4557 &     0.4652 &     0.4455 &     \underline{0.4723} &     \underline{0.4814} &     0.4560 &     0.4575 \\
yeast\_ml8      &     0.0000 &     0.1320 &     0.1560 &     0.1484 &     \underline{0.1502} &     \underline{0.1565} &     0.1445 &     0.1423 &     0.1386 &     0.1271 &     \underline{0.1527} &     0.1477 &     0.1538 &     \underline{0.1451} \\
scene           &     0.0109 &     0.2549 &     0.2528 &     \underline{0.2617} &     \underline{0.2578} &     \underline{0.2543} &     \underline{0.2552} &     0.2580 &     0.0705 &     0.0477 &     0.2352 &     0.2490 &     0.2368 &     0.2408 \\
libras\_move    &     0.4906 &     0.6951 &     0.6548 &     0.6638 &     0.6678 &     0.6398 &     0.6333 &     0.6510 &     0.6802 &     0.6834 &     \underline{0.7003} &     \underline{0.6769} &     \underline{0.6512} &     \underline{0.6878} \\
thyroid\_sick   &     0.8334 &     0.7835 &     0.7323 &     \underline{0.7920} &     \underline{0.7857} &     0.6269 &     0.7846 &     0.7381 &     0.8075 &     0.7323 &     0.7916 &     0.7840 &     \underline{0.7845} &     \underline{0.7854} \\
coil\_2000      &     0.0000 &     0.2120 &     0.2248 &     \underline{0.2184} &     \underline{0.2166} &     0.2074 &     \underline{0.2150} &     0.2199 &     0.0811 &     0.0101 &     0.2092 &     0.2165 &     \underline{0.2095} &     0.2092 \\
solar\_flare\_m0&     0.0164 &     0.1959 &     0.2459 &     \underline{0.1828} &     \underline{0.1918} &     \underline{0.1917} &     \underline{0.1754} &     0.1923 &     0.0659 &     0.1572 &     0.1712 &     0.1807 &     0.1701 &     0.1708 \\
oil             &     0.3640 &     0.3905 &     0.3752 &     0.3659 &     0.3993 &     0.3904 &     0.3585 &     0.3487 &     0.3782 &     0.3906 &     \underline{0.4600} &     \underline{0.4522} &     \underline{0.4064} &     \underline{0.4418} \\
car\_eval\_4    &     0.0000 &     0.4061 &     0.5011 &     0.4387 &     0.4383 &     0.4290 &     0.4213 &     0.4280 &     0.4034 &     0.5403 &     \underline{0.4696} &     \underline{0.4750} &     \underline{0.4696} &     \underline{0.4403} \\
wine\_quality   &     0.0764 &     0.2317 &     0.1821 &     \underline{0.2091} &     \underline{0.2246} &     \underline{0.2191} &     \underline{0.1949} &     0.2006 &     0.1765 &     0.2753 &     0.2015 &     0.2241 &     0.2104 &     0.1842 \\
letter\_img     &     0.6064 &     0.4611 &     0.5567 &     0.5507 &     0.4252 &     0.5268 &     0.4499 &     0.4832 &     0.5624 &     0.5206 &     \underline{0.6195} &     \underline{0.4331} &     \underline{0.6236} &     \underline{0.5385} \\
yeast\_me2      &     0.0972 &     0.2700 &     0.2836 &     0.2768 &     0.3272 &     \underline{0.2999} &     0.2610 &     0.2935 &     0.2727 &     0.3121 &     \underline{0.3071} &     \underline{0.3366} &     0.2858 &     \underline{0.2930} \\
ozone\_level    &     0.0528 &     0.2384 &     0.2198 &     0.2354 &     0.2633 &     0.2240 &     0.2280 &     0.2402 &     0.2393 &     0.2395 &     \underline{0.2846} &     \underline{0.2823} &     \underline{0.2559} &     \underline{0.2775} \\
abalone\_19     &     0.0000 &     0.0471 &     0.0367 &     0.0439 &     \underline{0.0565} &     0.0448 &     0.0448 &     0.0591 &     0.0415 &     0.0390 &     \underline{0.0442} &     0.0522 &     \underline{0.0499} &     \underline{0.0460} \\ \midrule
\textbf{mean}   &     0.3089 &     0.4470 &     0.4537 &     0.4578 &     0.4520 &     0.4450 &     0.4383 &     0.4462 &     0.4330 &     0.4261 &     \textcolor{red}{\textbf{\underline{0.4806}}} &     \underline{0.4599} &     \textcolor{blue}{\textbf{\underline{0.4683}}} &     \underline{0.4618} \\
\textbf{rank}   &    12.1429 &     8.4286 &     6.9524 &     6.8095 &     6.2381 &     7.8095 &     9.4286 &     7.3810 &     8.5238 &     9.4762 &     \textcolor{red}{\textbf{\underline{4.1429}}} &     \textcolor{blue}{\textbf{\underline{5.3810}}} &     \underline{5.8095} &     \underline{6.4762} \\ \bottomrule

\end{tabular}

}


\end{table*}


In our work, we classified the existing approaches to geometric data modeling and sampling based on the neighborhood relation size and arity, highlighting their connection with the issues of low data coverage and low sample quality. We proposed a new instance of geometric oversampling called Simplicial SMOTE. As the original SMOTE algorithm, it models data locally by the neighborhood size $k$ much less than the amount $n$ of data points. Yet, instead of a graph model of data, which samples synthetic points as random convex combinations from the neighborhood graph edges, it uses a simplicial complex to model the data to sample synthetic points as random convex combinations from its simplices, formed by points being in $p$-ary neighborhood relation. This results in better coverage of the true data distribution and allows the generation of synthetic points of the minor class closer to the major class on the decision boundary, effectively moving the decision boundary away from the minor class.

 We have shown on model and real imbalanced datasets that the proposed approach to data modeling and sampling performs better than several sampling methods, including global sampling, original SMOTE, and several of its popular variants, to solve the classification problem in the presence of data imbalance. Moreover, the mean projection distance to the geometric model of the minority class gets smaller with increasing maximal relation arity parameter $p$, effectively allowing the move of the local decision boundary to the major class (Fig. \ref{figure:decision_boundary}). 

In our experiments, we have concluded that choosing $p$ -- the best number of points to span a simplex generally follows the similar tradeoff a choosing the neighborhood size $k$, with optimal value of $p$ is often neither too small nor too large (Appendix \ref{appendix_sensitivity}). The synthetic points, which are a convex combination of a large number of existing data points, could be potentially either too similar for a small neighborhood size or oversmoothed for the large one. Thus, we recommend doing a grid search over the maximal simplex dimension $p$.

Our method improves the original SMOTE algorithm only in terms of sampling, yet it is orthogonal and compatible with one of the most popular SMOTE variants. We demonstrated how the most cited SMOTE variants, such as Borderline SMOTE, Safe-level SMOTE, and ADASYN, can be generalized to use simplicial sampling. We provided their evaluation, with all simplicial extensions outperforming their original graph-based counterparts.






\begin{acks}
The article was prepared within the framework of the HSE University Basic Research Program.
\end{acks}

\clearpage
\clearpage
\bibliographystyle{ACM-Reference-Format}
\bibliography{main}

\appendix

\section{Datasets summary}
\vspace{-0.5em}

\begin{table}[h!]
\centering
\caption{Benchmark datasets and their properties.} 
\label{table:datasets}

\vspace{-0.75em}

\begin{tabular}{l|rrrrr} \toprule
  &\textbf{Features} &\textbf{Size} &\textbf{Minor} &\textbf{Major} &\textbf{Ratio} \\ \midrule
    ecoli & 7 &336 & 35 &301 &9 \\
    optical\_digits &64 & 5620 &554 & 5066 & 10 \\
    pen\_digits &16 &10992 & 1055 & 9937 & 10 \\
    abalone &10 & 4177 &391 & 3786 & 10 \\
    sick\_euthyroid &42 & 3163 &293 & 2870 & 10 \\
    spectrometer &93 &531 & 45 &486 & 11 \\
    car\_eval\_34 &21 & 1728 &134 & 1594 & 12 \\
    us\_crime & 100 & 1994 &150 & 1844 & 13 \\
    yeast\_ml8 & 103 & 2417 &178 & 2239 & 13 \\
    scene & 294 & 2407 &177 & 2230 & 13 \\
    libras\_move &90 &360 & 24 &336 & 14 \\
    thyroid\_sick &52 & 3772 &231 & 3541 & 16 \\
    coil\_2000 &85 & 9822 &586 & 9236 & 16 \\
    solar\_flare\_m0 &32 & 1389 & 68 & 1321 & 20 \\
    oil &49 &937 & 41 &896 & 22 \\
    car\_eval\_4 &21 & 1728 & 65 & 1663 & 26 \\
    wine\_quality &11 & 4898 &183 & 4715 & 26 \\
    letter\_img &16 &20000 &734 &19266 & 27 \\
    yeast\_me2 & 8 & 1484 & 51 & 1433 & 29 \\
    ozone\_level &72 & 2536 & 73 & 2463 & 34 \\
    abalone\_19 &10 & 4177 & 32 & 4145 &130 \\ \bottomrule
\end{tabular}
\end{table}

\clearpage
\onecolumn
\section{Sensitivity to hyperparameters}
\label{appendix_sensitivity}
\vspace{-0.5em}
\vspace{-0.3em}

\begin{figure}[h!]
\centering

\subfigure[ecoli]{\includegraphics[width=5.6cm,height=2.12cm]{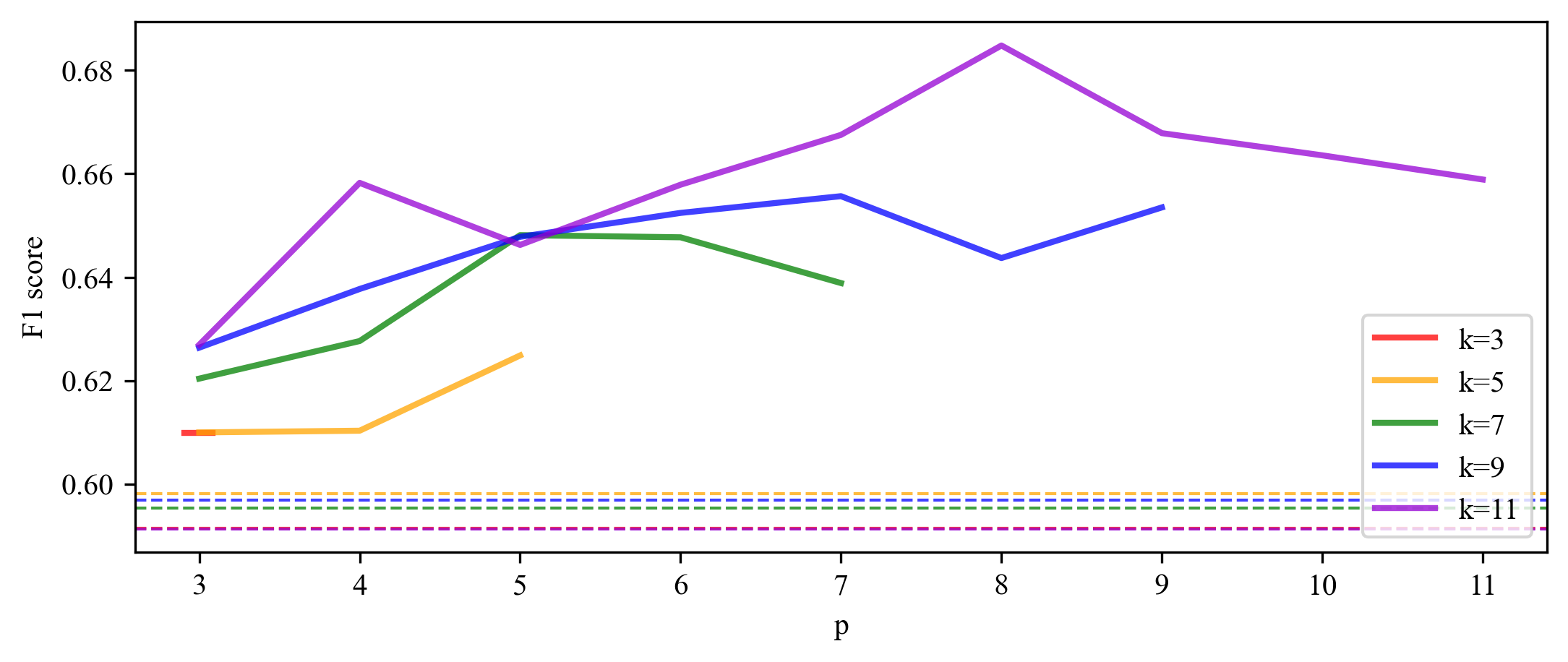}}
\hfill
\subfigure[optical\_digits]{\includegraphics[width=5.6cm,height=2.12cm]{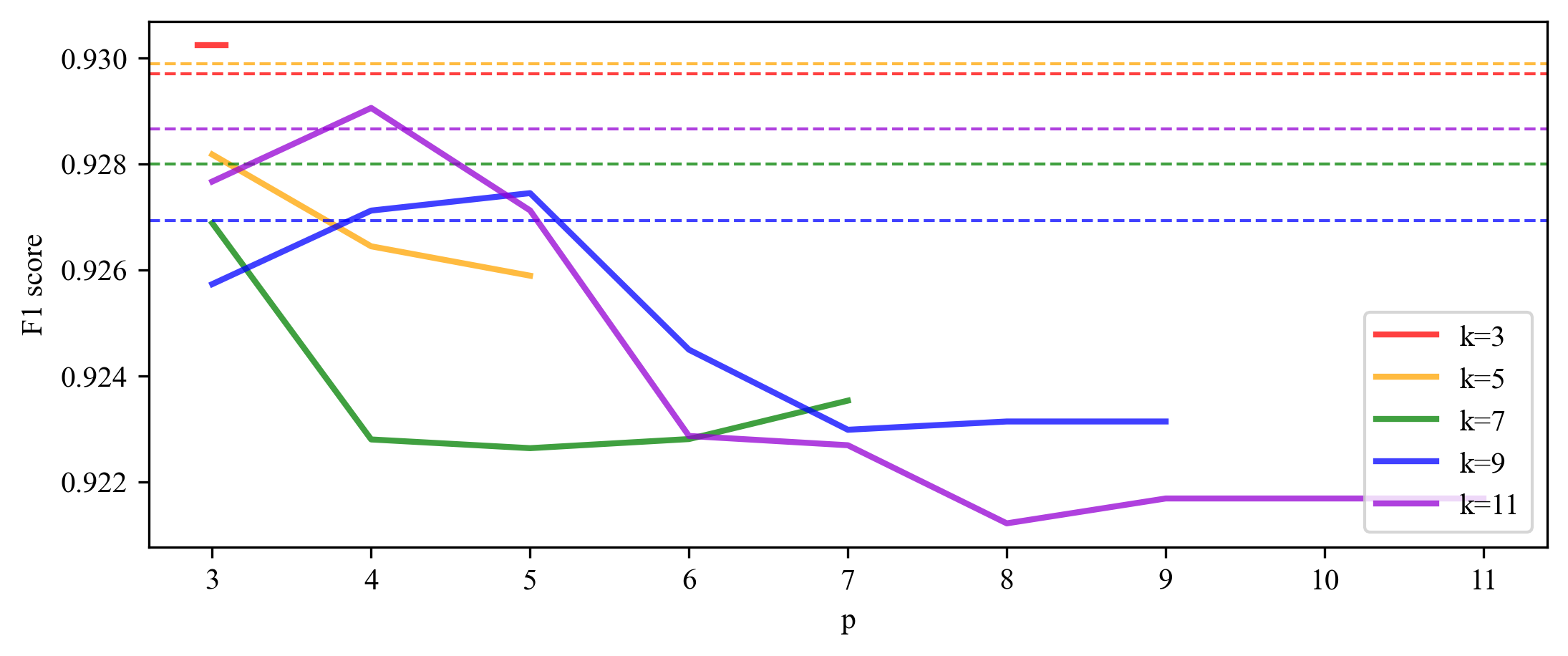}}
\hfill
\subfigure[pen\_digits]{\includegraphics[width=5.6cm,height=2.12cm]{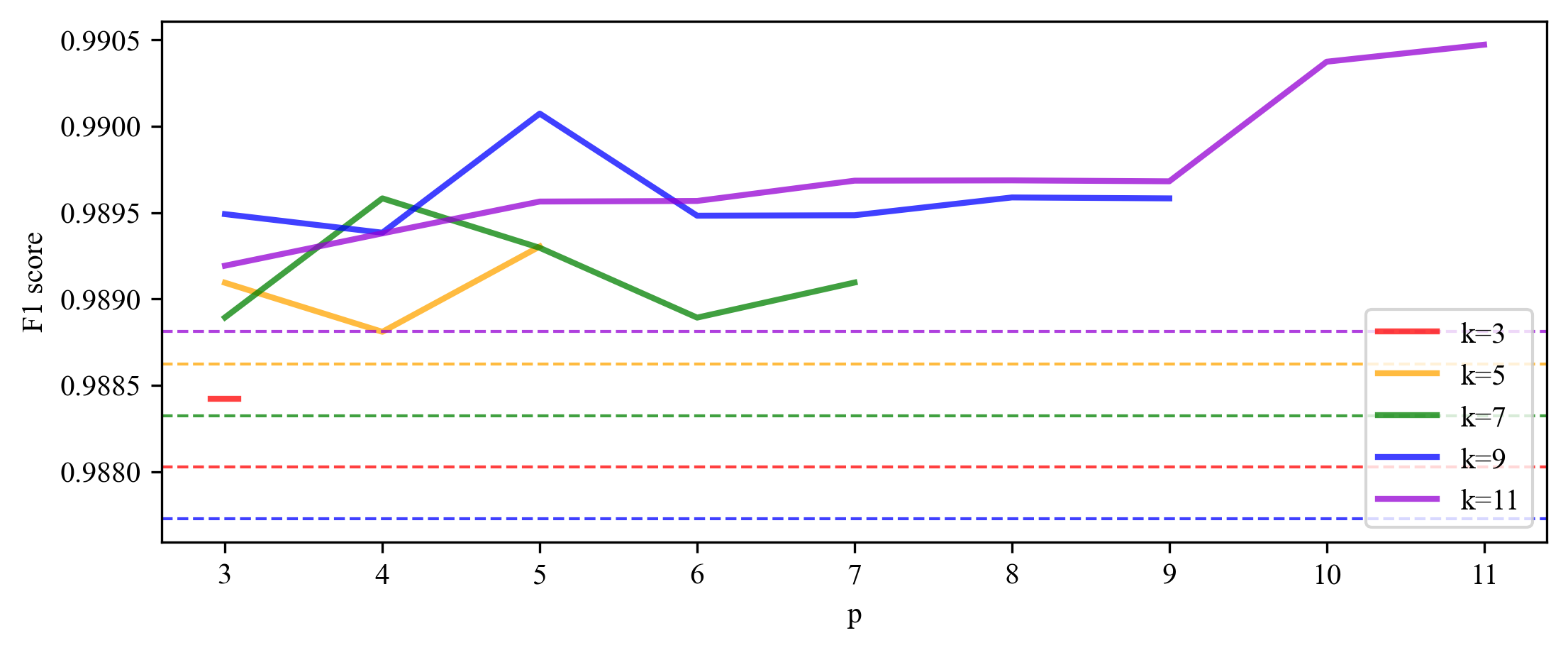}}

\vspace{-0.35em}

\subfigure[abalone]{\includegraphics[width=5.6cm,height=2.12cm]{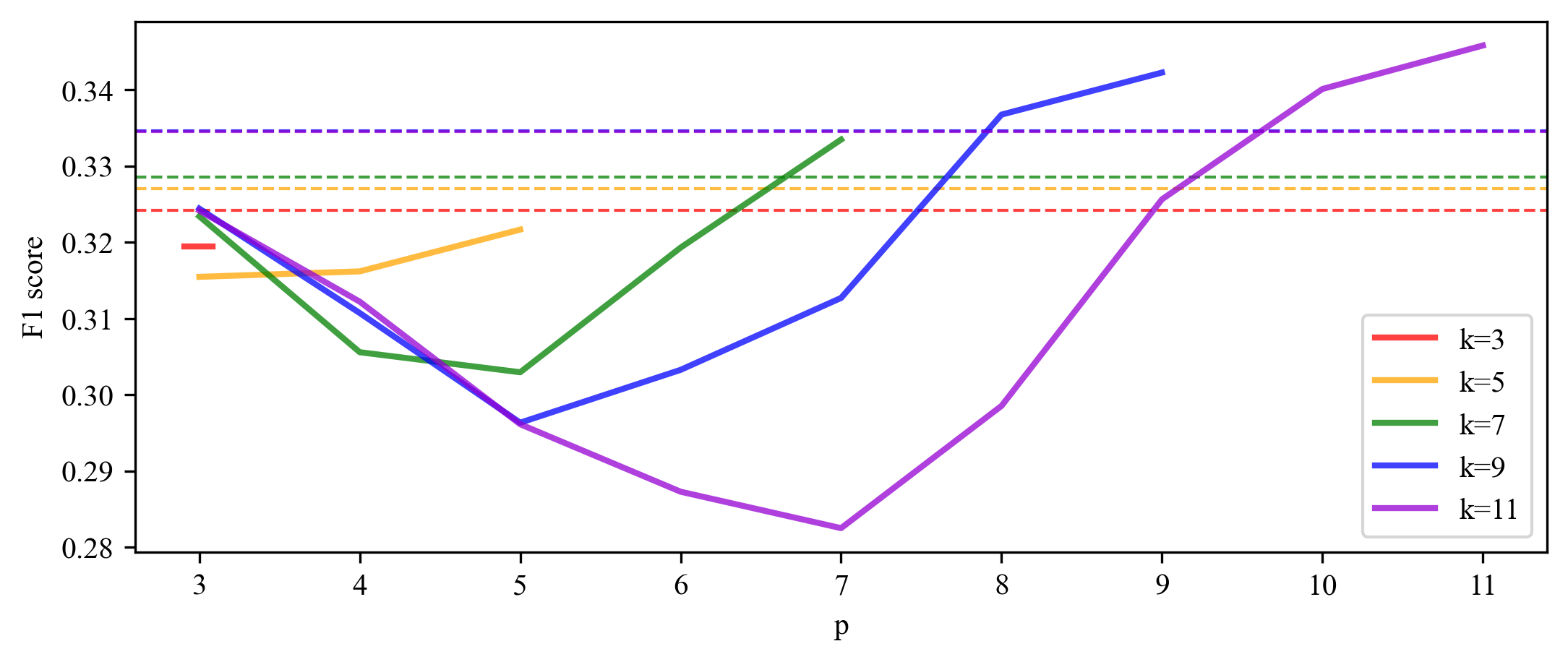}}
\hfill
\subfigure[sick\_euthyroid]{\includegraphics[width=5.6cm,height=2.12cm]{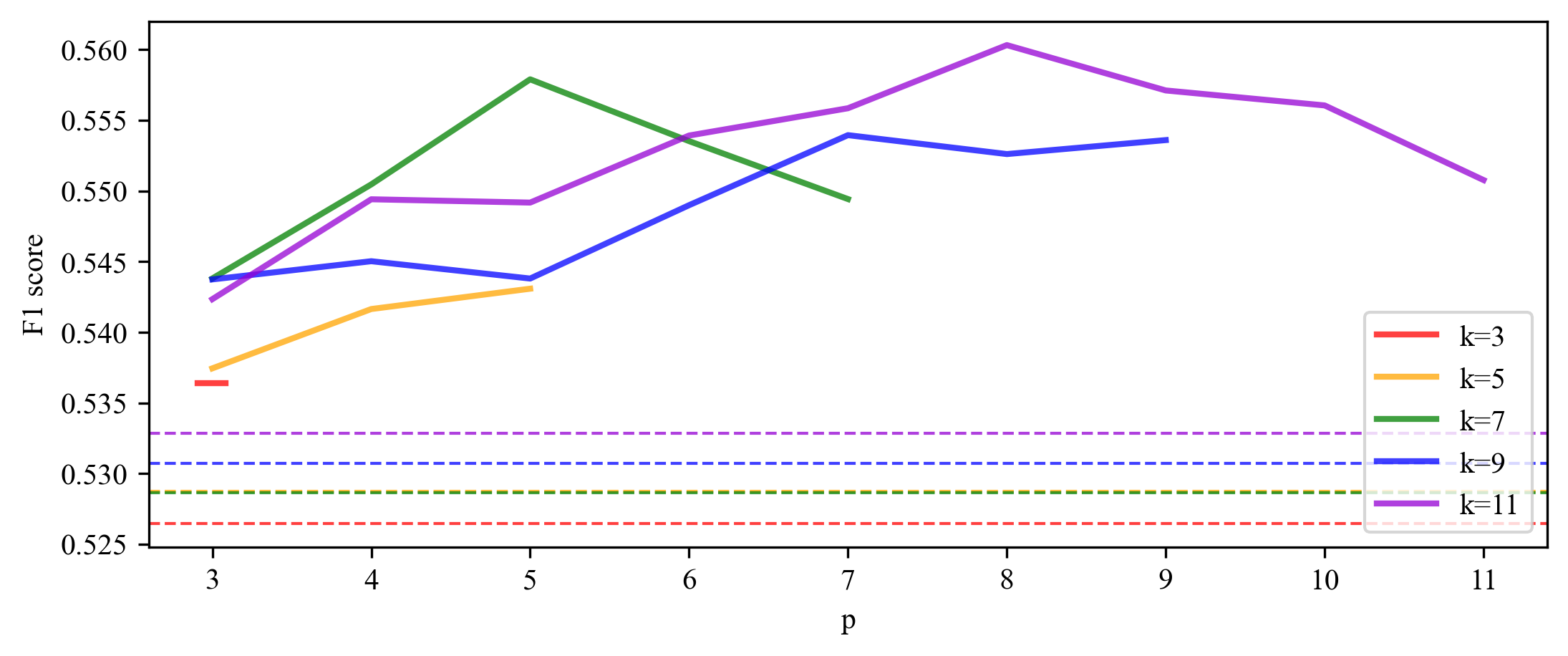}}
\hfill
\subfigure[spectrometer]{\includegraphics[width=5.6cm,height=2.12cm]{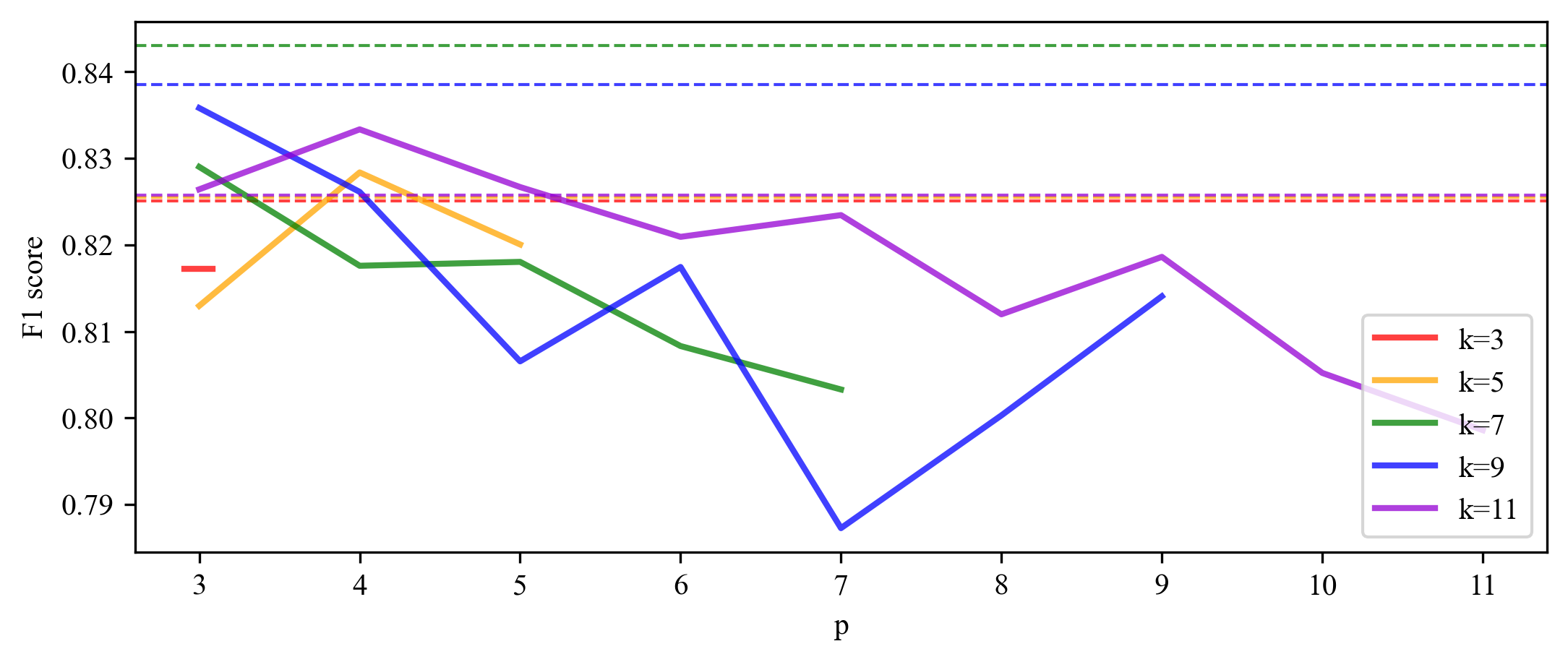}}

\vspace{-0.35em}

\subfigure[car\_eval\_34]{\includegraphics[width=5.6cm,height=2.12cm]{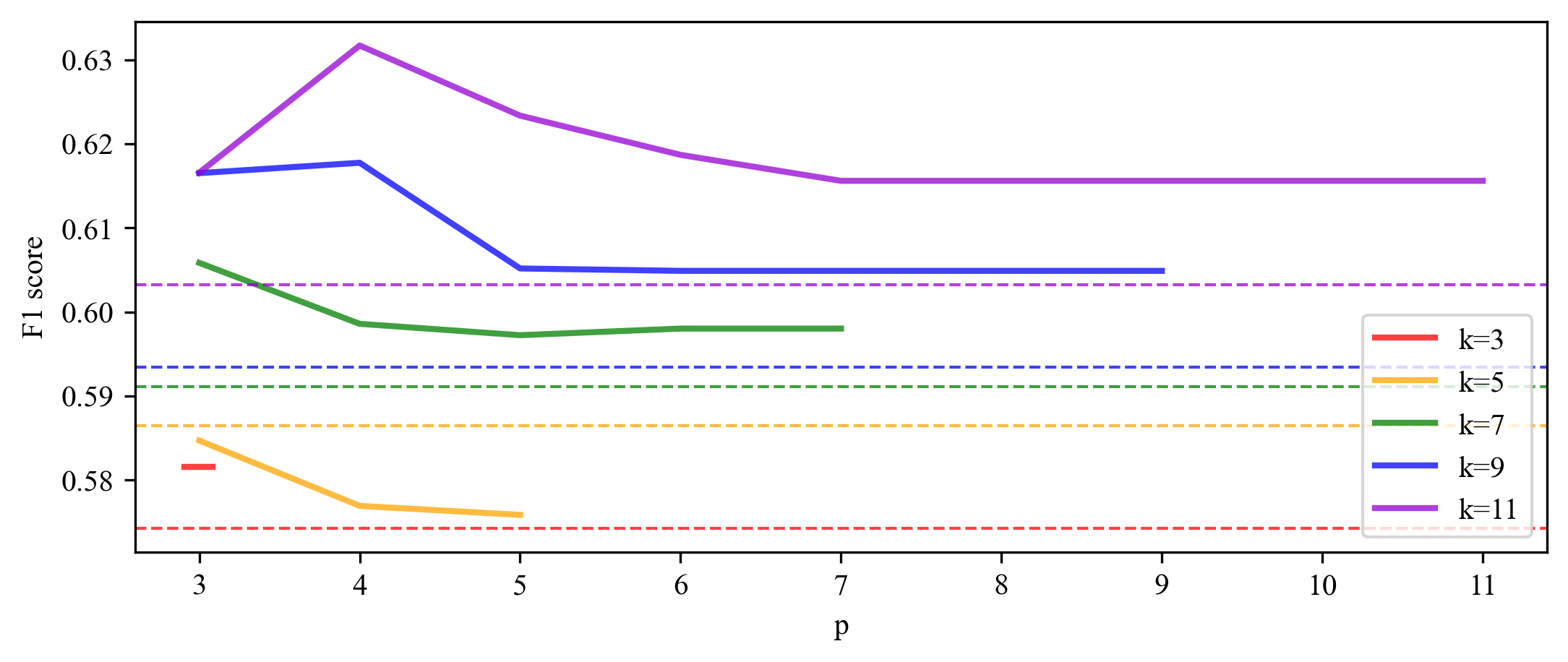}}
\hfill
\subfigure[us\_crime]{\includegraphics[width=5.6cm,height=2.12cm]{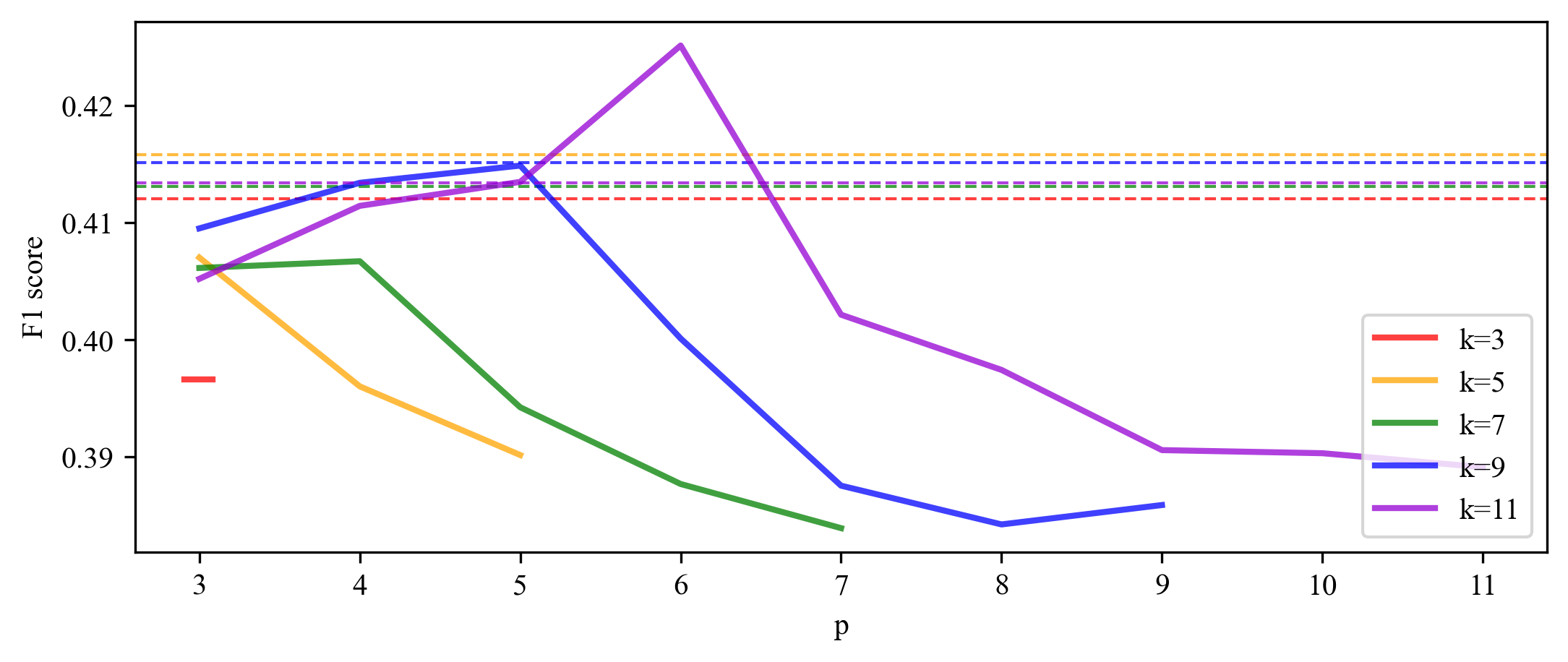}}
\hfill
\subfigure[yeast\_ml8]{\includegraphics[width=5.6cm,height=2.12cm]{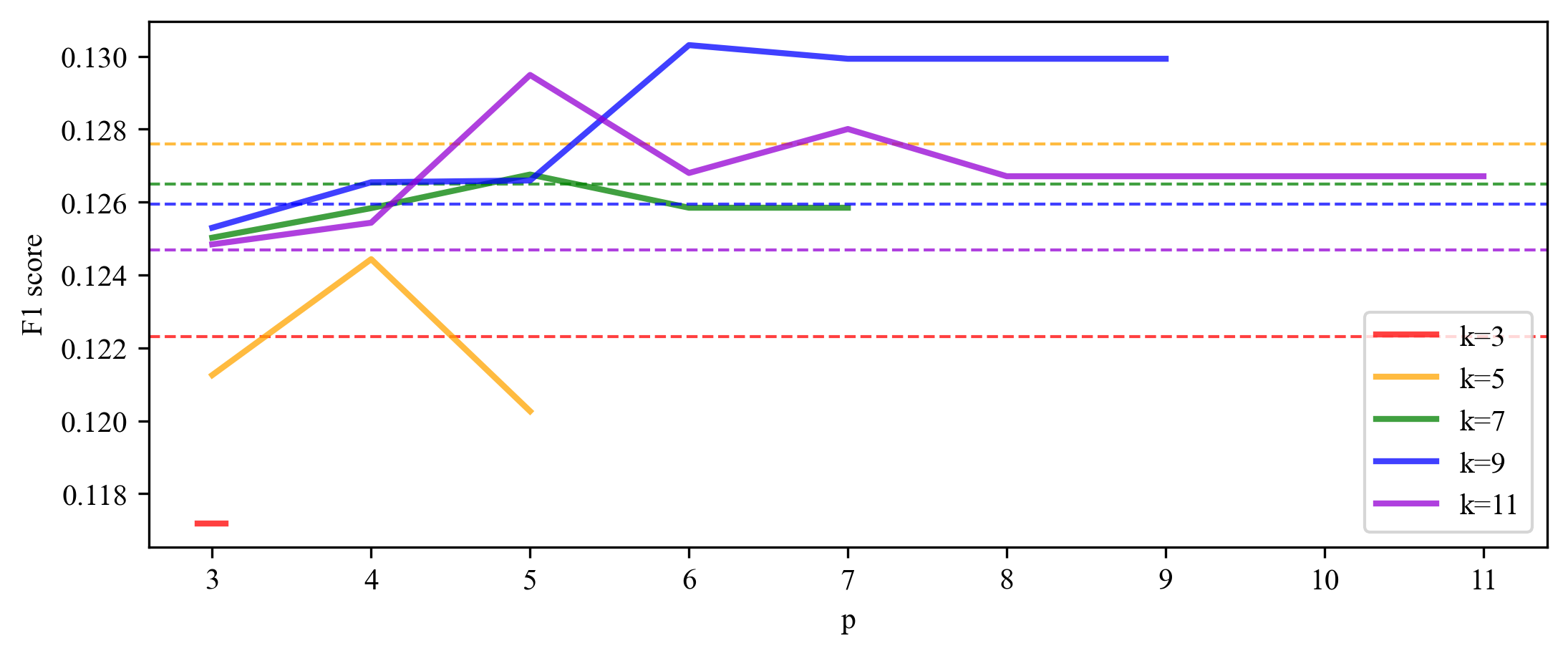}}

\vspace{-0.35em}

\subfigure[scene]{\includegraphics[width=5.6cm,height=2.12cm]{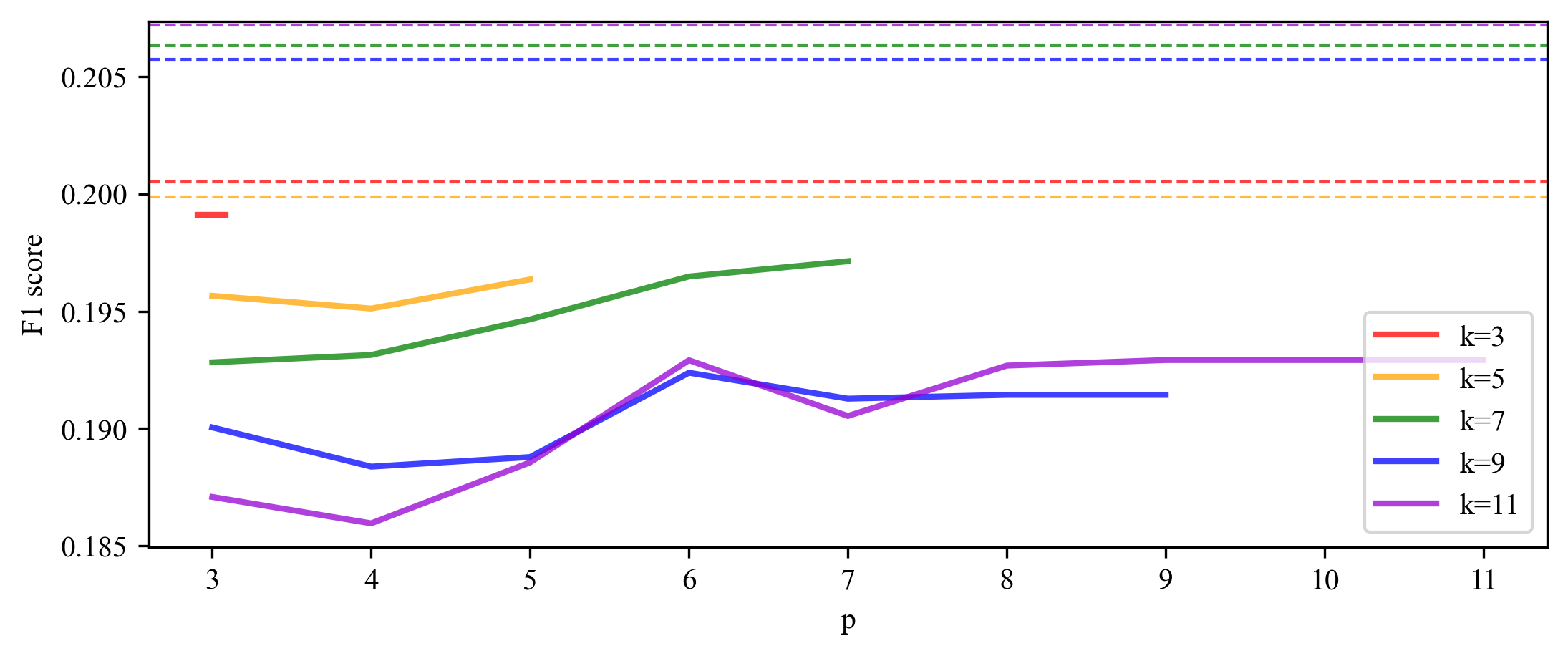}}
\hfill
\subfigure[libras\_move]{\includegraphics[width=5.6cm,height=2.12cm]{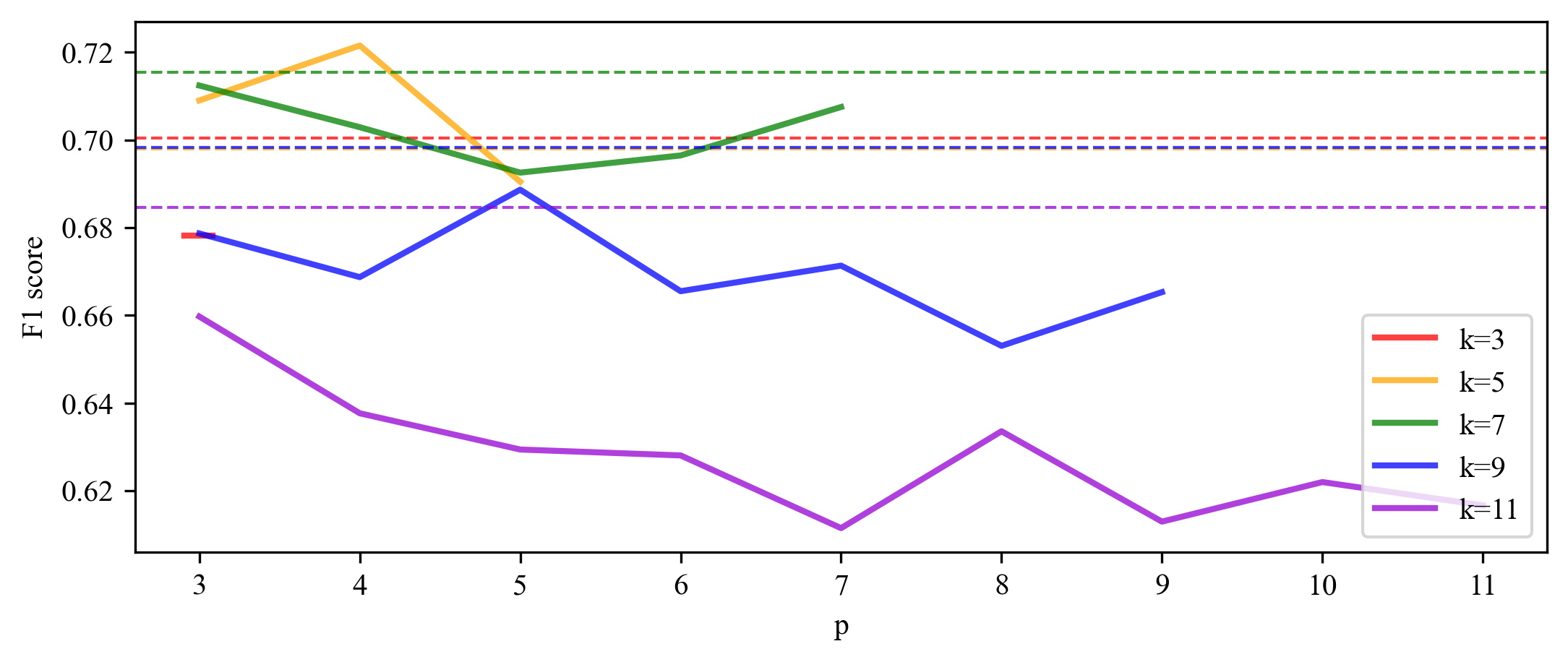}}
\hfill
\subfigure[thyroid\_sick]{\includegraphics[width=5.6cm,height=2.12cm]{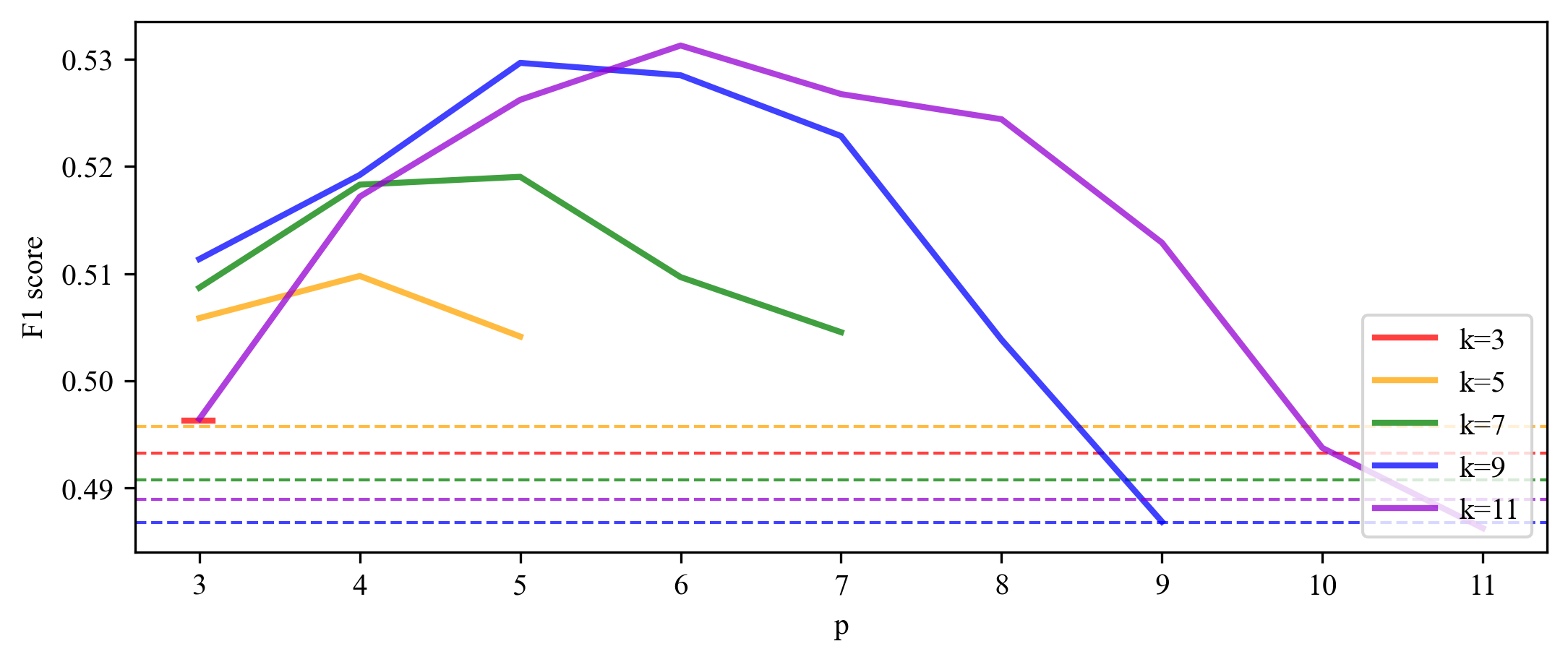}}

\vspace{-0.35em}

\subfigure[coil\_2000]{\includegraphics[width=5.6cm,height=2.12cm]{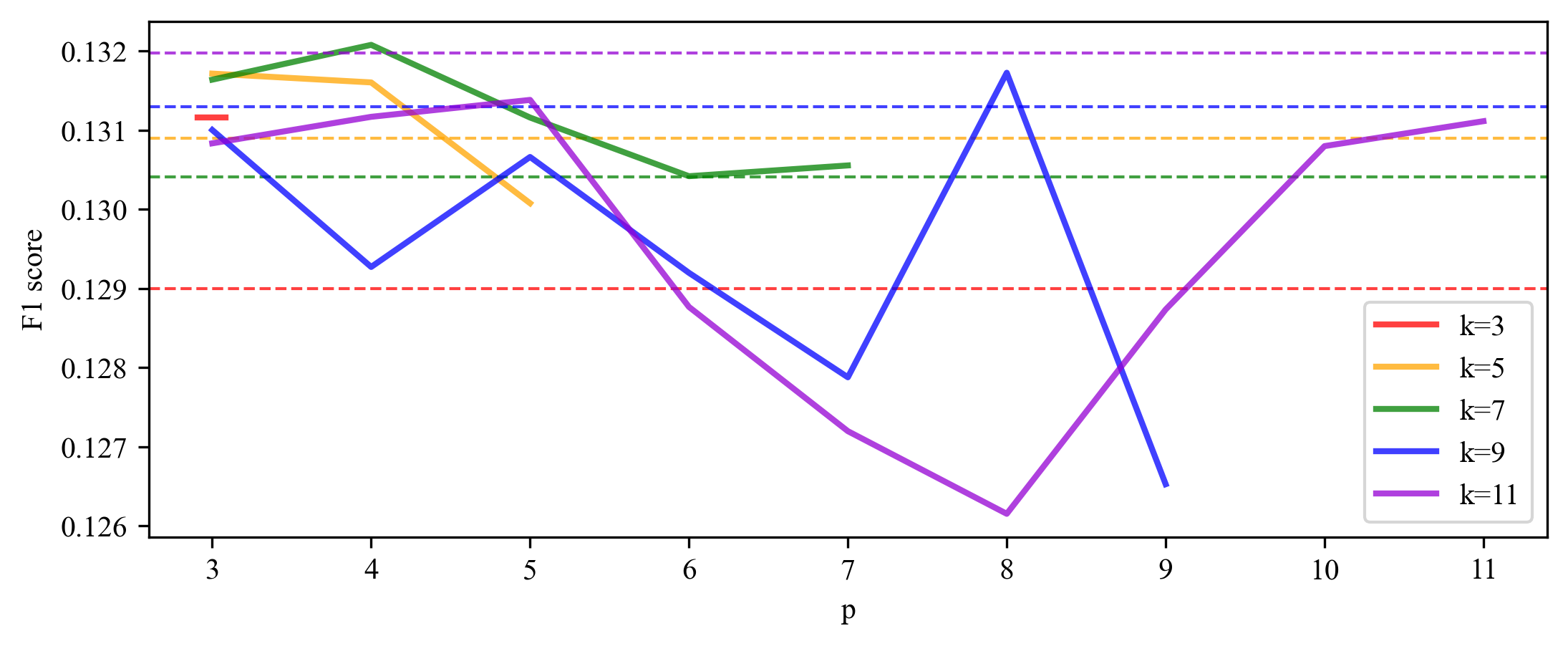}}
\hfill
\subfigure[solar\_flare\_m0]{\includegraphics[width=5.6cm,height=2.12cm]{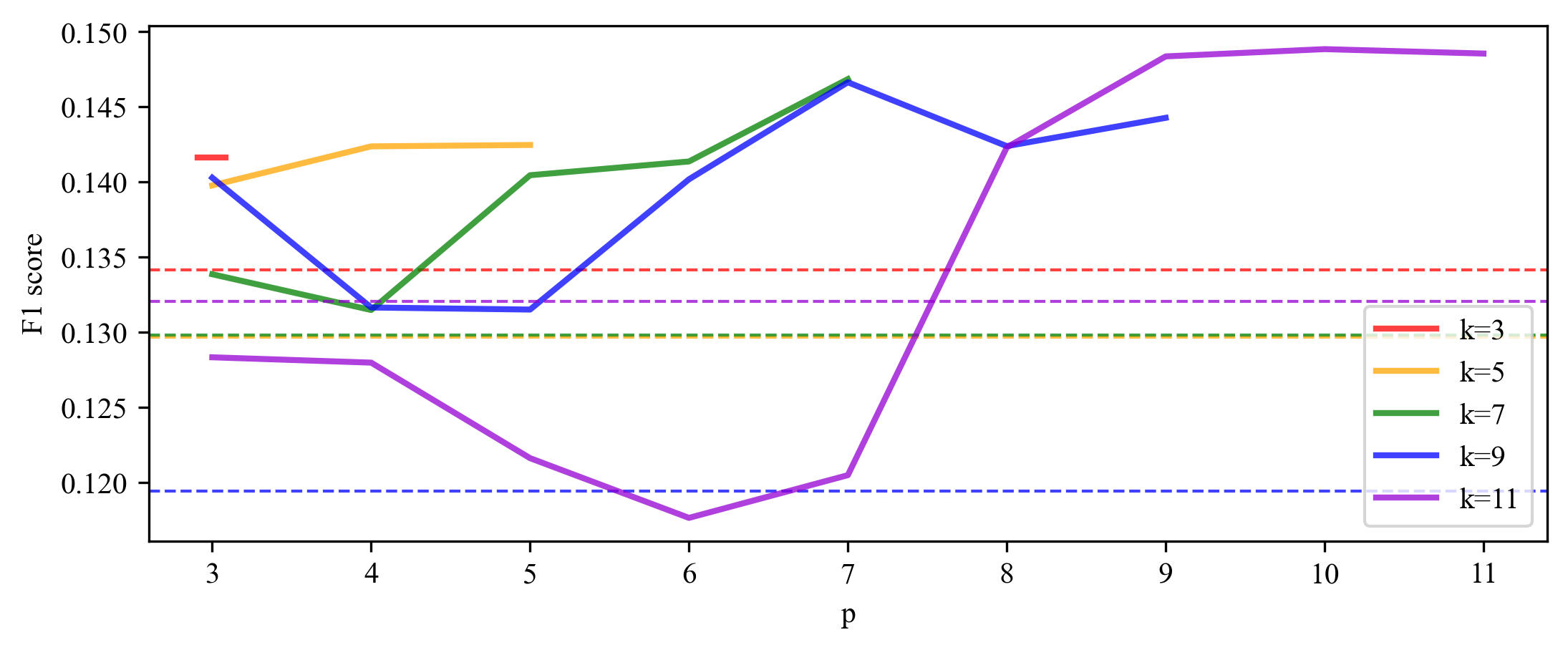}}
\hfill
\subfigure[oil]{\includegraphics[width=5.6cm,height=2.12cm]{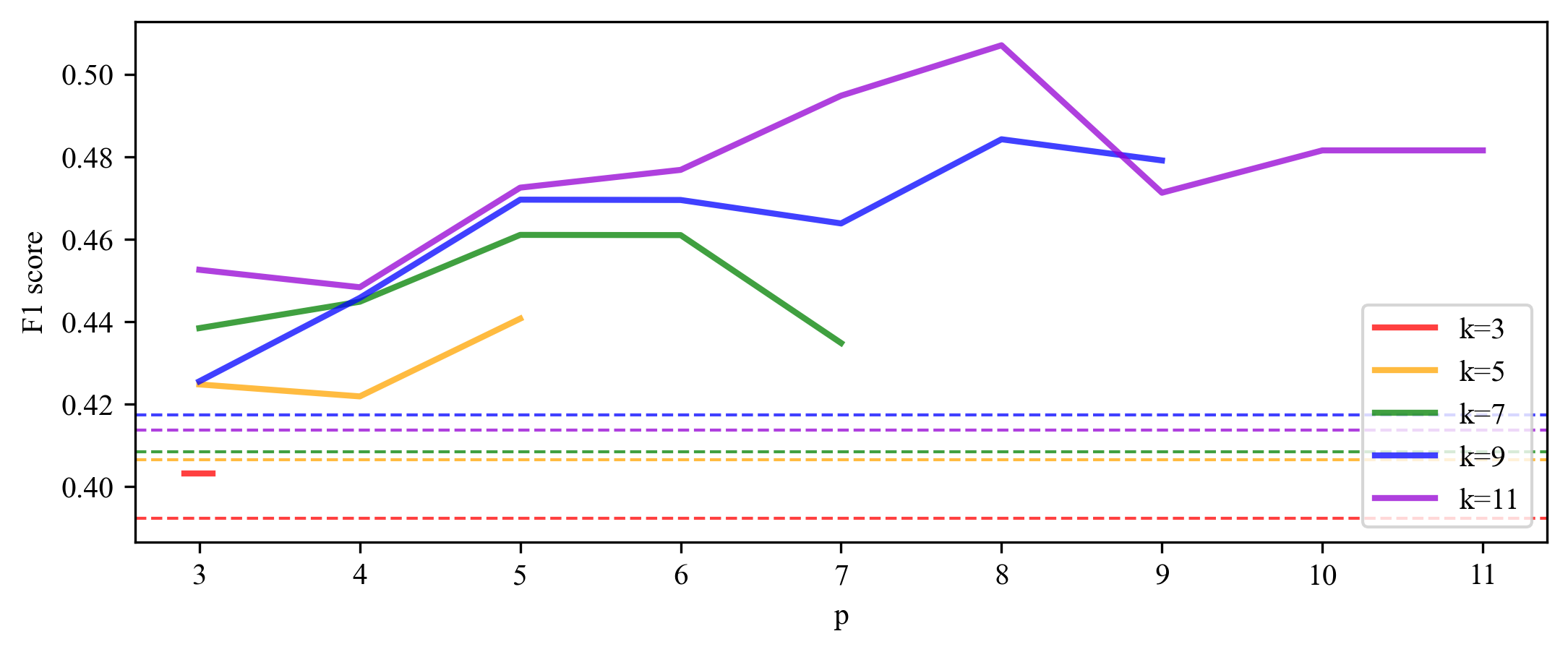}}

\vspace{-0.35em}

\subfigure[car\_eval\_4]{\includegraphics[width=5.6cm,height=2.12cm]{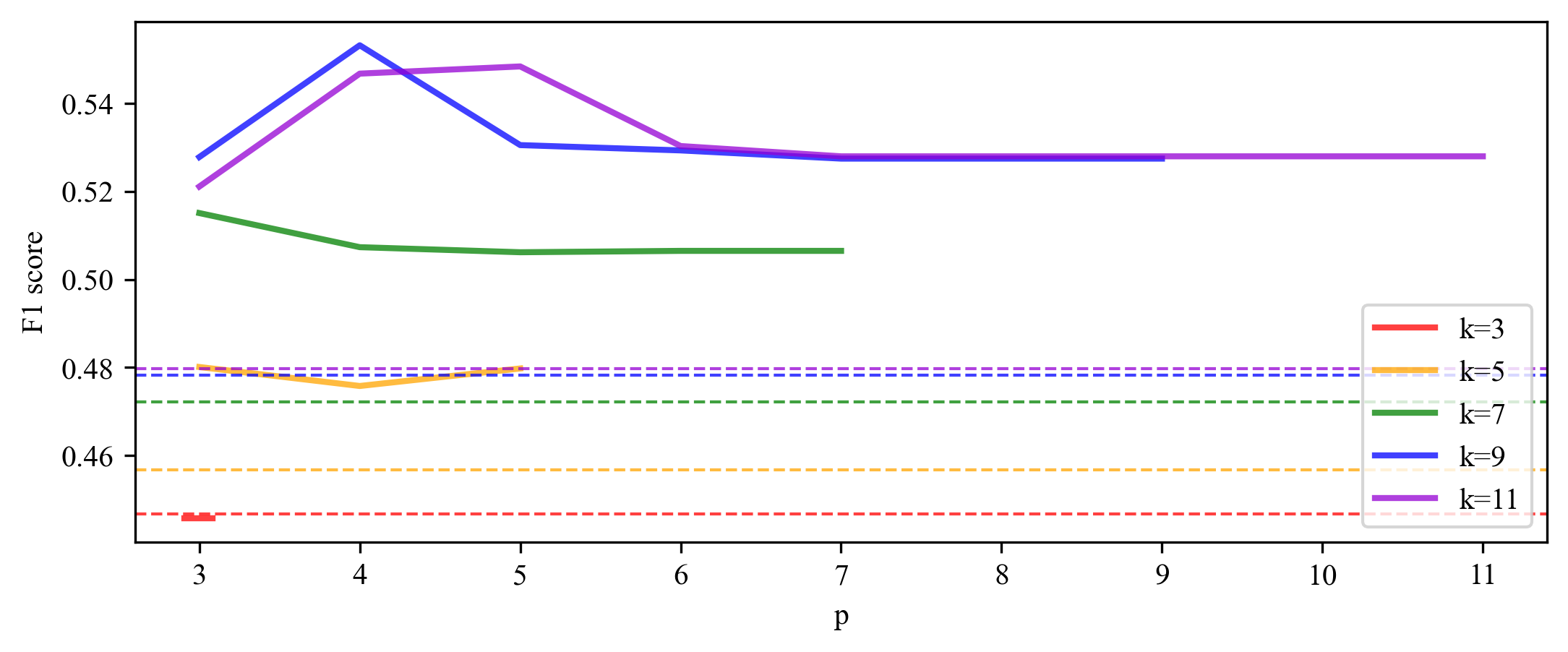}}
\hfill
\subfigure[wine\_quality]{\includegraphics[width=5.6cm,height=2.12cm]{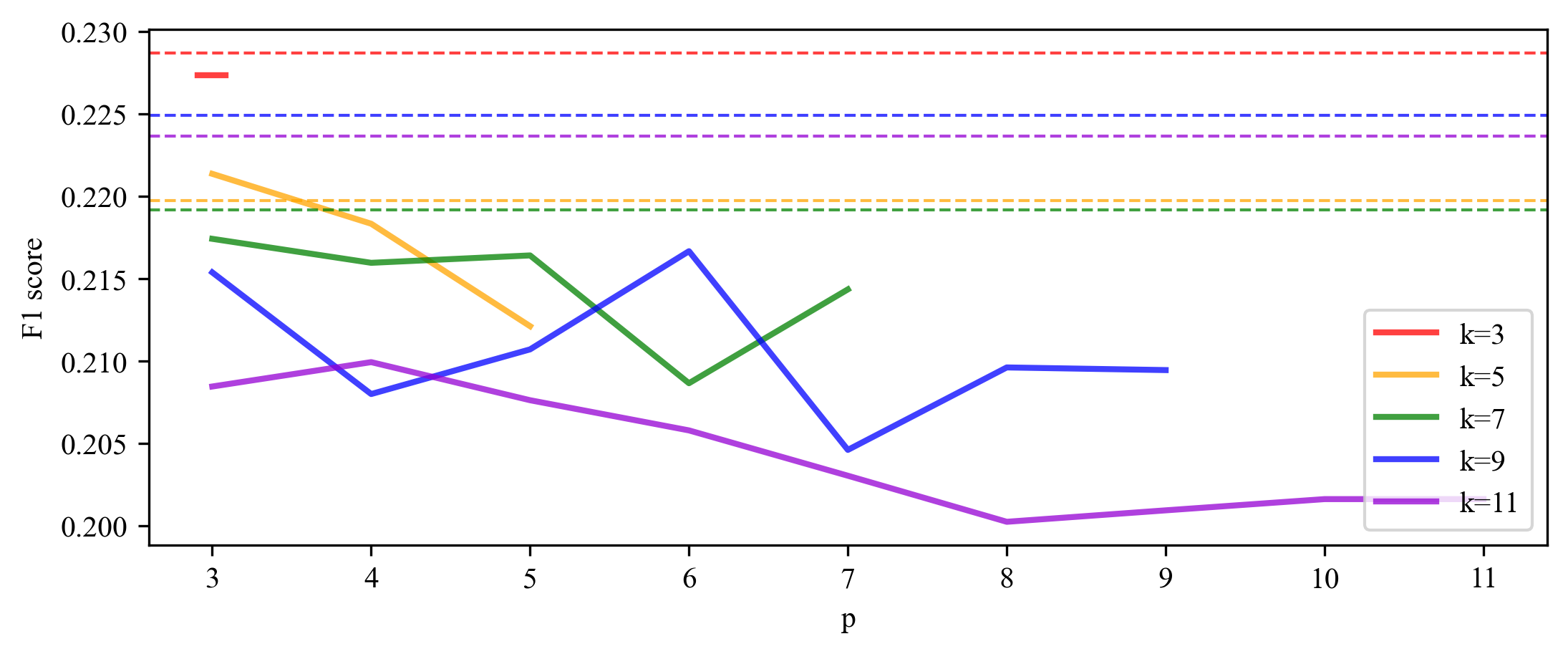}}
\hfill
\subfigure[letter\_img]{\includegraphics[width=5.6cm,height=2.12cm]{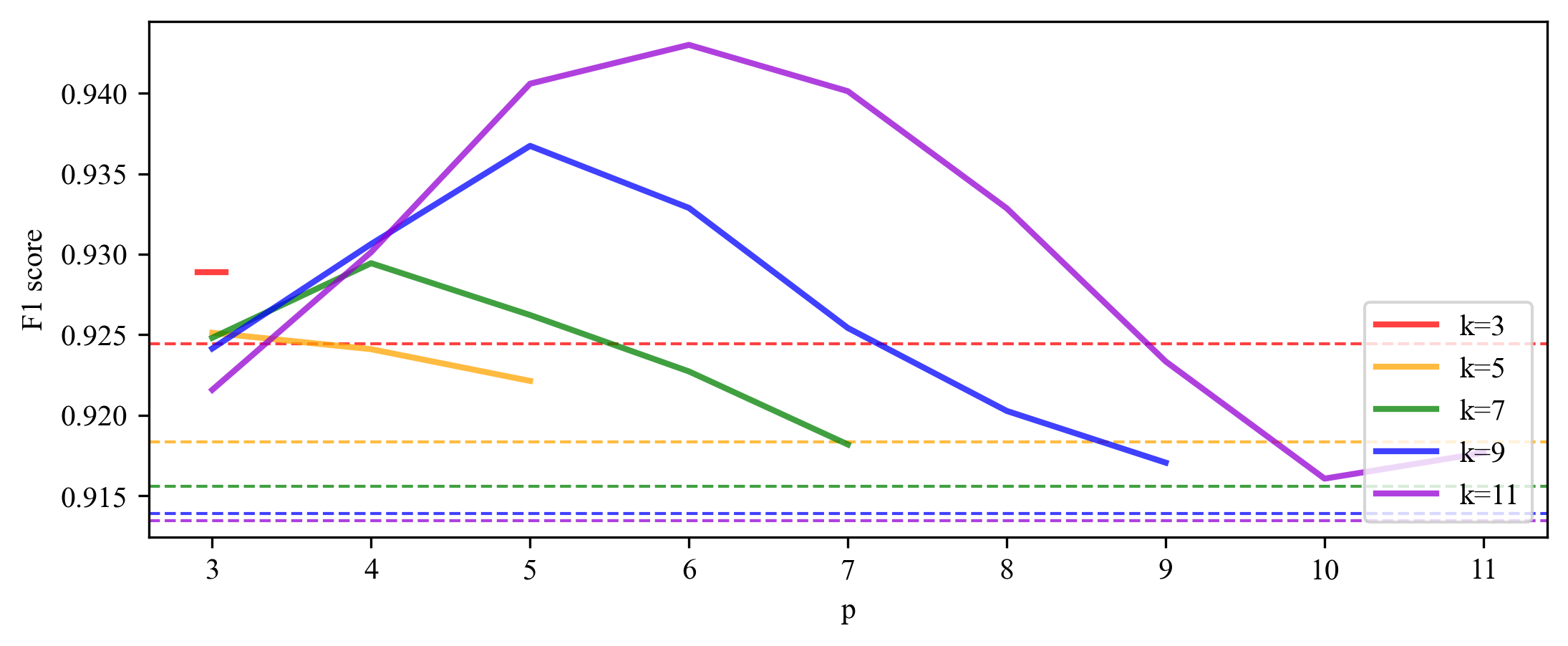}}

\vspace{-0.35em}

\subfigure[yeast\_me2]{\includegraphics[width=5.6cm,height=2.12cm]{figures/parameters/wine_quality_knn.png}}
\hfill
\subfigure[ozone\_level]{\includegraphics[width=5.6cm,height=2.12cm]{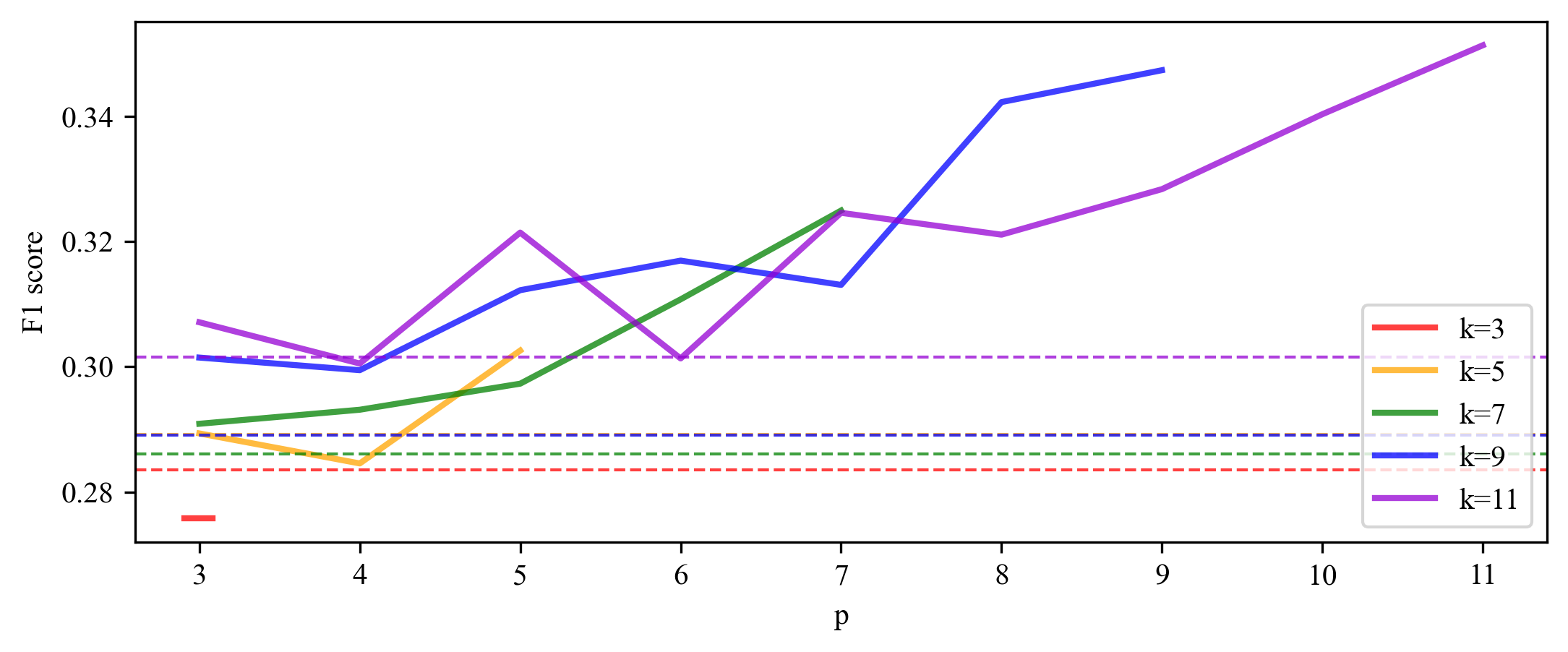}}
\hfill
\subfigure[abalone\_19]{\includegraphics[width=5.6cm,height=2.12cm]{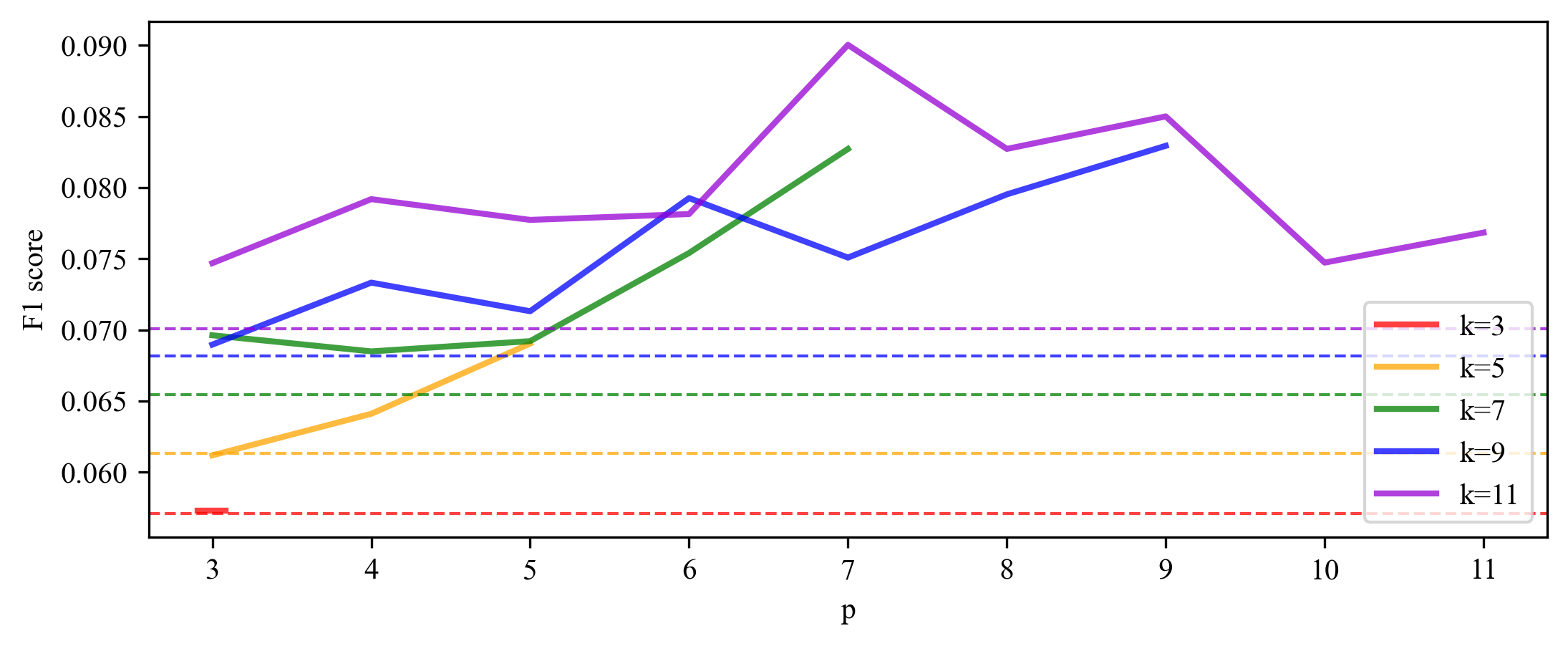}}

\vspace{-0.75em}
\caption{Sensitivity for Simplicial SMOTE's hyperparameters -- neighborhood size $k$ and maximum clique size $p$, followed by the nearest neighbor classifier. Performances in terms of F1 score for various $k$ and $p$ are shown as solid lines. Baseline SMOTE performance for the same $k$ is shown as a dashed line of the same color.}\label{sensitivity_knn}
\Description{Sensitivity of Simplicial SMOTE to hyperparameters}
\end{figure}

\begin{figure}[h!]
\centering

\subfigure[ecoli]{\includegraphics[width=5.6cm,height=2.12cm]{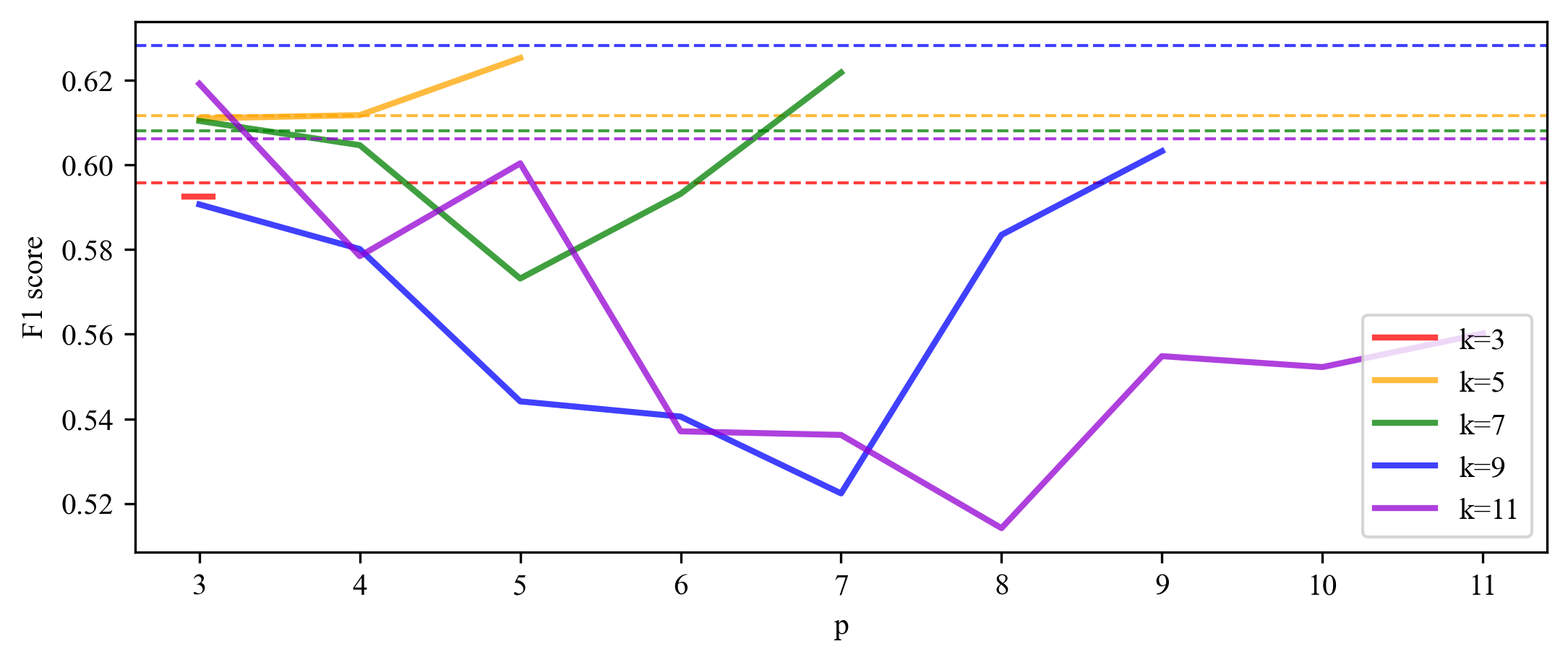}}
\hfill
\subfigure[optical\_digits]{\includegraphics[width=5.6cm,height=2.12cm]{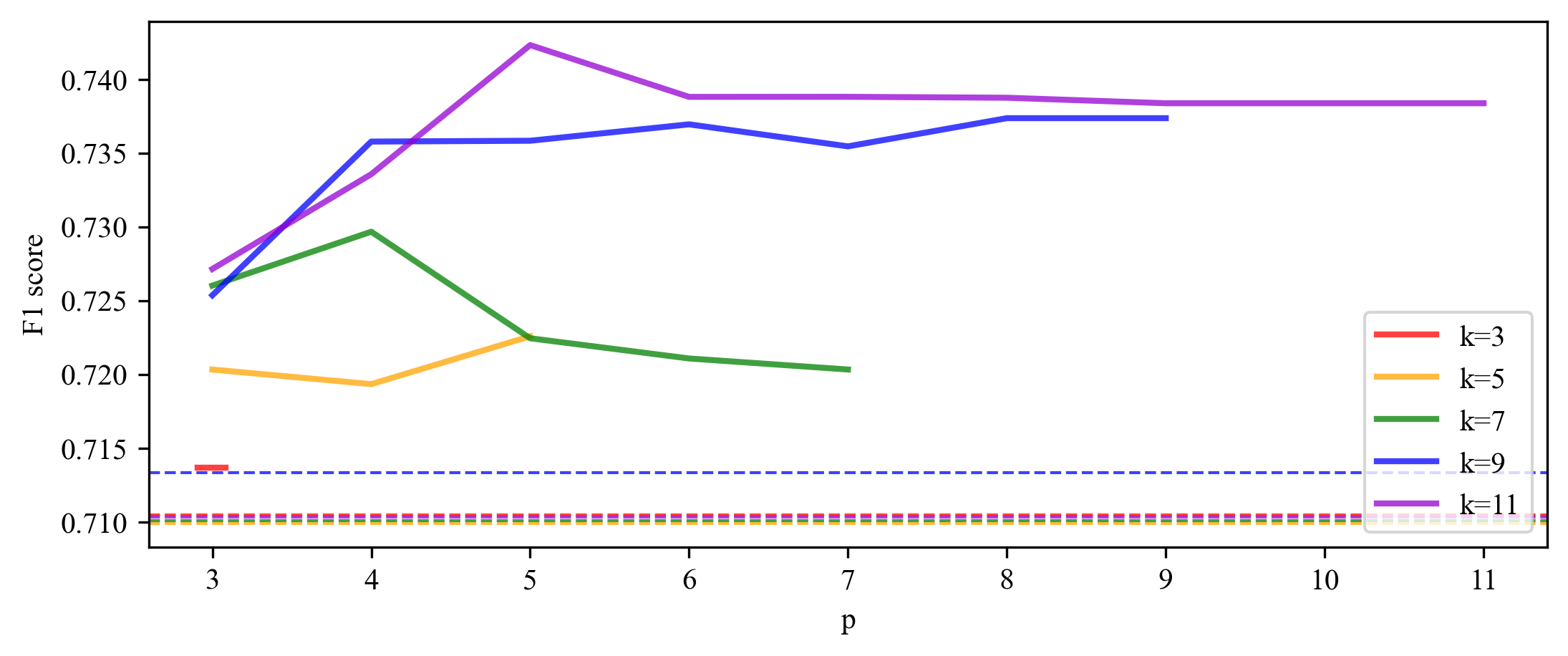}}
\hfill
\subfigure[pen\_digits]{\includegraphics[width=5.6cm,height=2.12cm]{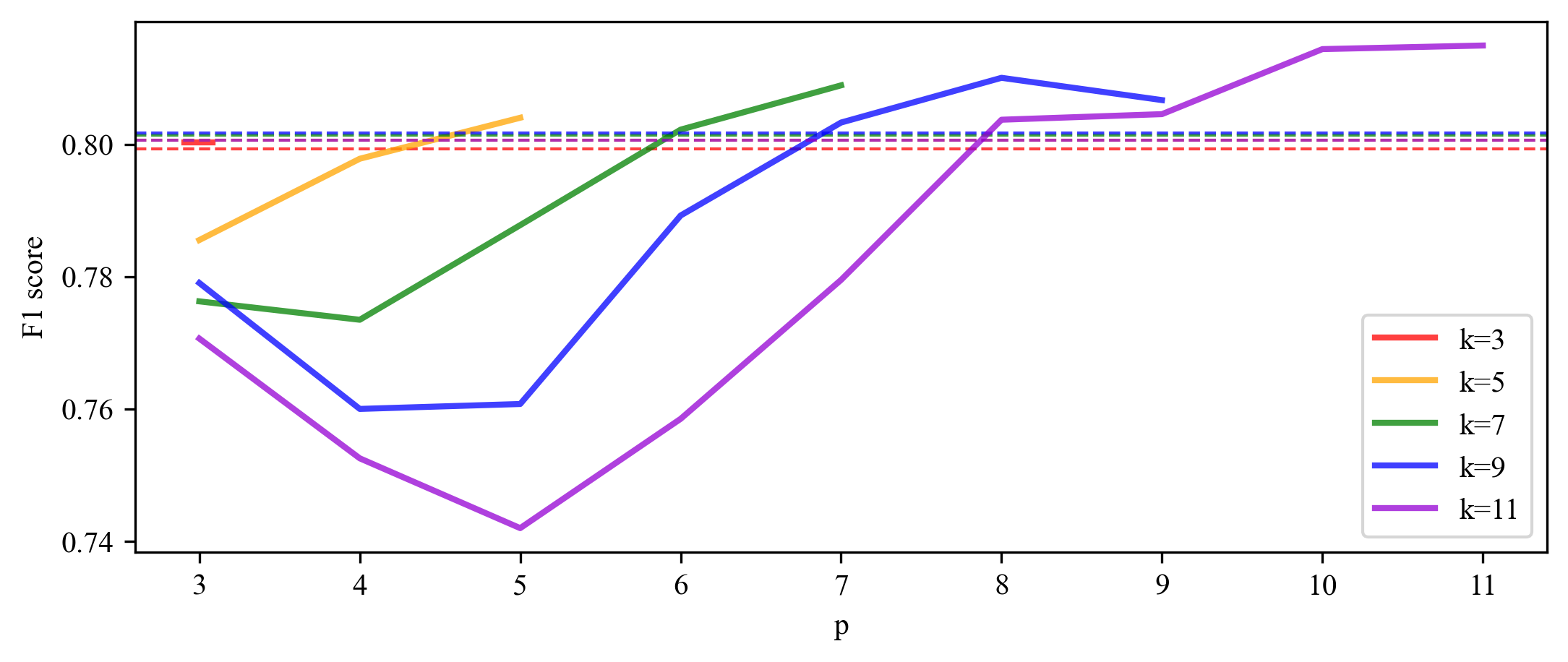}}

\subfigure[abalone]{\includegraphics[width=5.6cm,height=2.12cm]{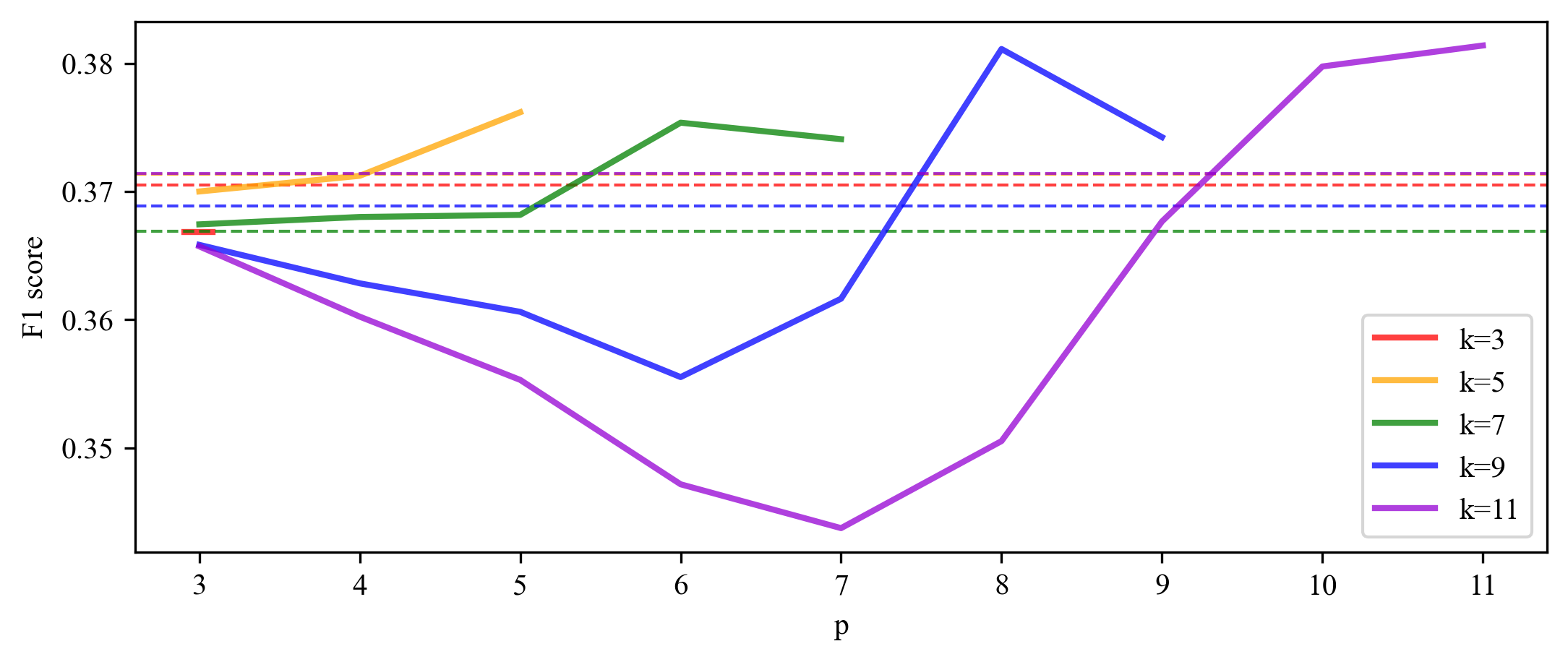}}
\hfill
\subfigure[sick\_euthyroid]{\includegraphics[width=5.6cm,height=2.12cm]{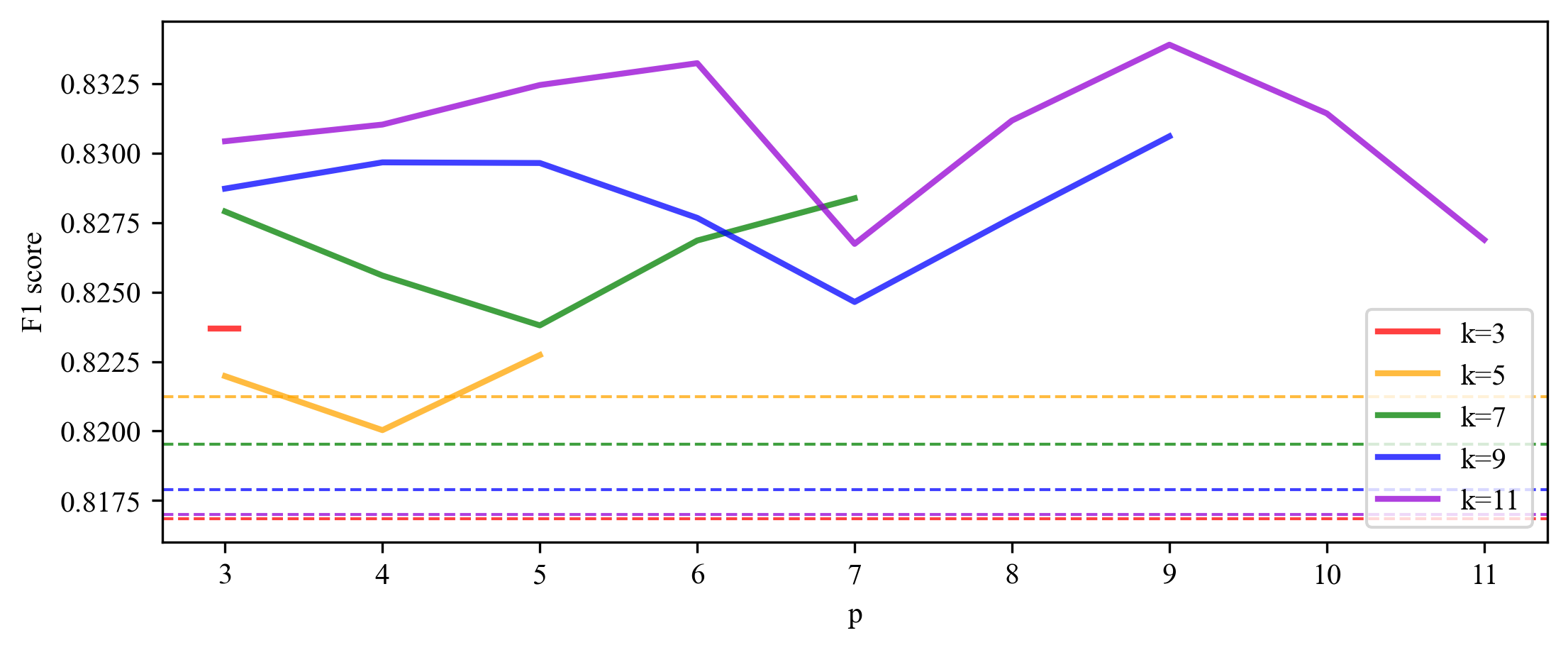}}
\hfill
\subfigure[spectrometer]{\includegraphics[width=5.6cm,height=2.12cm]{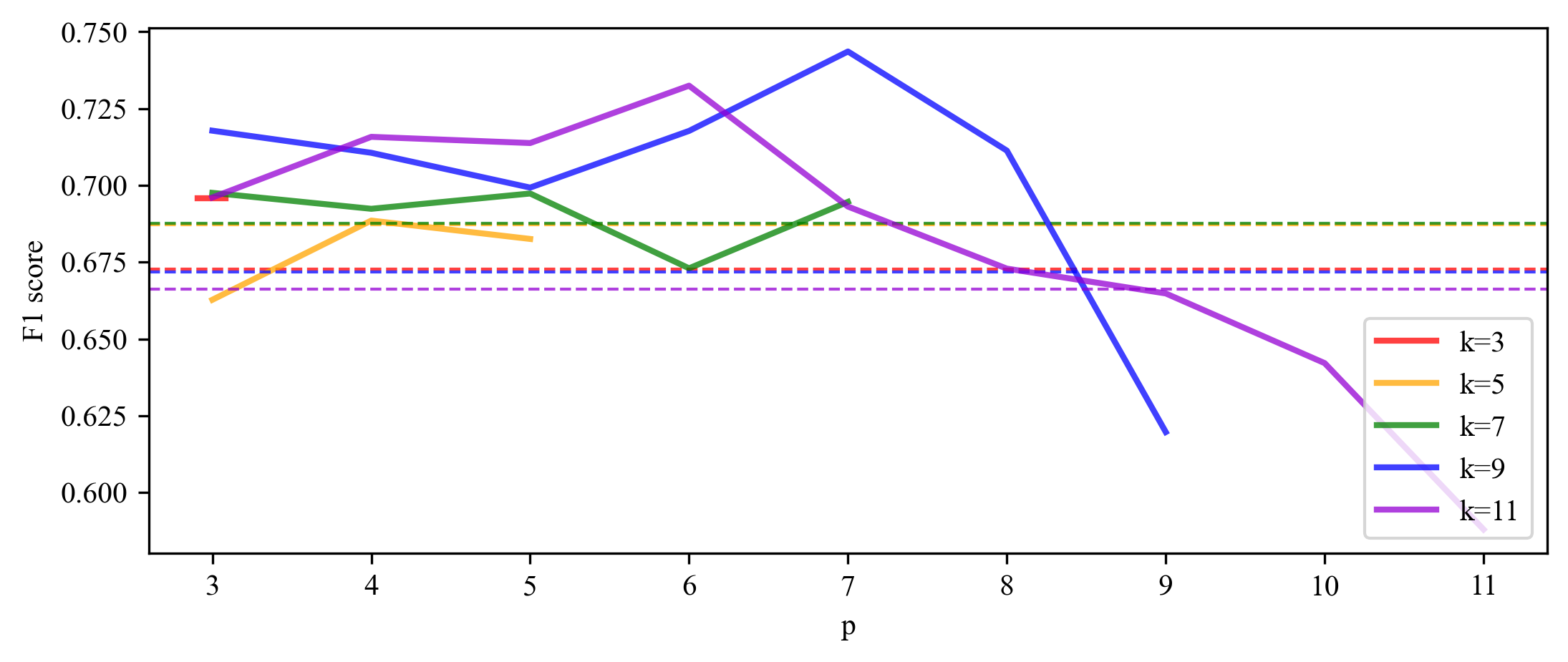}}

\subfigure[car\_eval\_34]{\includegraphics[width=5.6cm,height=2.12cm]{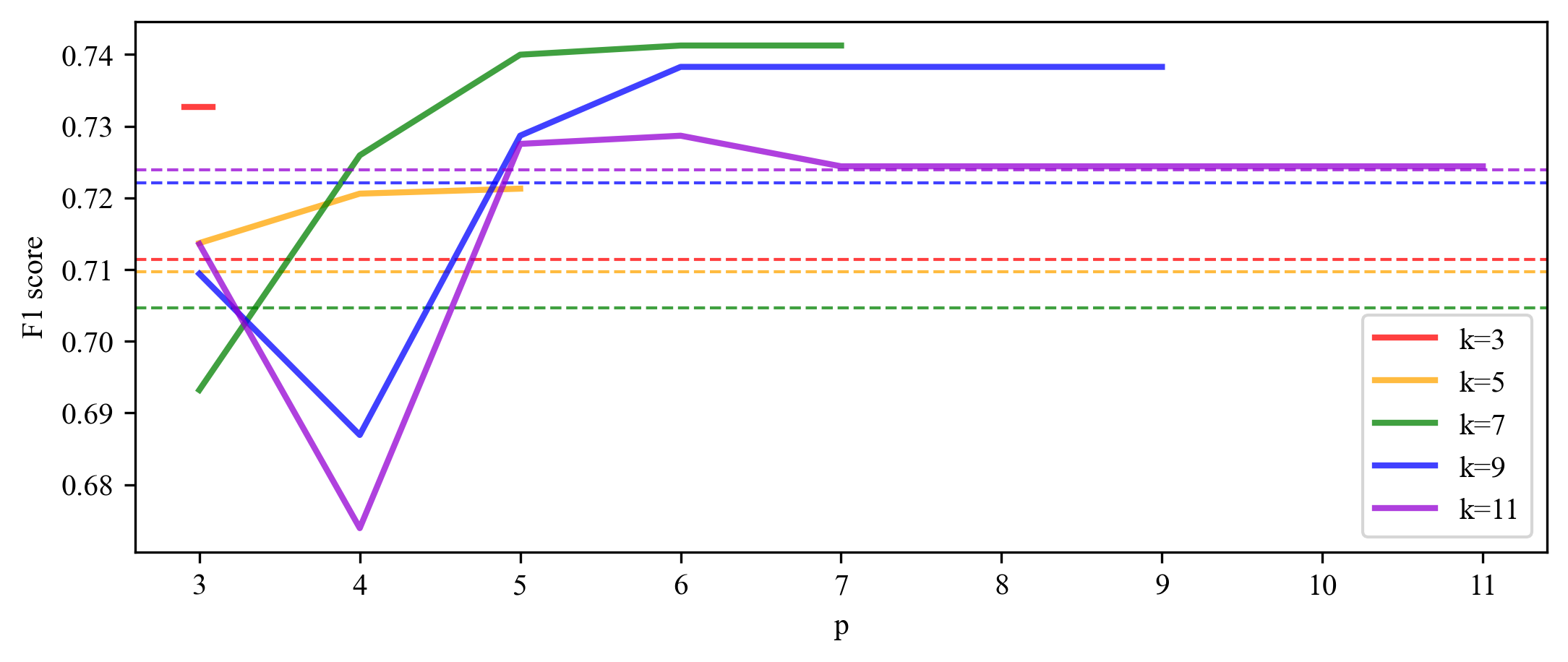}}
\hfill
\subfigure[us\_crime]{\includegraphics[width=5.6cm,height=2.12cm]{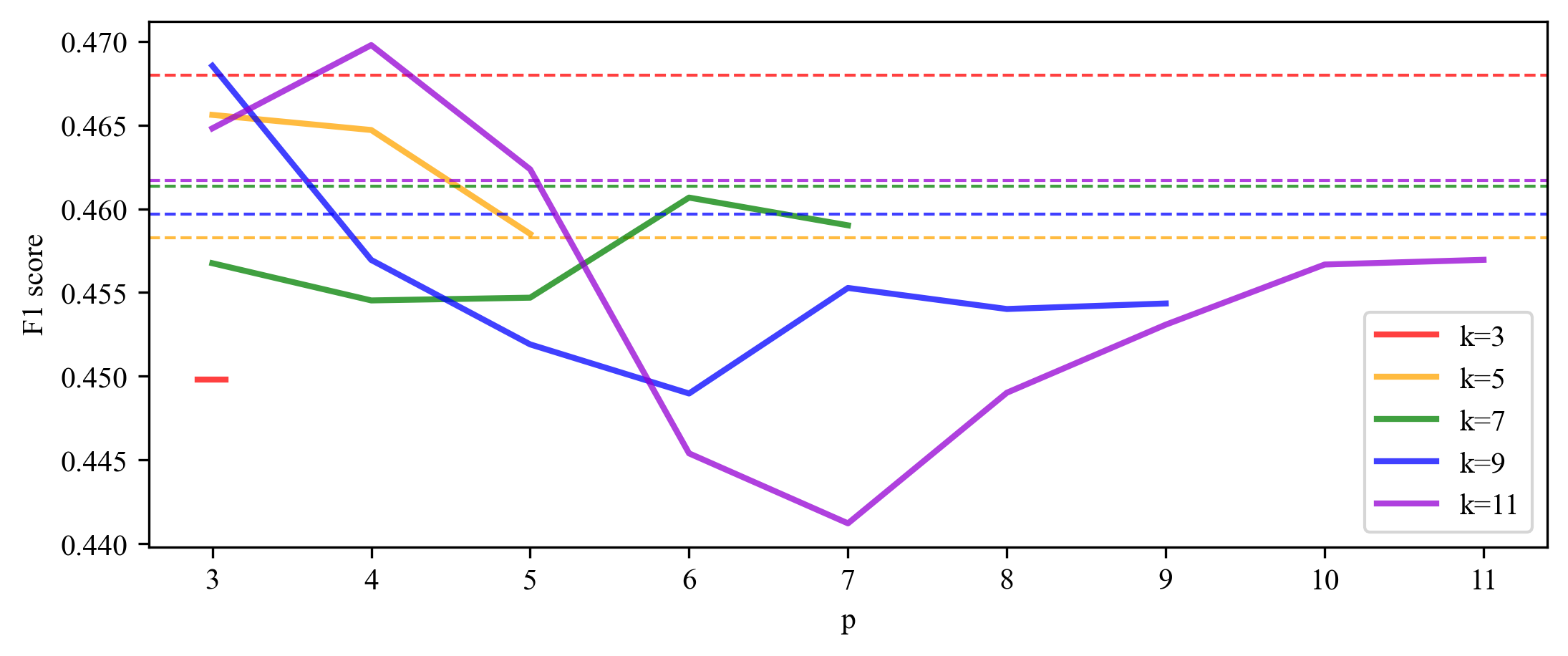}}
\hfill
\subfigure[yeast\_ml8]{\includegraphics[width=5.6cm,height=2.12cm]{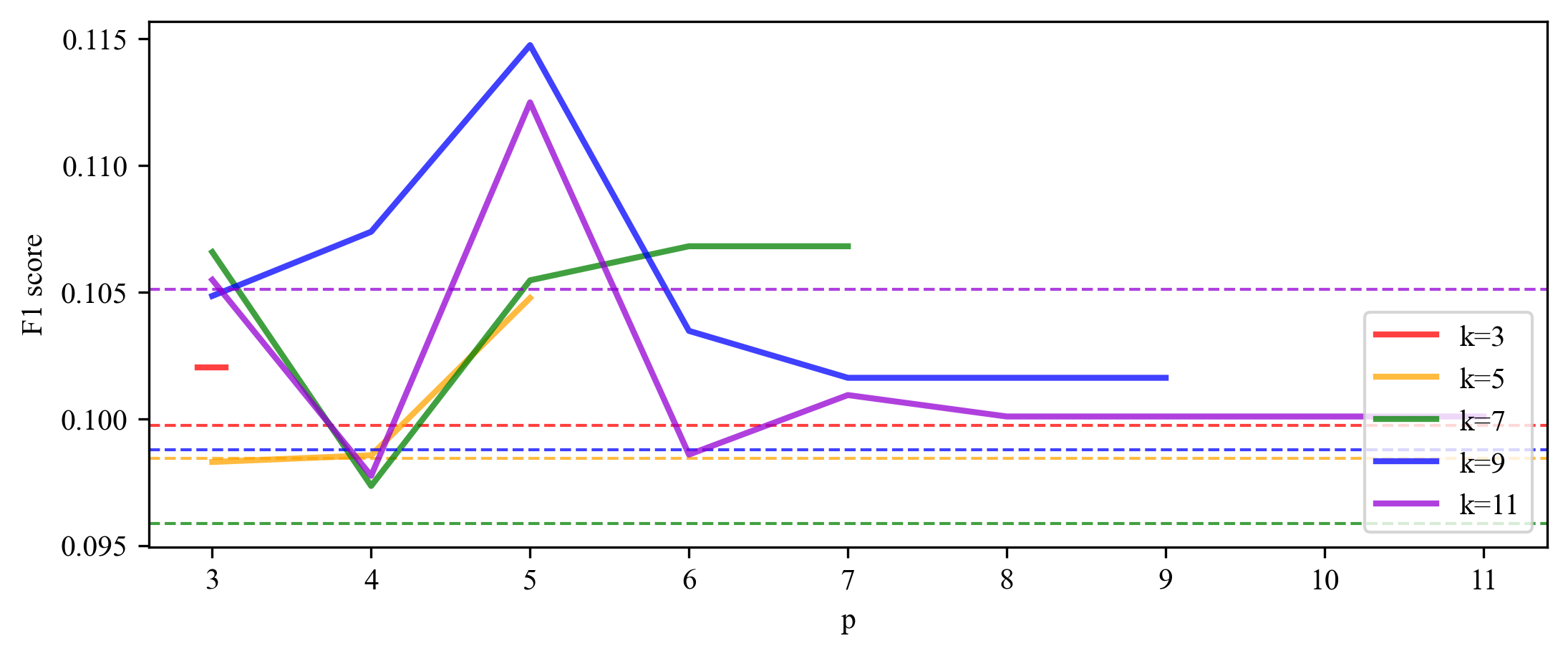}}

\subfigure[scene]{\includegraphics[width=5.6cm,height=2.12cm]{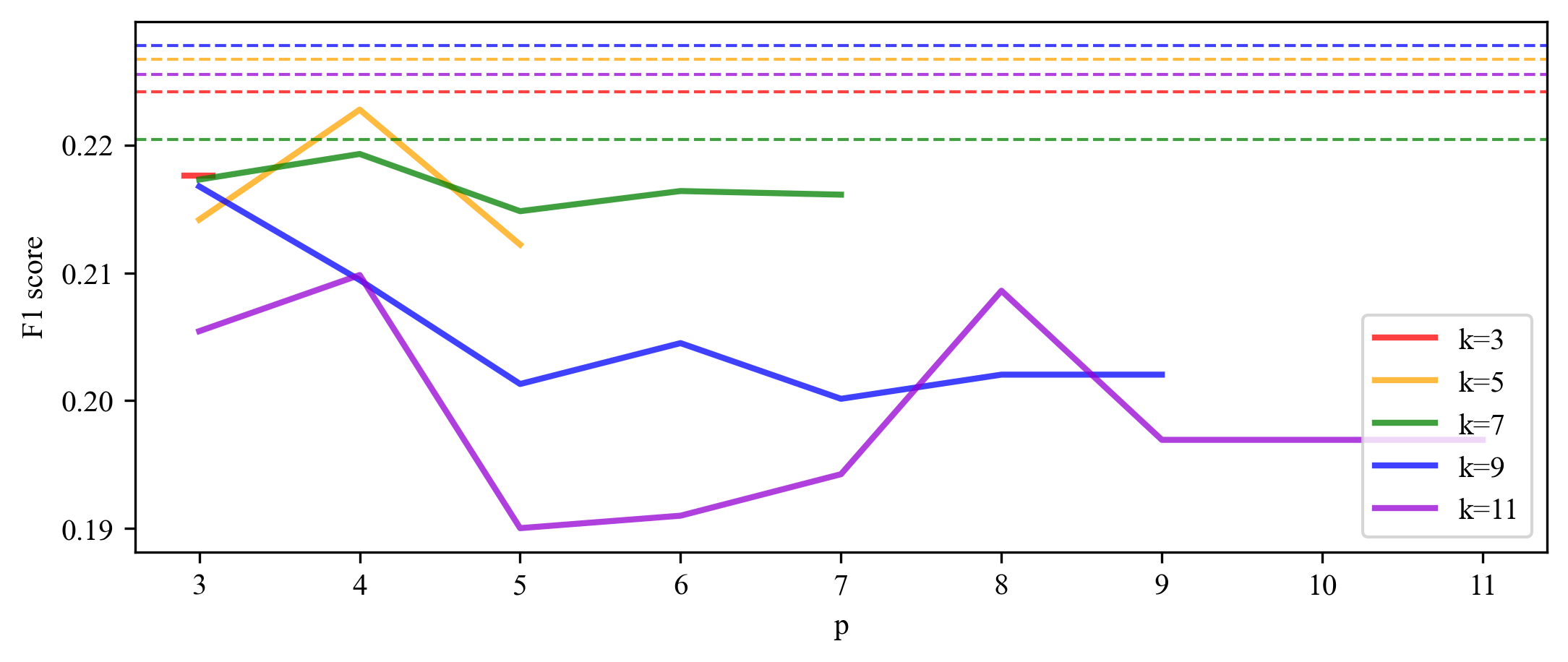}}
\hfill
\subfigure[libras\_move]{\includegraphics[width=5.6cm,height=2.12cm]{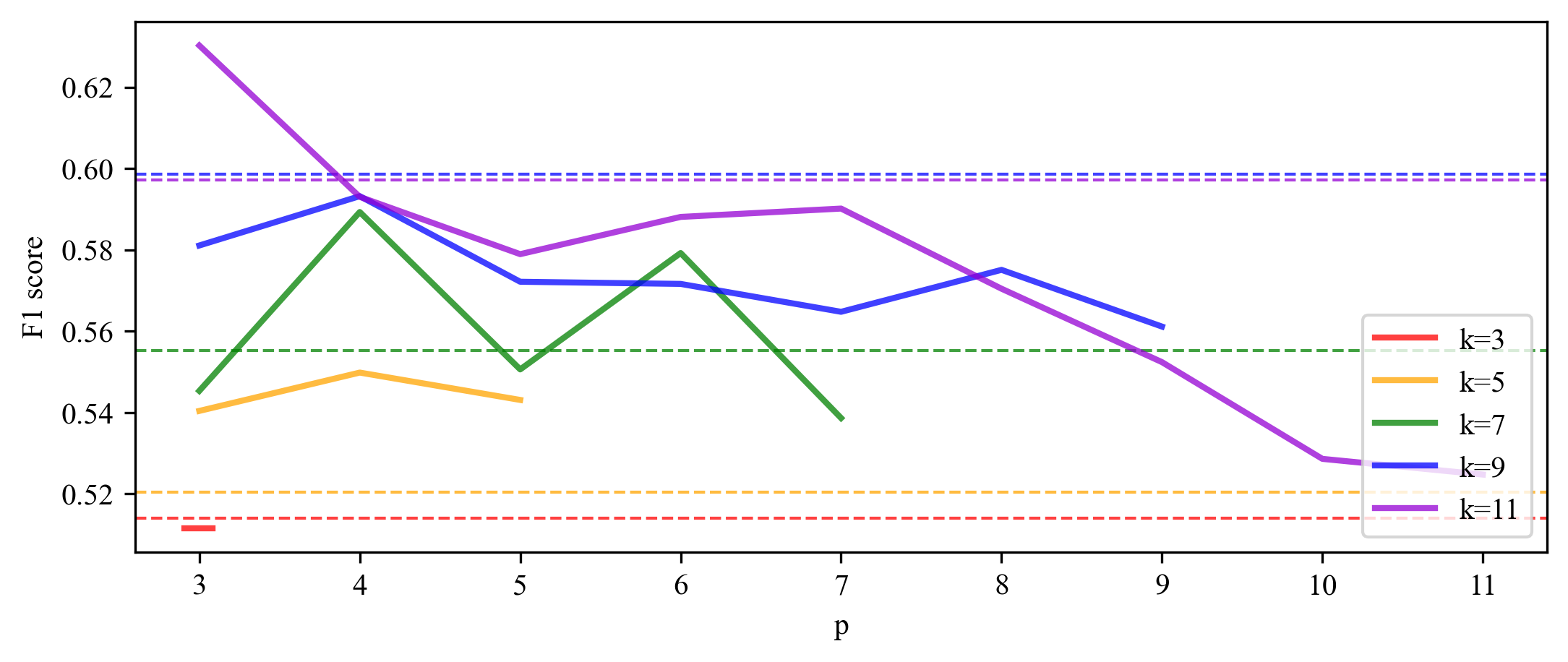}}
\hfill
\subfigure[thyroid\_sick]{\includegraphics[width=5.6cm,height=2.12cm]{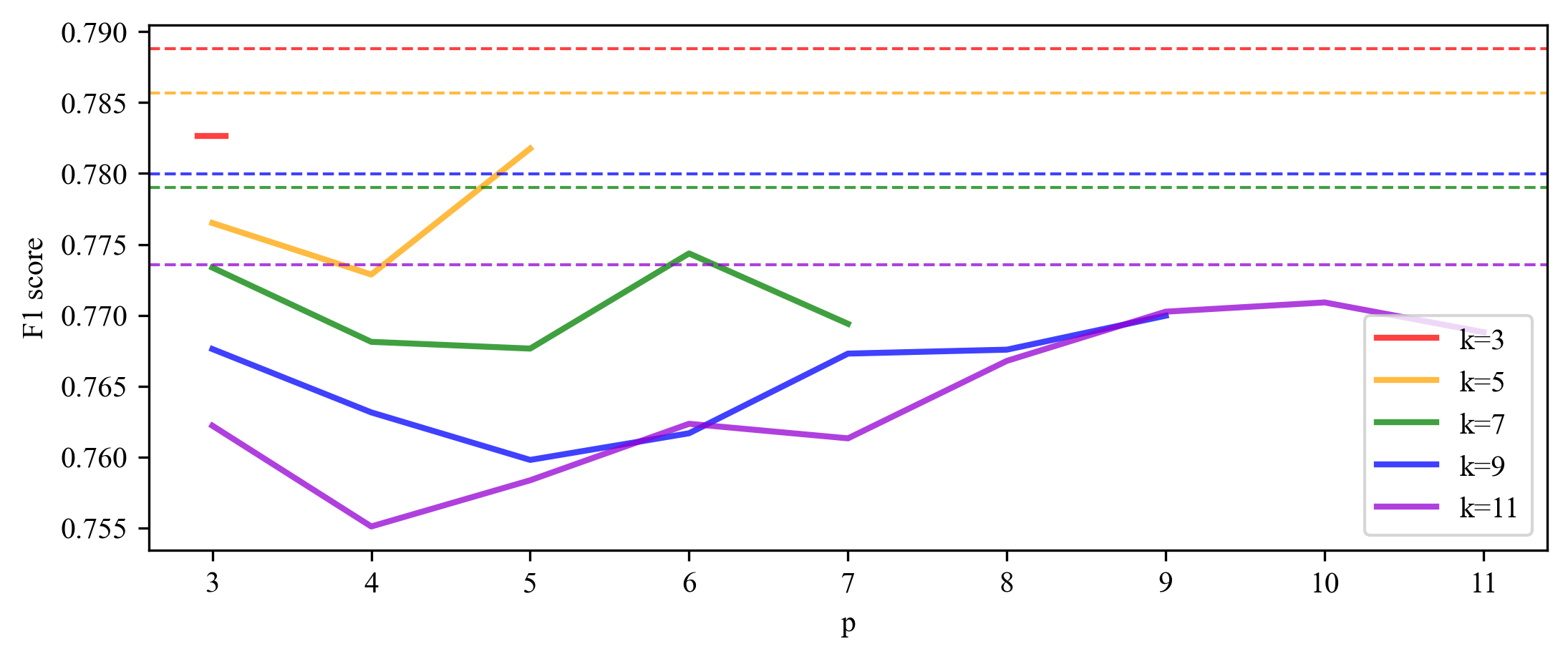}}

\subfigure[coil\_2000]{\includegraphics[width=5.6cm,height=2.12cm]{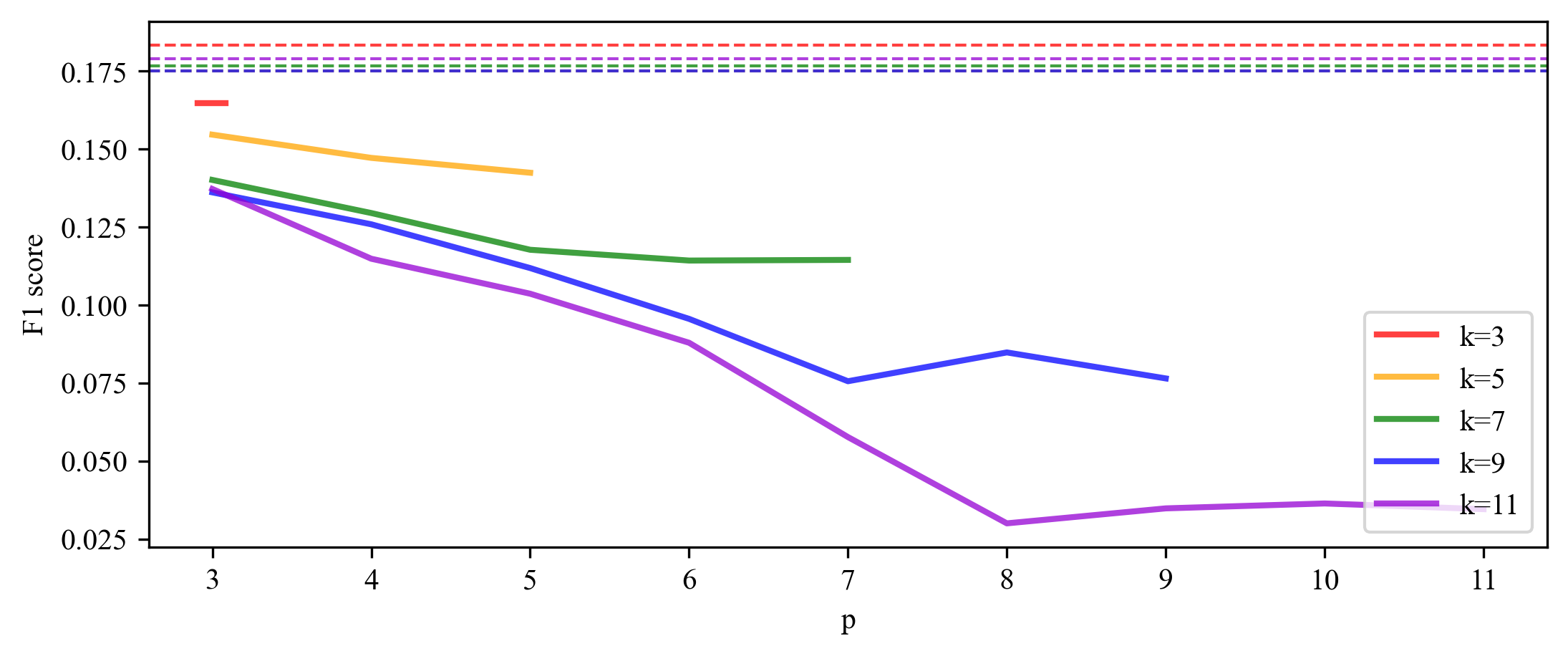}}
\hfill
\subfigure[solar\_flare\_m0]{\includegraphics[width=5.6cm,height=2.12cm]{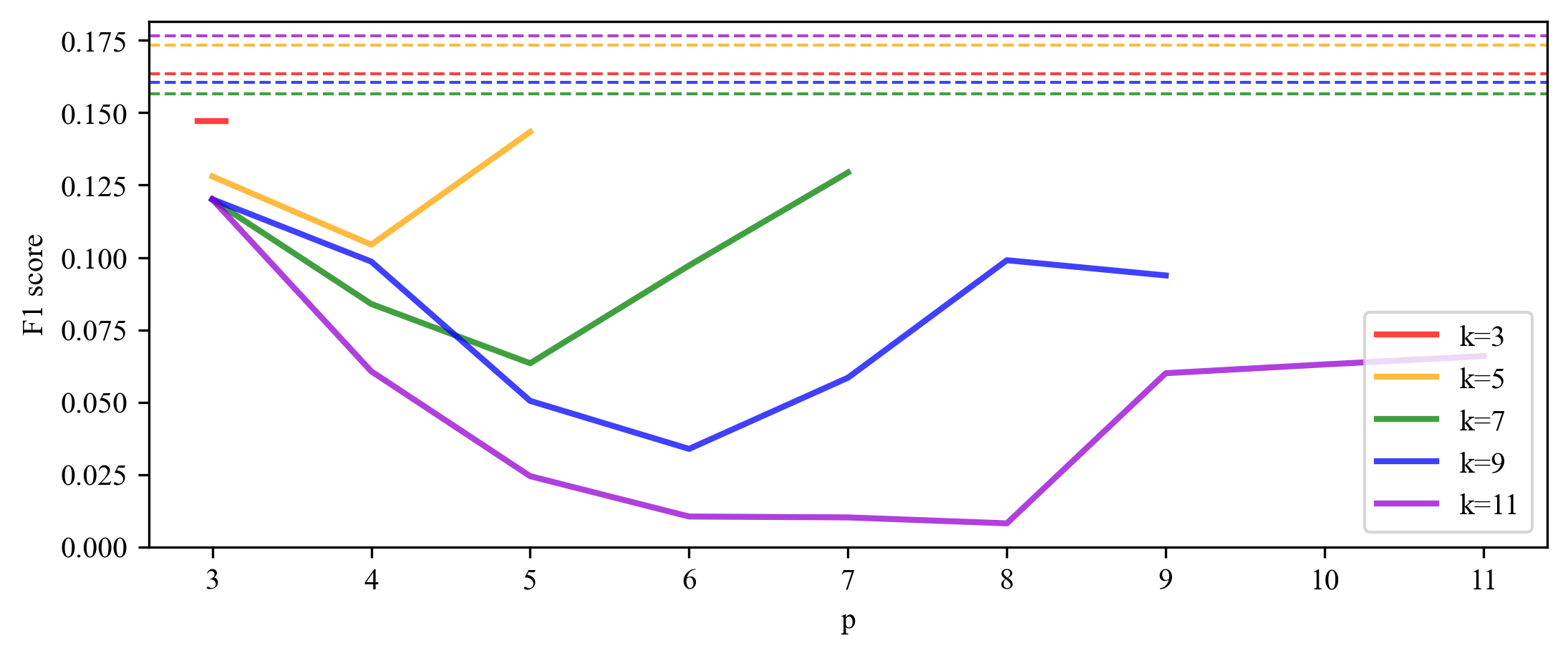}}
\hfill
\subfigure[oil]{\includegraphics[width=5.6cm,height=2.12cm]{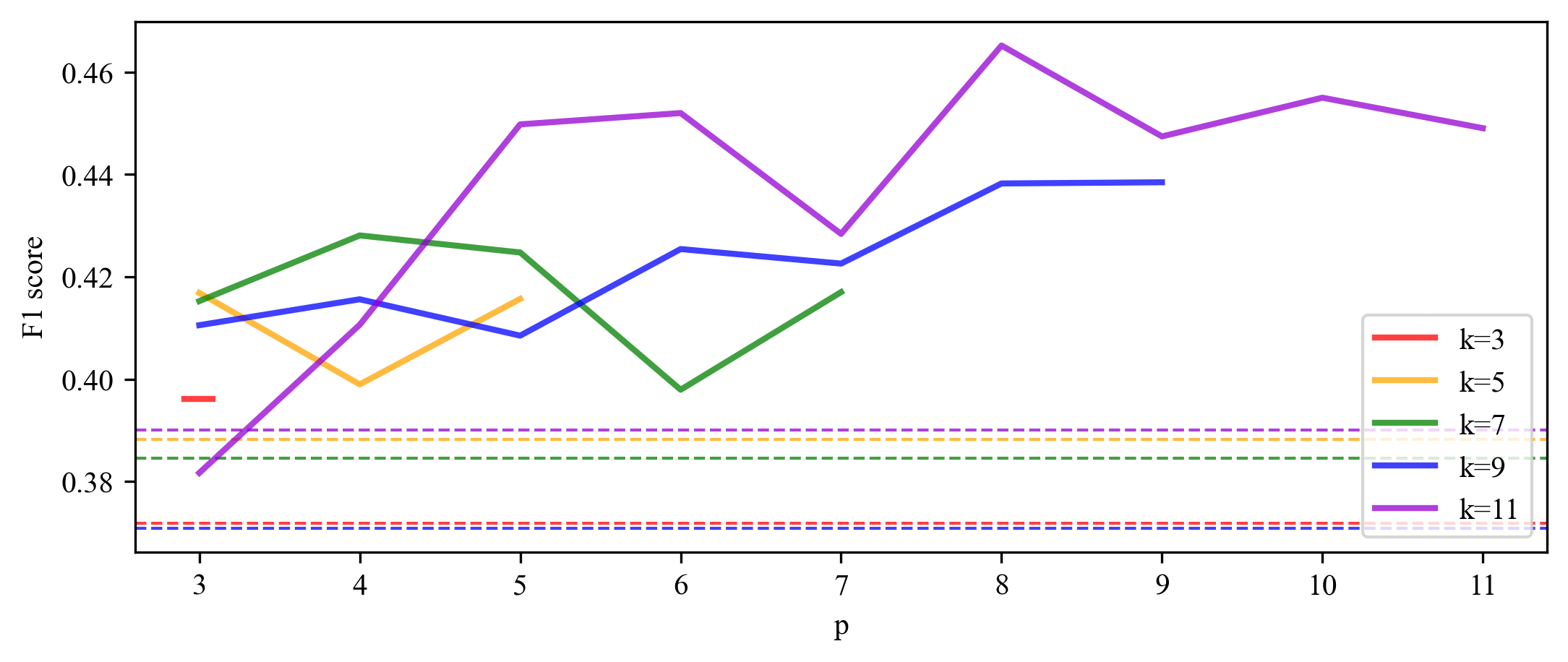}}

\subfigure[car\_eval\_4]{\includegraphics[width=5.6cm,height=2.12cm]{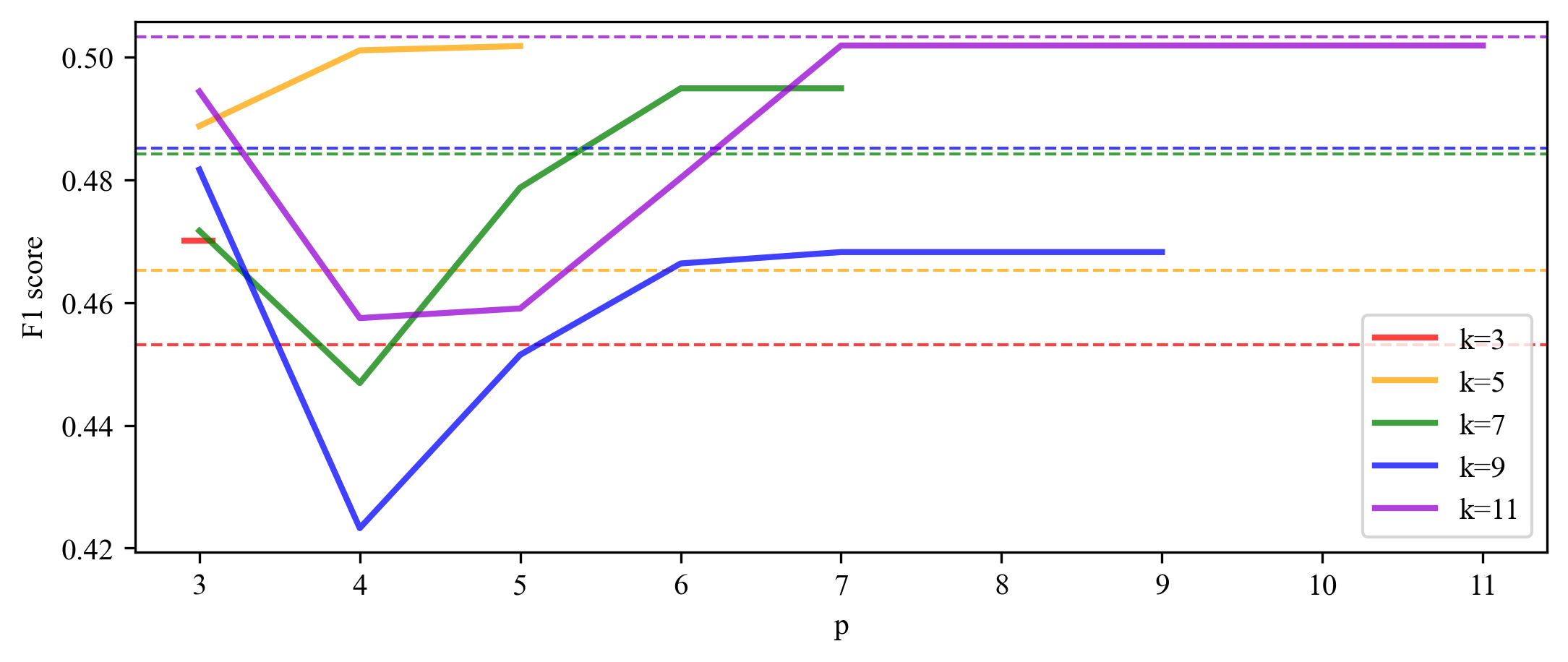}}
\hfill
\subfigure[wine\_quality]{\includegraphics[width=5.6cm,height=2.12cm]{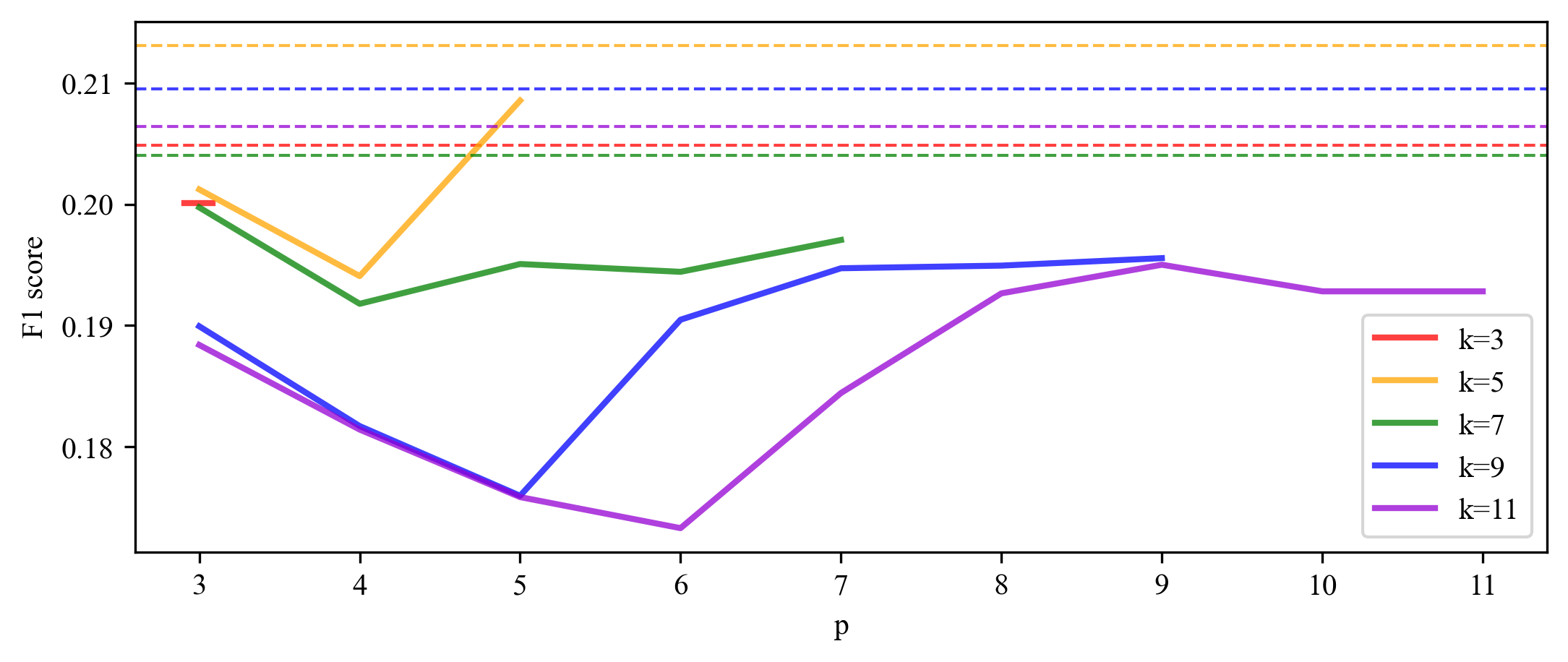}}
\hfill
\subfigure[letter\_img]{\includegraphics[width=5.6cm,height=2.12cm]{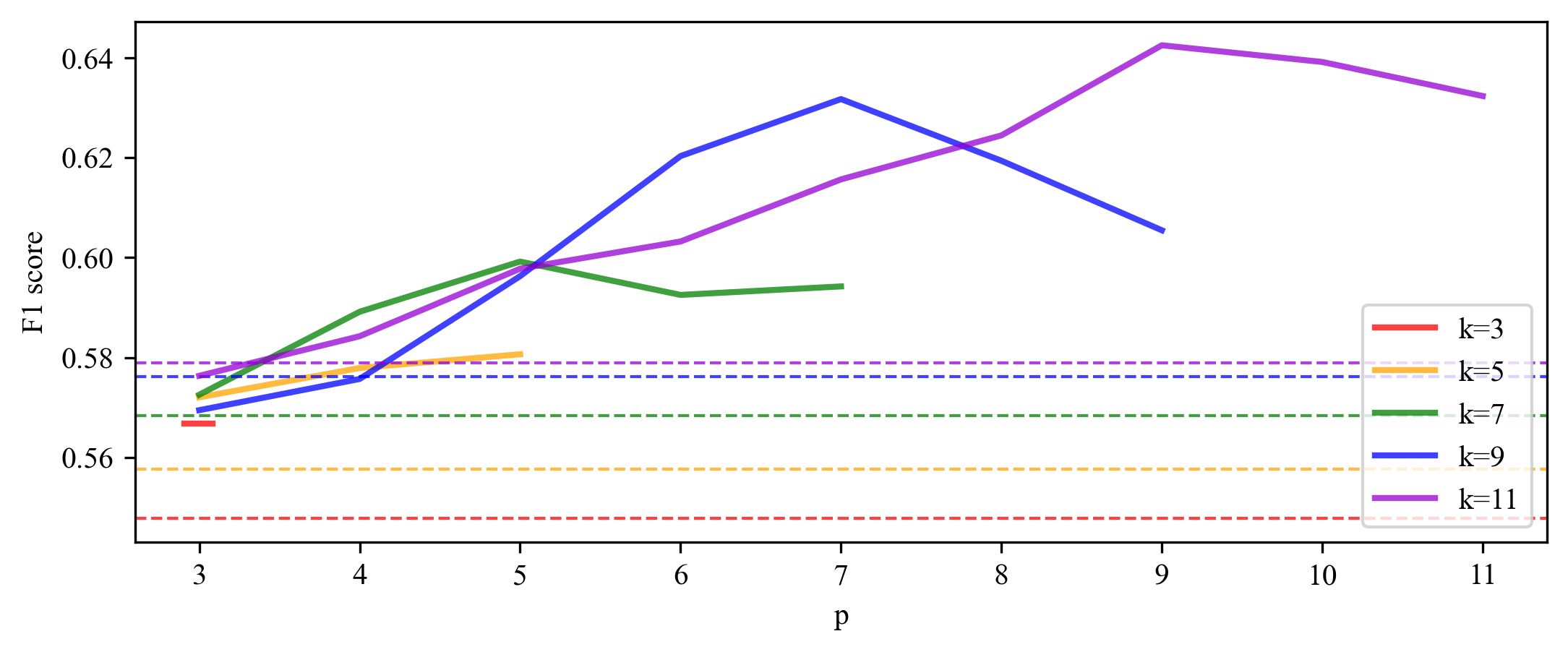}}

\subfigure[yeast\_me2]{\includegraphics[width=5.6cm,height=2.12cm]{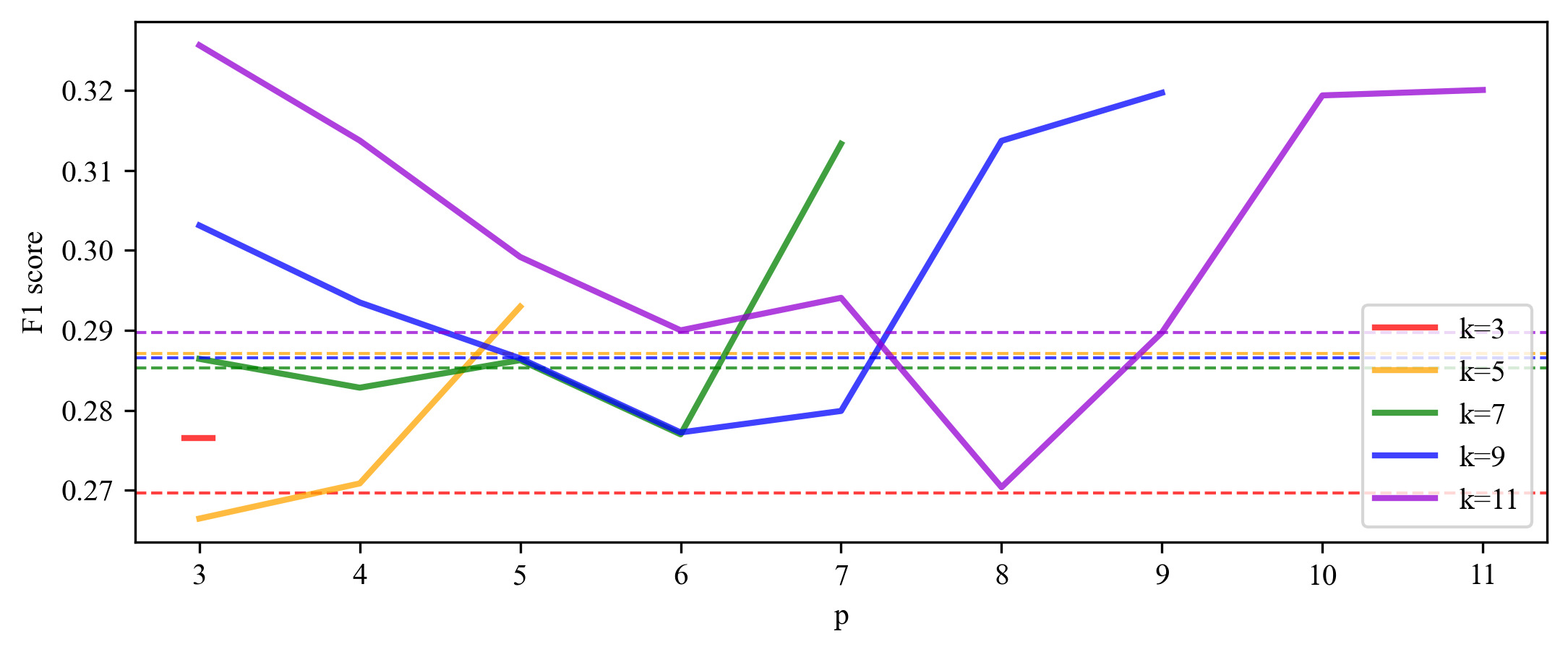}}
\hfill
\subfigure[ozone\_level]{\includegraphics[width=5.6cm,height=2.12cm]{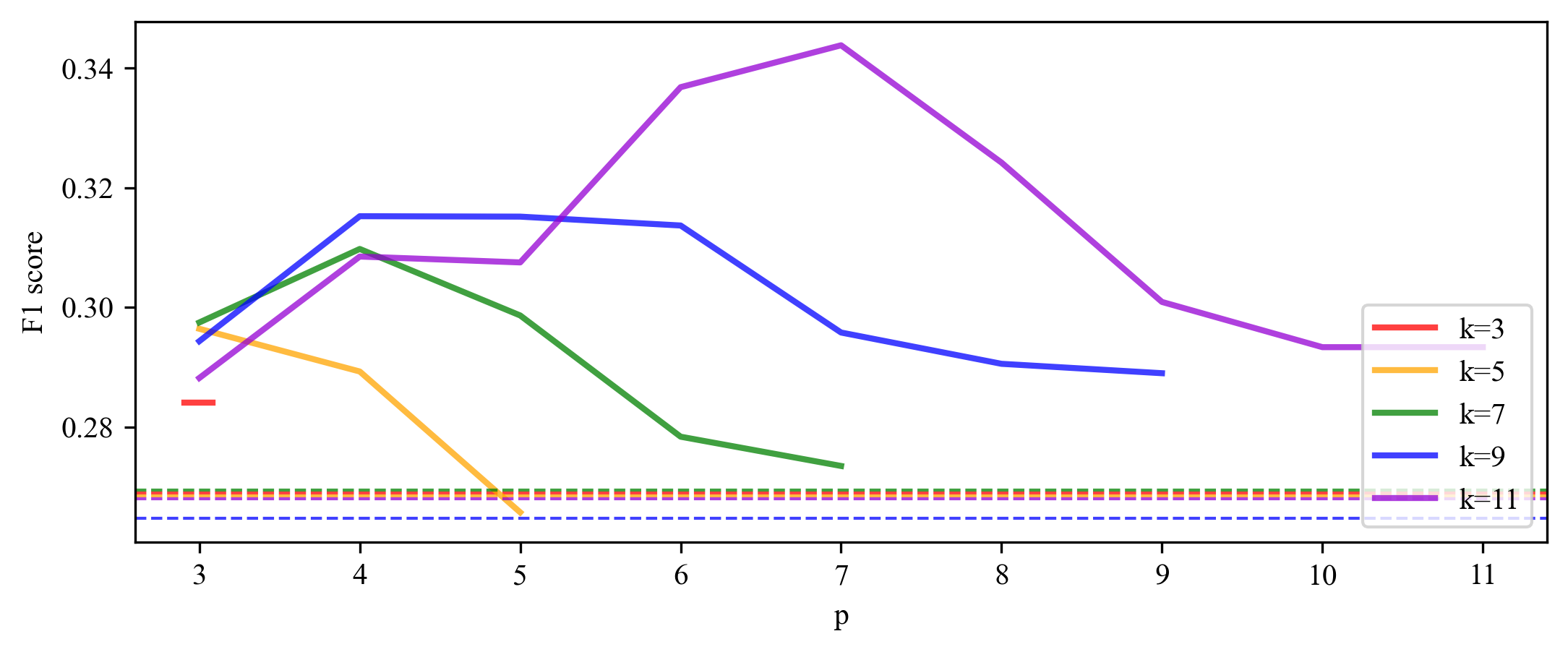}}
\hfill
\subfigure[abalone\_19]{\includegraphics[width=5.6cm,height=2.12cm]{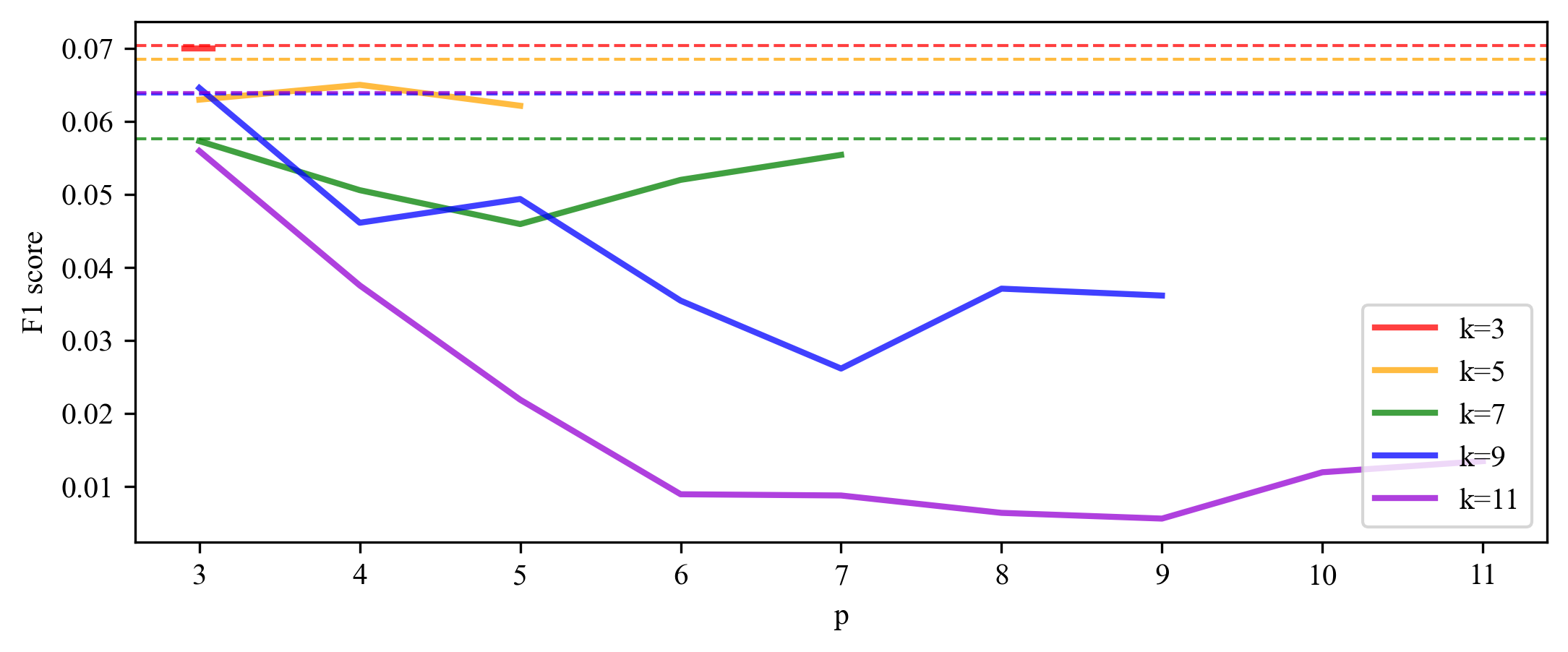}}

\vspace{-0.5em}
\caption{Sensitivity for Simplicial SMOTE's hyperparameters -- neighborhood size $k$ and maximum clique size $p$, followed by the gradient boosting classifier. Performances in terms of F1 score for various $k$ and $p$ are shown as solid lines. Baseline SMOTE performance for the same $k$ is shown as a dashed line of the same color.}
\end{figure} 

\clearpage

\section{Matthew's Correlation Coefficients}
\label{appendix_more_results}

\begin{table*}[h!]

\caption{Classification results on benchmark datasets for the $k$-NN classifier. Matthew's correlation coefficient averaged over $5$ repeats of $5$-fold (outer) cross-validation is reported. \textcolor{red}{\textbf{Best}} and \textcolor{blue}{\textbf{second-best}} results are highlighted.}\label{table:knnmcc}


\centering

\renewcommand{\arraystretch}{1.2}

\resizebox{\textwidth}{!}{

\begin{tabular}{l|rrrrrrrrrr|rrrr}
\toprule
 & \fontsize{8}{9}\selectfont\textbf{Imbalanced} & \fontsize{8}{9}\selectfont\textbf{Random} & \fontsize{8}{9}\selectfont\textbf{Global} & \fontsize{8}{9}\selectfont\textbf{SMOTE} & \fontsize{8}{9}\selectfont\textbf{Border.} & \fontsize{8}{9}\selectfont\textbf{Safelevel} & \fontsize{8}{9}\selectfont\textbf{ADASYN} & \fontsize{8}{9}\selectfont\textbf{MWMOTE} & \fontsize{8}{9}\selectfont\textbf{DBSMOTE} & \fontsize{8}{9}\selectfont\textbf{LVQ} & \fontsize{8}{9}\selectfont\textbf{Simplicial} & \fontsize{8}{9}\selectfont\textbf{S-Border.} & \fontsize{8}{9}\selectfont\textbf{S-Safe.} & \fontsize{8}{9}\selectfont\textbf{S-ADASYN} \\  \midrule

ecoli &     0.5469 &     0.5144 &     0.5622 &     0.5567 &     0.5581 &     0.5560 &     0.5431 &     0.5704 &     0.5999 &     0.5591 &     0.5942 &     0.5848 &     0.5307 &     0.5959 \\
optical\_digits &     0.9638 &     0.9439 &     0.9374 &     0.9360 &     0.9513 &     0.9386 &     0.9387 &     0.9319 &     0.9438 &     0.9577 &     0.9391 &     0.9511 &     0.9372 &     0.9367 \\
pen\_digits &     0.9920 &     0.9896 &     0.9884 &     0.9896 &     0.9917 &     0.9890 &     0.9909 &     0.9898 &     0.9913 &     0.9921 &     0.9906 &     0.9917 &     0.9903 &     0.9901 \\
abalone &     0.1409 &     0.2619 &     0.3279 &     0.2988 &     0.3037 &     0.3323 &     0.2935 &     0.3311 &     0.2542 &     0.2302 &     0.3146 &     0.3160 &     0.2997 &     0.3045 \\
sick\_euthyroid &     0.5413 &     0.5379 &     0.5526 &     0.5431 &     0.5386 &     0.4939 &     0.5402 &     0.5387 &     0.5673 &     0.5461 &     0.5659 &     0.5609 &     0.5565 &     0.5659 \\
spectrometer &     0.7704 &     0.8416 &     0.8279 &     0.8325 &     0.8470 &     0.8196 &     0.8325 &     0.8474 &     0.8012 &     0.8150 &     0.8492 &     0.8398 &     0.8292 &     0.8381 \\
car\_eval\_34 &     0.6199 &     0.5864 &     0.5909 &     0.5900 &     0.6048 &     0.5894 &     0.5984 &     0.6592 &     0.5864 &     0.7098 &     0.6363 &     0.6204 &     0.6390 &     0.6433 \\
us\_crime &     0.3765 &     0.4007 &     0.4252 &     0.4015 &     0.4357 &     0.4161 &     0.4010 &     0.3902 &     0.4007 &     0.4197 &     0.4194 &     0.4498 &     0.4055 &     0.4140 \\
yeast\_ml8 &     0.0668 &     0.0470 &     0.0892 &     0.0740 &     0.0858 &     0.0581 &     0.0706 &     0.0848 &     0.0470 &     0.0733 &     0.0830 &     0.0884 &     0.0842 &     0.0815 \\
scene &     0.1411 &     0.1816 &     0.2280 &     0.2100 &     0.2085 &     0.1813 &     0.1997 &     0.2133 &     0.1304 &     0.1961 &     0.1957 &     0.2182 &     0.1894 &     0.1992 \\
libras\_move &     0.7200 &     0.8003 &     0.7717 &     0.7669 &     0.7645 &     0.7388 &     0.7505 &     0.7843 &     0.8003 &     0.8040 &     0.7529 &     0.7561 &     0.7616 &     0.7506 \\
thyroid\_sick &     0.5149 &     0.5001 &     0.5028 &     0.5072 &     0.5081 &     0.4419 &     0.5041 &     0.5078 &     0.4892 &     0.4986 &     0.5218 &     0.5168 &     0.4968 &     0.5314 \\
coil\_2000 &     0.0530 &     0.1105 &     0.1111 &     0.1099 &     0.1141 &     0.1082 &     0.1103 &     0.1151 &     0.0582 &     0.0787 &     0.1090 &     0.1118 &     0.1069 &     0.1136 \\
solar\_flare\_m0 &     0.0483 &     0.1762 &     0.1648 &     0.1671 &     0.1862 &     0.1888 &     0.1784 &     0.1717 &     0.0385 &     0.1674 &     0.1778 &     0.1926 &     0.1923 &     0.1759 \\
oil &     0.3795 &     0.4285 &     0.4505 &     0.4382 &     0.4575 &     0.3641 &     0.4178 &     0.4066 &     0.4285 &     0.4457 &     0.4943 &     0.4917 &     0.4087 &     0.4654 \\
car\_eval\_4 &     0.2031 &     0.3981 &     0.5383 &     0.5035 &     0.5054 &     0.4718 &     0.4975 &     0.5606 &     0.3981 &     0.7011 &     0.6117 &     0.6143 &     0.6353 &     0.6050 \\
wine\_quality &     0.1857 &     0.2785 &     0.2254 &     0.2556 &     0.2628 &     0.2335 &     0.2525 &     0.2252 &     0.1383 &     0.1981 &     0.2522 &     0.2599 &     0.2623 &     0.2528 \\
letter\_img &     0.9712 &     0.9516 &     0.9087 &     0.9403 &     0.9598 &     0.9286 &     0.9520 &     0.9129 &     0.9648 &     0.9661 &     0.9546 &     0.9600 &     0.9537 &     0.9482 \\
yeast\_me2 &     0.2672 &     0.3118 &     0.2999 &     0.3187 &     0.3572 &     0.2955 &     0.3157 &     0.3297 &     0.2939 &     0.2942 &     0.3389 &     0.3662 &     0.2990 &     0.3305 \\
ozone\_level &     0.2007 &     0.2451 &     0.2526 &     0.2441 &     0.2470 &     0.2747 &     0.2453 &     0.2419 &     0.2451 &     0.2247 &     0.2531 &     0.2722 &     0.2305 &     0.2468 \\
abalone\_19 &    -0.0001 &     0.0186 &     0.0746 &     0.0506 &     0.0423 &     0.0290 &     0.0566 &     0.0318 &     0.0124 &     0.0333 &     0.0653 &     0.0387 &     0.0406 &     0.0627 \\ \midrule
\textbf{mean} &     0.4144 &     0.4535 &     0.4681 &     0.4635 &     0.4729 &     0.4500 &     0.4614 &     0.4688 &     0.4376 &     0.4720 &     \textcolor{blue}{\textbf{0.4819}} &     \textcolor{red}{\textbf{0.4858}} &     0.4690 &     0.4787 \\
\textbf{rank} &    10.6667 &     9.1667 &     6.9048 &     8.3810 &     4.9762 &     9.9524 &     8.2381 &     7.3333 &     9.7857 &     7.2857 &     \textcolor{blue}{\textbf{4.9524}} &     \textcolor{red}{\textbf{3.6429}} &     7.9048 &     5.8095 \\ \bottomrule

\end{tabular}

}

\end{table*}
\begin{table*}[h!]

\caption{Classification results on benchmark datasets for the gradient boosting classifier. Matthew's correlation coefficient averaged over $5$ repeats of $5$-fold (outer) cross-validation is reported. \textcolor{red}{\textbf{Best}} and \textcolor{blue}{\textbf{second-best}} results are highlighted.}

\vspace{-0.5em}

\centering

\renewcommand{\arraystretch}{1.2}

\resizebox{\textwidth}{!}{

\begin{tabular}{l|rrrrrrrrrr|rrrr}
\toprule
 & \fontsize{8}{9}\selectfont\textbf{Imbalanced} & \fontsize{8}{9}\selectfont\textbf{Random} & \fontsize{8}{9}\selectfont\textbf{Global} & \fontsize{8}{9}\selectfont\textbf{SMOTE} & \fontsize{8}{9}\selectfont\textbf{Border.} & \fontsize{8}{9}\selectfont\textbf{Safelevel} & \fontsize{8}{9}\selectfont\textbf{ADASYN} & \fontsize{8}{9}\selectfont\textbf{MWMOTE} & \fontsize{8}{9}\selectfont\textbf{DBSMOTE} & \fontsize{8}{9}\selectfont\textbf{LVQ} & \fontsize{8}{9}\selectfont\textbf{Simplicial} & \fontsize{8}{9}\selectfont\textbf{S-Border.} & \fontsize{8}{9}\selectfont\textbf{S-Safe.} & \fontsize{8}{9}\selectfont\textbf{S-ADASYN} \\  \midrule

ecoli           &     0.5656 &     0.5484 &     0.5850 &     0.5765 &     0.5399 &     0.5477 &     0.5673 &     0.5763 &     0.5556 &     0.5589 &     0.6059 &     0.5693 &     0.5598 &     0.5985 \\
optical\_digits &     0.5905 &     0.6539 &     0.7158 &     0.6973 &     0.6563 &     0.6969 &     0.6648 &     0.7059 &     0.6549 &     0.6258 &     0.7300 &     0.6657 &     0.7388 &     0.7103 \\
pen\_digits     &     0.6846 &     0.7876 &     0.6739 &     0.7960 &     0.6744 &     0.7969 &     0.6692 &     0.7158 &     0.7871 &     0.7808 &     0.8121 &     0.6657 &     0.8106 &     0.7043 \\
abalone         &    -0.0004 &     0.3610 &     0.3539 &     0.3630 &     0.3612 &     0.3614 &     0.3602 &     0.3670 &     0.3602 &     0.3524 &     0.3595 &     0.3624 &     0.3586 &     0.3647 \\
sick\_euthyroid &     0.8344 &     0.8100 &     0.8071 &     0.8147 &     0.8105 &     0.7208 &     0.8132 &     0.8153 &     0.8241 &     0.7961 &     0.8243 &     0.8178 &     0.8263 &     0.8168 \\
spectrometer    &     0.6347 &     0.7044 &     0.6099 &     0.7002 &     0.7287 &     0.7534 &     0.6840 &     0.6647 &     0.7828 &     0.5904 &     0.7989 &     0.7348 &     0.7277 &     0.7684 \\
car\_eval\_34   &     0.3449 &     0.6552 &     0.7297 &     0.7098 &     0.7169 &     0.6839 &     0.7231 &     0.6979 &     0.6556 &     0.5903 &     0.7269 &     0.7159 &     0.7269 &     0.6994 \\
us\_crime       &     0.4424 &     0.4628 &     0.4487 &     0.4566 &     0.4607 &     0.4662 &     0.4529 &     0.4447 &     0.4638 &     0.4008 &     0.4416 &     0.4493 &     0.4229 &     0.4231 \\
yeast\_ml8      &    -0.0005 &     0.0199 &     0.0589 &     0.0461 &     0.0507 &     0.0593 &     0.0383 &     0.0351 &     0.0302 &     0.0311 &     0.0549 &     0.0484 &     0.0560 &     0.0425 \\
scene           &     0.0242 &     0.2055 &     0.2034 &     0.2176 &     0.2043 &     0.1981 &     0.2112 &     0.2105 &     0.0920 &     0.0240 &     0.1732 &     0.1827 &     0.1736 &     0.1813 \\
libras\_move    &     0.5333 &     0.6799 &     0.6419 &     0.6504 &     0.6577 &     0.6273 &     0.6232 &     0.6335 &     0.6641 &     0.6718 &     0.6919 &     0.6656 &     0.6418 &     0.6803 \\
thyroid\_sick   &     0.8286 &     0.7762 &     0.7241 &     0.7837 &     0.7783 &     0.6337 &     0.7778 &     0.7297 &     0.7981 &     0.7203 &     0.7831 &     0.7732 &     0.7751 &     0.7764 \\
coil\_2000      &    -0.0022 &     0.1911 &     0.1705 &     0.1723 &     0.1694 &     0.1715 &     0.1711 &     0.1655 &     0.0720 &     0.0063 &     0.1544 &     0.1655 &     0.1550 &     0.1535 \\
solar\_flare\_m0&     0.0236 &     0.1855 &     0.2054 &     0.1426 &     0.1520 &     0.1699 &     0.1345 &     0.1510 &     0.0186 &     0.1412 &     0.1237 &     0.1393 &     0.1223 &     0.1231 \\
oil             &     0.3920 &     0.3960 &     0.3791 &     0.3764 &     0.3990 &     0.3912 &     0.3709 &     0.3516 &     0.3812 &     0.3789 &     0.4495 &     0.4347 &     0.3927 &     0.4333 \\
car\_eval\_4    &     0.0000 &     0.4745 &     0.5345 &     0.5025 &     0.5023 &     0.4938 &     0.4874 &     0.4934 &     0.4721 &     0.5443 &     0.4835 &     0.4876 &     0.4835 &     0.4413 \\
wine\_quality   &     0.1235 &     0.2432 &     0.1961 &     0.2211 &     0.2321 &     0.2317 &     0.2098 &     0.2119 &     0.1609 &     0.2672 &     0.2114 &     0.2296 &     0.2217 &     0.1957 \\
letter\_img     &     0.6425 &     0.5151 &     0.5810 &     0.5844 &     0.4825 &     0.5660 &     0.5043 &     0.5292 &     0.5664 &     0.5507 &     0.6318 &     0.4819 &     0.6362 &     0.5725 \\
yeast\_me2      &     0.1124 &     0.3099 &     0.3240 &     0.3127 &     0.3442 &     0.3334 &     0.3055 &     0.3209 &     0.2719 &     0.3451 &     0.3266 &     0.3389 &     0.3133 &     0.3108 \\
ozone\_level    &     0.0668 &     0.2838 &     0.2627 &     0.2734 &     0.2832 &     0.2582 &     0.2652 &     0.2732 &     0.2856 &     0.2675 &     0.2971 &     0.2790 &     0.2789 &     0.2920 \\
abalone\_19     &    -0.0029 &     0.0972 &     0.0657 &     0.0890 &     0.0738 &     0.0880 &     0.0900 &     0.0800 &     0.0481 &     0.0831 &     0.0781 &     0.0665 &     0.0871 &     0.0806 \\ \midrule
\textbf{mean}   &     0.3256 &     0.4458 &     0.4415 &     0.4517 &     0.4418 &     0.4404 &     0.4345 &     0.4368 &     0.4260 &     0.4156 &     \textcolor{red}{\textbf{0.4647}} &     0.4416 &     \textcolor{blue}{\textbf{0.4528}} &     0.4461 \\
\textbf{rank}   &    10.9524 &     6.9524 &     7.4762 &     \textcolor{blue}{\textbf{5.6667}} &     6.6190 &     6.8095 &     8.4286 &     7.8095 &     8.7619 &     9.4286 &     \textcolor{red}{\textbf{5.1905}} &     7.0476 &     6.8571 &     7.0000 \\ \bottomrule

\end{tabular}

}

\end{table*}

\clearpage
\section{Statistical significance}\label{appendix_significance} 

\begin{table*}[h!]

\caption{P-values of the Conover-Friedman post-hoc test for the $k$-NN classifier and F1 score.}
\label{pvalues_knn_results}

\centering

\resizebox{\textwidth}{!}{

\begin{tabular}{l|rrrrrrrrrr|rrrr}
\toprule
 & \fontsize{8}{9}\selectfont\textbf{Imbalanced} & \fontsize{8}{9}\selectfont\textbf{Random} & \fontsize{8}{9}\selectfont\textbf{Global} & \fontsize{8}{9}\selectfont\textbf{SMOTE} & \fontsize{8}{9}\selectfont\textbf{Border.} & \fontsize{8}{9}\selectfont\textbf{Safelevel} & \fontsize{8}{9}\selectfont\textbf{ADASYN} & \fontsize{8}{9}\selectfont\textbf{MWMOTE} & \fontsize{8}{9}\selectfont\textbf{DBSMOTE} & \fontsize{8}{9}\selectfont\textbf{LVQ} & \fontsize{8}{9}\selectfont\textbf{Simplicial} & \fontsize{8}{9}\selectfont\textbf{S-Border.} & \fontsize{8}{9}\selectfont\textbf{S-Safe.} & \fontsize{8}{9}\selectfont\textbf{S-ADASYN} \\ \midrule

\textbf{Imbalanced} &      1.000 &      0.000 &      0.002 &      0.025 &      0.000 &      0.105 &      0.042 &      0.001 &      0.009 &      0.000 &      0.000 &      0.000 &      0.000 &      0.000 \\
\textbf{Random} &      0.000 &      1.000 &      0.518 &      0.140 &      0.023 &      0.036 &      0.092 &      0.819 &      0.279 &      0.349 &      0.085 &      0.001 &      0.917 &      0.574 \\
\textbf{Global} &      0.002 &      0.518 &      1.000 &      0.405 &      0.004 &      0.145 &      0.298 &      0.677 &      0.662 &      0.114 &      0.018 &      0.000 &      0.588 &      0.227 \\
\textbf{SMOTE} &      0.025 &      0.140 &      0.405 &      1.000 &      0.000 &      0.532 &      0.835 &      0.212 &      0.692 &      0.016 &      0.001 &      0.000 &      0.170 &      0.042 \\
\textbf{Border.} &      0.000 &      0.023 &      0.004 &      0.000 &      1.000 &      0.000 &      0.000 &      0.012 &      0.001 &      0.176 &      0.574 &      0.360 &      0.017 &      0.085 \\
\textbf{Safelevel} &      0.105 &      0.036 &      0.145 &      0.532 &      0.000 &      1.000 &      0.677 &      0.062 &      0.308 &      0.003 &      0.000 &      0.000 &      0.046 &      0.008 \\
\textbf{ADASYN} &      0.042 &      0.092 &      0.298 &      0.835 &      0.000 &      0.677 &      1.000 &      0.145 &      0.546 &      0.009 &      0.001 &      0.000 &      0.114 &      0.025 \\
\textbf{MWMOTE} &      0.001 &      0.819 &      0.677 &      0.212 &      0.012 &      0.062 &      0.145 &      1.000 &      0.393 &      0.244 &      0.051 &      0.001 &      0.900 &      0.429 \\
\textbf{DBSMOTE} &      0.009 &      0.279 &      0.662 &      0.692 &      0.001 &      0.308 &      0.546 &      0.393 &      1.000 &      0.044 &      0.005 &      0.000 &      0.328 &      0.101 \\
\textbf{LVQ} &      0.000 &      0.349 &      0.114 &      0.016 &      0.176 &      0.003 &      0.009 &      0.244 &      0.044 &      1.000 &      0.429 &      0.024 &      0.298 &      0.708 \\ \hline
\textbf{Simplicial} &      0.000 &      0.085 &      0.018 &      0.001 &      0.574 &      0.000 &      0.001 &      0.051 &      0.005 &      0.429 &      1.000 &      0.140 &      0.068 &      0.244 \\ 
\textbf{S-Border.} &      0.000 &      0.001 &      0.000 &      0.000 &      0.360 &      0.000 &      0.000 &      0.001 &      0.000 &      0.024 &      0.140 &      1.000 &      0.001 &      0.009 \\
\textbf{S-Safelevel} &      0.000 &      0.917 &      0.588 &      0.170 &      0.017 &      0.046 &      0.114 &      0.900 &      0.328 &      0.298 &      0.068 &      0.001 &      1.000 &      0.505 \\
\textbf{S-ADASYN} &      0.000 &      0.574 &      0.227 &      0.042 &      0.085 &      0.008 &      0.025 &      0.429 &      0.101 &      0.708 &      0.244 &      0.009 &      0.505 &      1.000 \\ \bottomrule

\end{tabular}

}

\end{table*}

\begin{figure*}[h!] 

\centering
\includegraphics[width=0.9\textwidth]{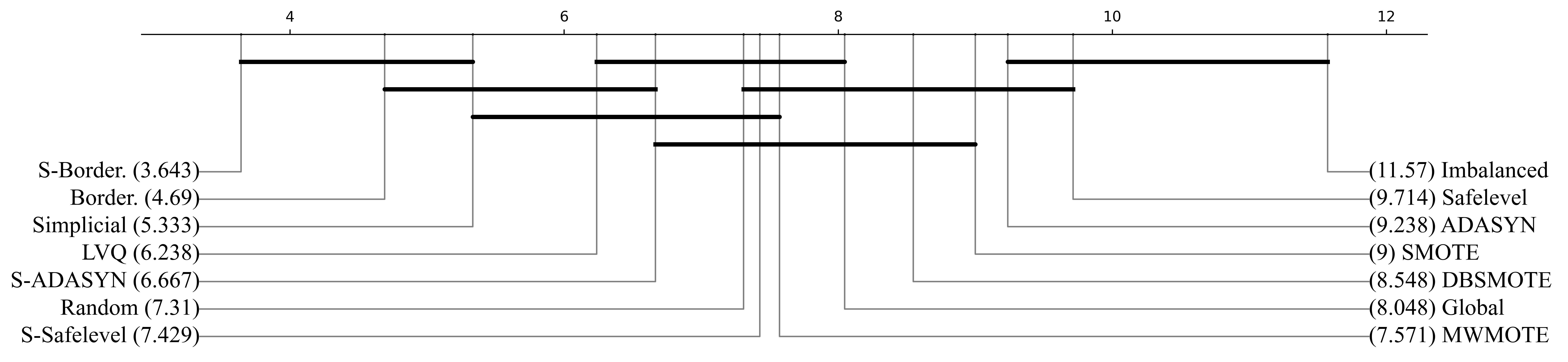}

\vspace{-0.5em}
\caption{Critical difference diagram for the $k$-NN classifier and F1 score.}
\label{cd_knn_f1}
\Description{Critical difference diagram for the k-NN classifier and F1 score.}
\end{figure*} 

\begin{table*}[h!]
\label{main_results1}

\caption{P-values of the Conover-Friedman post-hoc test for the $k$-NN classifier and Matthew's correlation coefficient.}

\centering

\resizebox{\textwidth}{!}{

\begin{tabular}{l|rrrrrrrrrr|rrrr}
\toprule
 & \fontsize{8}{9}\selectfont\textbf{Imbalanced} & \fontsize{8}{9}\selectfont\textbf{Random} & \fontsize{8}{9}\selectfont\textbf{Global} & \fontsize{8}{9}\selectfont\textbf{SMOTE} & \fontsize{8}{9}\selectfont\textbf{Border.} & \fontsize{8}{9}\selectfont\textbf{Safelevel} & \fontsize{8}{9}\selectfont\textbf{ADASYN} & \fontsize{8}{9}\selectfont\textbf{MWMOTE} & \fontsize{8}{9}\selectfont\textbf{DBSMOTE} & \fontsize{8}{9}\selectfont\textbf{LVQ} & \fontsize{8}{9}\selectfont\textbf{Simplicial} & \fontsize{8}{9}\selectfont\textbf{S-Border.} & \fontsize{8}{9}\selectfont\textbf{S-Safe.} & \fontsize{8}{9}\selectfont\textbf{S-ADASYN} \\ \midrule

\textbf{Imbalanced} &      1.000 &      0.192 &      0.001 &      0.047 &      0.000 &      0.534 &      0.035 &      0.004 &      0.443 &      0.003 &      0.000 &      0.000 &      0.017 &      0.000 \\
\textbf{Random} &      0.192 &      1.000 &      0.049 &      0.494 &      0.000 &      0.494 &      0.419 &      0.111 &      0.590 &      0.102 &      0.000 &      0.000 &      0.272 &      0.004 \\
\textbf{Global} &      0.001 &      0.049 &      1.000 &      0.199 &      0.094 &      0.008 &      0.246 &      0.709 &      0.013 &      0.740 &      0.090 &      0.005 &      0.384 &      0.340 \\
\textbf{SMOTE} &      0.047 &      0.494 &      0.199 &      1.000 &      0.003 &      0.172 &      0.901 &      0.362 &      0.221 &      0.340 &      0.003 &      0.000 &      0.678 &      0.026 \\
\textbf{Border.} &      0.000 &      0.000 &      0.094 &      0.003 &      1.000 &      0.000 &      0.005 &      0.041 &      0.000 &      0.045 &      0.983 &      0.246 &      0.011 &      0.468 \\
\textbf{Safelevel} &      0.534 &      0.494 &      0.008 &      0.172 &      0.000 &      1.000 &      0.136 &      0.023 &      0.884 &      0.021 &      0.000 &      0.000 &      0.075 &      0.000 \\
\textbf{ADASYN} &      0.035 &      0.419 &      0.246 &      0.901 &      0.005 &      0.136 &      1.000 &      0.431 &      0.178 &      0.407 &      0.004 &      0.000 &      0.771 &      0.035 \\
\textbf{MWMOTE} &      0.004 &      0.111 &      0.709 &      0.362 &      0.041 &      0.023 &      0.431 &      1.000 &      0.033 &      0.967 &      0.039 &      0.001 &      0.618 &      0.185 \\
\textbf{DBSMOTE} &      0.443 &      0.590 &      0.013 &      0.221 &      0.000 &      0.884 &      0.178 &      0.033 &      1.000 &      0.030 &      0.000 &      0.000 &      0.102 &      0.001 \\
\textbf{LVQ} &      0.003 &      0.102 &      0.740 &      0.340 &      0.045 &      0.021 &      0.407 &      0.967 &      0.030 &      1.000 &      0.043 &      0.002 &      0.590 &      0.199 \\ \hline
\textbf{Simplicial} &      0.000 &      0.000 &      0.090 &      0.003 &      0.983 &      0.000 &      0.004 &      0.039 &      0.000 &      0.043 &      1.000 &      0.254 &      0.011 &      0.455 \\
\textbf{S-Border.} &      0.000 &      0.000 &      0.005 &      0.000 &      0.246 &      0.000 &      0.000 &      0.001 &      0.000 &      0.002 &      0.254 &      1.000 &      0.000 &      0.060 \\
\textbf{S-Safelevel} &      0.017 &      0.272 &      0.384 &      0.678 &      0.011 &      0.075 &      0.771 &      0.618 &      0.102 &      0.590 &      0.011 &      0.000 &      1.000 &      0.069 \\
\textbf{S-ADASYN} &      0.000 &      0.004 &      0.340 &      0.026 &      0.468 &      0.000 &      0.035 &      0.185 &      0.001 &      0.199 &      0.455 &      0.060 &      0.069 &      1.000 \\ \bottomrule

\end{tabular}

}

\end{table*}

\begin{figure*}[!h] 


\centering
\includegraphics[width=0.9\textwidth]{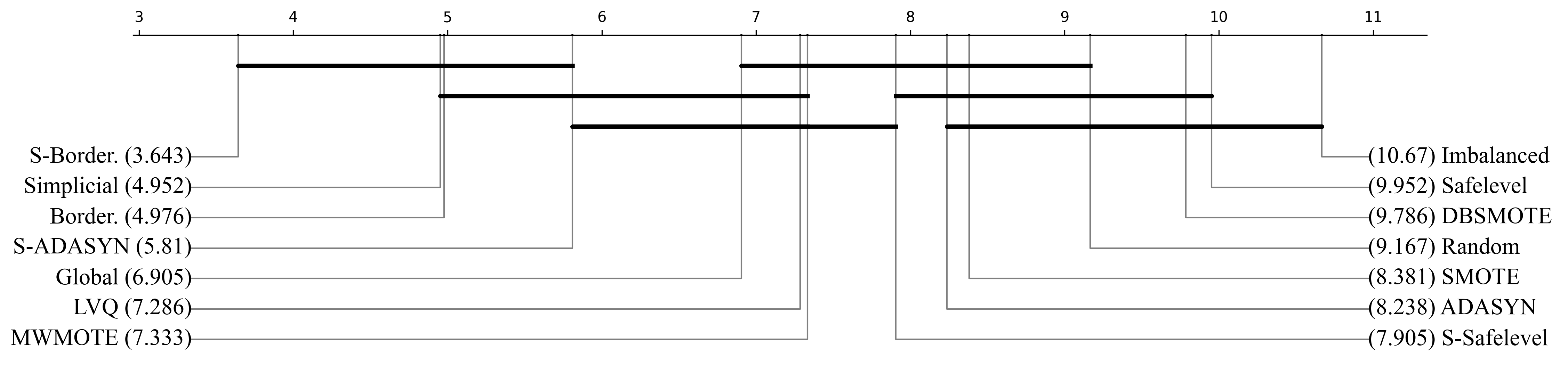}

\caption{Critical difference diagram for the $k$-NN classifier and Matthew's correlation coefficient.}\label{cd_knn_mcc}
\Description{Critical difference diagram for the k-NN classifier and MCC}

\end{figure*}

\begin{table*}[h!]

\caption{P-values of the Conover-Friedman post-hoc test for the gradient boosting classifier and F1 score.}
\label{pvalues_gb_f1}

\centering

\resizebox{\textwidth}{!}{

\begin{tabular}{l|rrrrrrrrrr|rrrr}
\toprule
 & \textbf{Imbalanced} & \textbf{Random} & \textbf{Global} & \textbf{SMOTE} & \textbf{Border.} & \textbf{Safelevel} & \textbf{ADASYN} & \textbf{MWMOTE} & \textbf{DBSMOTE} & \textbf{LVQ} & \textbf{Simplicial} & \textbf{S-Border.} & \textbf{S-Safe.} & \textbf{S-ADASYN} \\ \midrule

\textbf{Imbalanced} &      1.000 &      0.002 &      0.000 &      0.000 &      0.000 &      0.000 &      0.020 &      0.000 &      0.002 &      0.022 &      0.000 &      0.000 &      0.000 &      0.000 \\
\textbf{Random} &      0.002 &      1.000 &      0.203 &      0.163 &      0.060 &      0.593 &      0.388 &      0.366 &      0.934 &      0.366 &      0.000 &      0.009 &      0.024 &      0.093 \\
\textbf{Global} &      0.000 &      0.203 &      1.000 &      0.902 &      0.538 &      0.460 &      0.033 &      0.712 &      0.176 &      0.030 &      0.016 &      0.176 &      0.324 &      0.681 \\
\textbf{SMOTE} &      0.000 &      0.163 &      0.902 &      1.000 &      0.622 &      0.388 &      0.024 &      0.622 &      0.140 &      0.022 &      0.022 &      0.218 &      0.388 &      0.774 \\
\textbf{Border.} &      0.000 &      0.060 &      0.538 &      0.622 &      1.000 &      0.176 &      0.006 &      0.324 &      0.049 &      0.006 &      0.071 &      0.460 &      0.712 &      0.837 \\
\textbf{Safelevel} &      0.000 &      0.593 &      0.460 &      0.388 &      0.176 &      1.000 &      0.163 &      0.712 &      0.538 &      0.151 &      0.002 &      0.037 &      0.085 &      0.250 \\
\textbf{ADASYN} &      0.020 &      0.388 &      0.033 &      0.024 &      0.006 &      0.163 &      1.000 &      0.078 &      0.435 &      0.967 &      0.000 &      0.001 &      0.002 &      0.011 \\
\textbf{MWMOTE} &      0.000 &      0.366 &      0.712 &      0.622 &      0.324 &      0.712 &      0.078 &      1.000 &      0.324 &      0.071 &      0.006 &      0.085 &      0.176 &      0.435 \\
\textbf{DBSMOTE} &      0.002 &      0.934 &      0.176 &      0.140 &      0.049 &      0.538 &      0.435 &      0.324 &      1.000 &      0.411 &      0.000 &      0.007 &      0.020 &      0.078 \\
\textbf{LVQ} &      0.022 &      0.366 &      0.030 &      0.022 &      0.006 &      0.151 &      0.967 &      0.071 &      0.411 &      1.000 &      0.000 &      0.000 &      0.002 &      0.010 \\ \hline 
\textbf{Simplicial} &      0.000 &      0.000 &      0.016 &      0.022 &      0.071 &      0.002 &      0.000 &      0.006 &      0.000 &      0.000 &      1.000 &      0.286 &      0.151 &      0.045 \\
\textbf{S-Border.} &      0.000 &      0.009 &      0.176 &      0.218 &      0.460 &      0.037 &      0.001 &      0.085 &      0.007 &      0.000 &      0.286 &      1.000 &      0.712 &      0.345 \\
\textbf{S-Safelevel} &      0.000 &      0.024 &      0.324 &      0.388 &      0.712 &      0.085 &      0.002 &      0.176 &      0.020 &      0.002 &      0.151 &      0.712 &      1.000 &      0.565 \\
\textbf{S-ADASYN} &      0.000 &      0.093 &      0.681 &      0.774 &      0.837 &      0.250 &      0.011 &      0.435 &      0.078 &      0.010 &      0.045 &      0.345 &      0.565 &      1.000 \\ \bottomrule

\end{tabular}

}

\end{table*} 

\begin{table*}[h!]
\label{main_results1}

\caption{P-values of the Conover-Friedman post-hoc test for the gradient boosting classifier and Matthew's correlation coefficient.}
\vspace{-0.75em}

\centering

\renewcommand{\arraystretch}{1.1}

\resizebox{\textwidth}{!}{

\begin{tabular}{l|rrrrrrrrrr|rrrr}
\hline
 & \fontsize{8}{9}\selectfont\textbf{Imbalanced} & \fontsize{8}{9}\selectfont\textbf{Random} & \fontsize{8}{9}\selectfont\textbf{Global} & \fontsize{8}{9}\selectfont\textbf{SMOTE} & \fontsize{8}{9}\selectfont\textbf{Border.} & \fontsize{8}{9}\selectfont\textbf{Safelevel} & \fontsize{8}{9}\selectfont\textbf{ADASYN} & \fontsize{8}{9}\selectfont\textbf{MWMOTE} & \fontsize{8}{9}\selectfont\textbf{DBSMOTE} & \fontsize{8}{9}\selectfont\textbf{LVQ} & \fontsize{8}{9}\selectfont\textbf{Simplicial} & \fontsize{8}{9}\selectfont\textbf{S-Border.} & \fontsize{8}{9}\selectfont\textbf{S-Safe.} & \fontsize{8}{9}\selectfont\textbf{S-ADASYN} \\ \hline
\hline

\textbf{Imbalanced} &      1.000 &      0.001 &      0.005 &      0.000 &      0.001 &      0.001 &      0.042 &      0.011 &      0.077 &      0.218 &      0.000 &      0.002 &      0.001 &      0.002 \\
\textbf{Random} &      0.001 &      1.000 &      0.672 &      0.299 &      0.787 &      0.908 &      0.233 &      0.488 &      0.144 &      0.046 &      0.155 &      0.939 &      0.939 &      0.969 \\
\textbf{Global} &      0.005 &      0.672 &      1.000 &      0.144 &      0.488 &      0.590 &      0.441 &      0.787 &      0.299 &      0.115 &      0.065 &      0.729 &      0.616 &      0.700 \\
\textbf{SMOTE} &      0.000 &      0.299 &      0.144 &      1.000 &      0.441 &      0.355 &      0.026 &      0.084 &      0.013 &      0.003 &      0.700 &      0.264 &      0.336 &      0.281 \\
\textbf{Border.} &      0.001 &      0.787 &      0.488 &      0.441 &      1.000 &      0.878 &      0.144 &      0.336 &      0.084 &      0.024 &      0.248 &      0.729 &      0.847 &      0.758 \\
\textbf{Safelevel} &      0.001 &      0.908 &      0.590 &      0.355 &      0.878 &      1.000 &      0.191 &      0.419 &      0.115 &      0.035 &      0.191 &      0.847 &      0.969 &      0.878 \\
\textbf{ADASYN} &      0.042 &      0.233 &      0.441 &      0.026 &      0.144 &      0.191 &      1.000 &      0.616 &      0.787 &      0.419 &      0.009 &      0.264 &      0.204 &      0.248 \\
\textbf{MWMOTE} &      0.011 &      0.488 &      0.787 &      0.084 &      0.336 &      0.419 &      0.616 &      1.000 &      0.441 &      0.191 &      0.035 &      0.538 &      0.441 &      0.513 \\
\textbf{DBSMOTE} &      0.077 &      0.144 &      0.299 &      0.013 &      0.084 &      0.115 &      0.787 &      0.441 &      1.000 &      0.590 &      0.004 &      0.166 &      0.124 &      0.155 \\
\textbf{LVQ} &      0.218 &      0.046 &      0.115 &      0.003 &      0.024 &      0.035 &      0.419 &      0.191 &      0.590 &      1.000 &      0.001 &      0.055 &      0.038 &      0.050 \\ \hline
\textbf{Simplicial} &      0.000 &      0.155 &      0.065 &      0.700 &      0.248 &      0.191 &      0.009 &      0.035 &      0.004 &      0.001 &      1.000 &      0.134 &      0.178 &      0.144 \\
\textbf{S-Border.} &      0.002 &      0.939 &      0.729 &      0.264 &      0.729 &      0.847 &      0.264 &      0.538 &      0.166 &      0.055 &      0.134 &      1.000 &      0.878 &      0.969 \\
\textbf{S-Safelevel} &      0.001 &      0.939 &      0.616 &      0.336 &      0.847 &      0.969 &      0.204 &      0.441 &      0.124 &      0.038 &      0.178 &      0.878 &      1.000 &      0.908 \\
\textbf{S-ADASYN} &      0.002 &      0.969 &      0.700 &      0.281 &      0.758 &      0.878 &      0.248 &      0.513 &      0.155 &      0.050 &      0.144 &      0.969 &      0.908 &      1.000 \\ \hline

\end{tabular}

}

\end{table*} 
\begin{figure*}[!h] 


\centering
\includegraphics[width=0.9\textwidth]{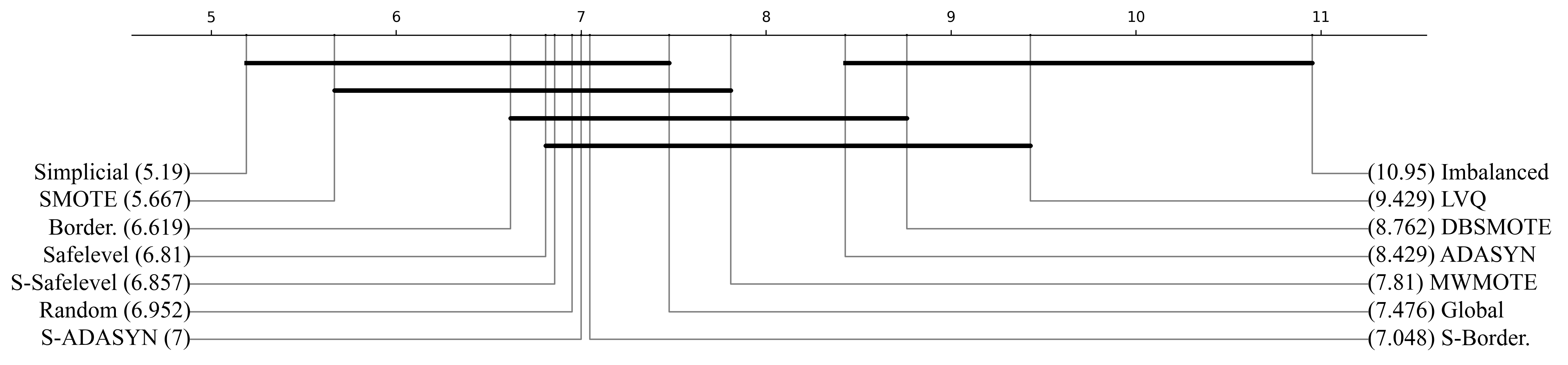}

\vspace{-1em}
\caption{Critical difference diagram for the gradient boosting classifier and Matthew's correlation coefficient.}

\end{figure*} 


\end{document}